\documentclass{article}

\usepackage{arxiv}

\usepackage[utf8]{inputenc} % allow utf-8 input
\usepackage[T1]{fontenc}    % use 8-bit T1 fonts
\usepackage{hyperref}       % hyperlinks
\usepackage{url}            % simple URL typesetting
\usepackage{booktabs}       % professional-quality tables
\usepackage{amsfonts}       % blackboard math symbols
\usepackage{nicefrac}       % compact symbols for 1/2, etc.
\usepackage{microtype}      % microtypography
\usepackage{lipsum}
\usepackage{graphicx}
\usepackage{subcaption}
\usepackage{times}
\usepackage{epsfig}
\usepackage{amsmath}
\usepackage{amssymb}
\usepackage{float}
\usepackage{multirow}
\usepackage[nodisplayskipstretch]{setspace}

\graphicspath{ {./images/} }

\title{Counterfactual Generation with Knockoffs}

\author{
 Oana-Iuliana Popescu \\
  Friedrich-Schiller University, Jena\\
  \texttt{oana-iuliana.popescu@uni-jena.de} \\
   \And
 Maha Shadaydeh \\
  Friedrich-Schiller University, Jena\\
  \texttt{maha.shadaydeh@uni-jena.de} \\
  \And
 Joachim Denzler \\
  Friedrich-Schiller University, Jena\\
  Institute for Data Science of the German Aerospace Center, Jena \\
  \texttt{joachim.denzler@uni-jena.de} \\
}

\begin{document}
\maketitle
\begin{abstract}
Human interpretability of deep neural networks' decisions is crucial, especially in domains where these directly affect human lives. Counterfactual explanations of already trained neural networks can be generated by perturbing input features and attributing importance according to the change in the classifier's outcome after perturbation. Perturbation can be done by replacing features using heuristic or generative in-filling methods. The choice of in-filling function significantly impacts the number of artifacts, i.e., false-positive attributions. Heuristic methods result in false-positive artifacts because the image after the perturbation is far from the original data distribution. Generative in-filling methods reduce artifacts by producing in-filling values that respect the original data distribution. However, current generative in-filling methods may also increase false-negatives due to the high correlation of in-filling values with the original data. In this paper, we propose to alleviate this by generating in-fillings with the statistically-grounded Knockoffs framework, which was developed by Barber and Candès in 2015 as a tool for variable selection with controllable false discovery rate. Knockoffs are statistically null-variables as decorrelated as possible from the original data, which can be swapped with the originals without changing the underlying data distribution. A comparison of different in-filling methods indicates that in-filling with knockoffs can reveal explanations in a more causal sense while still maintaining the compactness of the explanations. 
\end{abstract}

%%%%%%%%% BODY TEXT
\section{Introduction}

Human interpretability of predictive models' decisions is crucial for model validation, especially in critical domains such as medicine, where these decisions directly impact human lives. Decisions can be explained with post-hoc attribution methods, which reveal how much each feature contributes to an already trained model's decision \cite{ancona17}. Post-hoc methods are local interpretability methods, i.e., they explain the decision of the network for one sample and can be split into perturbation-based and gradient-based. Perturbation-based methods inspect the effect of changing or removing input features on the output. They pose a "what-if" question to the predictive model, thus generating counterfactuals. Gradient-based methods answer a "why" question using the gradients of the output with respect to the input features.

Counterfactual generation approaches differ in the way the feature is perturbed. Given that neural networks cannot handle missing values, a feature cannot be removed entirely. Therefore, perturbation aims to simulate its absence by replacing it with a reference value computed with an in-filling function. In-filling functions have been categorized into heuristic and generative, and it has been shown that the choice of function significantly impacts the number of artifacts, i.e., false-positive attributions, in the explanations \cite{chang18}.  Heuristic reference values generate more artifacts because the perturbed images are far from the original data distribution and thus bias the classifier. Generative in-filling methods reduce these artifacts using strong generative models to fill-in missing values according to the initial data distribution \cite{chang18}. However,  generative in-filling methods may  also increase false-negatives due to the high correlation of  in-filling values with the original data.  Such  generative  models  can also predict redundant features and hence  those features do not get attributed importance, which imposes a regularization effect. While this regularization considerably reduces the number of artifacts, the effect might also hold for less redundant features. This can lead to false-negatives, i.e., fewer pixels being signalized than would belong to the causal counterfactual. Therefore, we are interested in generating in-filling values as decorrelated as possible from the original data while keeping the perturbed image in-distribution. 

In this paper, we propose to use another theoretically-grounded approach, the Knockoffs framework \cite{Barber_2015, candes16}, to reference value generation. This approach was developed for variable selection problem using conditional independence testing with controllable false discovery rate \cite{Barber_2015}, which pursues the same goal as attribution methods: to explain which features are important for the outcome. Knockoffs are statistical null variables, i.e., they do not introduce any new information to the model. They are designed to be as decorrelated as possible from the original data but still belong to the original data distribution. They can therefore be swapped with the original variables without introducing additional bias to the model. In the original framework, they are used to compare the coefficients of the original variables and knockoffs when doing regression: a larger coefficient for the knockoff indicates that the variable is not predictive. Knockoffs work for multiple data distributions \cite{Barber_2015}, and with arbitrary models \cite{watson2019testing}.

Our contribution is the exploration of knockoffs as generative in-filling values. We present a comparison of in-filling with knockoffs, heuristic, and generative methods. We also compare the generated counterfactual explanations with three gradient-based explanations. We generate these explanations for convolutional neural networks (CNNs) trained to do classification tasks on hand-written digits using the MNIST dataset \cite{mnist10}. We show that knockoffs have the potential to deliver counterfactual explanations in a more causal sense, indicating which regions should be changed to transform one sample into a sample of another class while still keeping the explanations as compact as possible.

% Therefore, they can be swapped with the original values of a dataset without introducing bias to the predictive model. 

% Knockoffs \cite{Barber_2015, candes16} are statistical null-values that are as decorrelated from the original data as possible while still respecting the original data distribution. 

%-------------------------------------------------------------------------

\section{Related Work}

\subsection{Attribution Methods}

% There are multiple definitions of interpretability. We will use the definition of \cite{doshivelez2017rigorous}: interpretability is "the ability to explain or to present in understandable terms to a human." Interpretability serves essential purposes such as building trust, detecting bias, encouraging fair models, inspecting robustness, and reliability of models \cite{doshivelez2017rigorous} and enhancing human learning \cite{goyal19a}. Interpretability can be achieved with, among others, post-hoc attribution methods. These explain how much each feature contributes to the model's decision \cite{ancona17} locally, for one sample only, or globally, for a set of images belonging to a particular class or the entire dataset. We focus on local explanations, which are separated by their computational approach into perturbation-based and gradient-based methods. 

% It can be obtained through intrinsic or post-hoc methods. Intrinsic methods restrict model complexity to functions with interpretable coefficients, such as linear models, at the cost of predictive power \cite{lipton18}. Post-hoc interpretability methods extract information from the model after training and do not impose restrictions on the model class. Since they are developed by humans, they might be biased by the human's need for consistency and understandability \cite{lipton18}, but are are preferred when reducing predictive power is impossible, such as in computer vision.

Perturbation-based methods attribute importance by inspecting the changes in the network's output when the input features have different values \cite{Ancona2019GradientBasedAM}. This is usually approached as an optimization problem where the objective is to find the region that affects the classifier output maximally. This objective is  formulated by \cite{dabkowski17} as finding the following two regions:

\begin{itemize}
    \item[--] \textbf{Smallest Deletion Region (SDR)}: the smallest region that, when removed, prevents confident classification;
    \item[--] \textbf{Smallest Supporting Region (SSR)}: the smallest region that can alone allows a confident classification.
\end{itemize}

This objective translates into finding a binary mask that describes which pixels to perturb to change or support the model's decision. This mask can be found either by directly optimizing the mask \cite{fong17} or by formulating an alternative objective. One work trains a model to learn masks \cite{dabkowski17}, while \cite{chang18} formulate the objective as doing per-pixel dropout and optimizing dropout probabilities. 

Perturbations are done through replacement with a reference value obtained from an in-filling function, which can be heuristic or generative \cite{chang18}. A well-known heuristic method is pixel occlusion \cite{zeiler13}. Other heuristics were proposed in \cite{fong17} to generate reference values closer to the initial input values, e.g., by using a blurred version of the image or by adding Gaussian noise to initial pixel values. It has been shown in \cite{chang18} that the images after perturbation with heuristics are far from the original data distribution and thus bias the classifier and, implicitly, the explanation. To alleviate this problem, Chang et al. \cite{chang18} proposes an in-filling function using a conditional generative model, in the form of a variational auto-encoder (VAE) \cite{kingma13} or a generative adversarial network (GAN) \cite{gan}. These learn to replace masked-out pixels in the image according to the original data distribution.

Back-propagation methods measure the classifier's sensitivity to small perturbations in the input using the partial derivative of the network's output with respect to the input features. Their main advantage over perturbation-based methods is faster computation, but the resulting explanations are noisy and vary with other features that are not of interest \cite{ancona17}. However, methods have been further developed to improve the stability of the gradients. Gradient $\times$ Input multiplies gradients with the input features, while Integrated Gradients \cite{sundararajan17}, Layer-wise Relevance Propagation (LRP) \cite{Binder2016LayerWiseRP}, or DeepLIFT \cite{DBLP:journals/corr/deeplift} average over gradients between the input and a baseline. SmoothGrad \cite{smoothgrad} samples a larger region around the input feature by adding Gaussian noise and can be used with any of the methods mentioned above. Class activation maps (CAM) explore which higher-level features, extracted by the last layers of the neural network, are important by computing the gradients of the output layers with respect to the weights of the last convolutional feature layer \cite{ZhouKLOT15, grad_cam}.

\subsection{Knockoffs}

Given a set of observed variables, a response variable, and a predictive model, variable selection defines which subset of variables is important for the response. It is, therefore, similar in concept to global interpretability methods, which explain decisions by finding relevant variables for the decision over the entire dataset. Often, observed variables are marginally dependent and might become independent if some other variable is already observed. For example, income and health status are strongly correlated, and it might suffice to know one of the variables for the outcome. This scenario often occurs in high-dimensional datasets. In such cases, the Knockoffs framework can be used to test for conditional independence, i.e., whether a variable adds more information than what is already known from the already used variables \cite{candes16}.

Knockoffs are constructed to be in-distribution null-variables. To be in-distribution, knockoffs are exchangeable with the original variables: the correlation between knockoffs is the same as the correlation between the original variables \cite{candes16}. To be null-variables, knockoffs do not contain any information about the outcome $Y$. Formally, this translates to the construction of knockoffs  $\tilde{X}=(\tilde{X}_1,...,\tilde{X}_n)$ given the features $X=(X_1,...,X_n)$ such that they fulfill the two properties as follows:
\begin{itemize}
    \item[--] \textbf{Exchangeability} $\forall S \subset {1,...n} \quad (X, \tilde{X})_{swap(S)}=(\tilde{X}, X)$, where $(X, \tilde{X})_{swap(S)}=(\tilde{X}, X)$ is obtained by swapping the features $X_j$, indexed by set $j \in S$ with their knockoffs $\tilde{X}_j$;
    \item[--] \textbf{Null variables} $\tilde{X} \perp Y | X$, which is guaranteed if $\tilde{X}$ is constructed without looking at $Y$.
\end{itemize}

Initially designed for linear models and Gaussian data distributions, knockoffs have been extended by \cite{candes16} to non-linear predictive models and arbitrary data distributions and by \cite{watson2019testing} to work with arbitrary predictive models. \cite{Romano19, duarte2020knockoffinspired} have proposed knockoff generation methods for arbitrary data distributions.

%------------------------------------------------------------------------
\section{Counterfactual Generation with Knockoffs}

Motivated by the over-regularization imposed by existing generative in-filling functions, we propose to use knockoffs as an in-filling method for generating counterfactual explanations and combine knockoffs with the per-pixel dropout approach of Fill-In Drop-Out (FIDO) \cite{chang18} for mask generation. 

In the sections that follow, we denote the classifier function as $f$. Its output given input $x$, a probability distribution over the classes, is denoted as $f(x)$. To refer to the probability of a specific target class $c$, we will denote it as $f(c|x)$.

\subsection{Counterfactual generation}
\label{subsec:cf_generation}
 
The FIDO algorithm \cite{chang18} starts by sampling a binary mask, which indicates the pixels to be perturbed. Reference values are then computed using the in-filling function, and perturbation is done by replacing original feature values with the reference values. The resulting image is then fed to the classifier. The classifier output is used to update the per-pixel dropout probabilities from which a new binary mask can be sampled. This procedure is repeated for several iterations to optimize the SSR and SDR objectives. The saliency map is then generated from the counterfactual explanation in the form of dropout probabilities.

\paragraph{Mask sampling} FIDO defines counterfactual generation as doing per-pixel dropout on the $N$ image pixels with dropout probabilities $\theta \in [0,1]^{N}$ \cite{chang18}. The dropout probabilities are used to sample binary masks $z \in \{0,1\}^{N}$ from a Bernoulli distribution $q$ with parameter $\theta$ and should be found such that the objective is optimal using masks sampled from the distribution. Therefore, the optimization of the SSR and SDR objectives is done with respect to $\theta$. Since computing gradients for the masks in binary form is not possible, Concrete dropout \cite{gal2017concrete} is used to relax the Bernoulli distribution to a continuous distribution. The Concrete distribution approximates the binary mask $z$ as $\tilde{z}$ according to Equation \ref{eq:concrete}, where $u$ is a random uniform variable $u \sim Unif(0,1)$.

\begin{equation}
    \tilde{z}=\textrm{sigmoid}(\frac{1}{t})\cdot (\log p ) - \log(1-p) + \log u - \log(1-u))
    \label{eq:concrete}
\end{equation}

\paragraph{Value swapping} To perturb the features of a sample $x$ from the dataset, pixels are replaced as indicated by the mask, i.e., where $\tilde{z}=1$, using the in-filling function. The result is a new image $\hat{x}$ as described by function $\Theta$ in Equation \ref{eq:infill}, where $\odot$ is the Hadamard element-wise multiplication, which we call the corrupted image.

\begin{equation}
    \Theta(x, \tilde{z}) = \tilde{z} \odot x + (1-\tilde{z}) \odot \hat{x}
    \label{eq:infill}
\end{equation}

\paragraph{Mask parameters update} The corrupted image is fed to the classifier. The outcome is used to compute the log-odds score $s$ as described in Equation \ref{eq:logodds}. The score is plugged in the SSR or SDR objectives as in Equations \ref{eq:ssr_obj} and \ref{eq:sdr_obj}, respectively.

\begin{equation}
    s(c|x) = \log f(c|x) - \log(1-f(c|x))
    \label{eq:logodds}
\end{equation}

\begin{equation}
    \theta_{SSR} = \min_{\mathbb{E}_{q_{\theta}(z)}} s(c|\phi(x, \tilde{z})) + \lambda \|\tilde{z}\|_1\
    \label{eq:ssr_obj}0
\end{equation}
\begin{equation}
    \theta_{SDR} = \min_{\mathbb{E}_{q_{\theta}(z)}} -s(c|\phi(x, \tilde{z})) + \lambda \|1-\tilde{z}\|_1
    \label{eq:sdr_obj}.
\end{equation}

To speed up convergence and cover more possible solutions, multiple dropout masks can be sampled at an iteration, and the parameters can be updated using mini-batch gradient descent. Additional regularization can be done using Total Variation (TV) \cite{dabkowski17} or by optimizing the objectives for the down-sampled image. The resulting counterfactual, represented by the parameter $\theta$, which has values between 0 and 1, is visualized as a diverging saliency map by subtracting 0.5 from $\theta$ \cite{chang18}.

\subsection{In-filling methods}

\paragraph{Knockoffs}

We use knockoffs to perturb the feature of interest $r$ by taking the respective feature value from the knockoff image $\tilde{x}$. The resulting in-filling function can be defined mathematically as $\hat{x} \sim p(x_r|x_{-r}, \tilde{x}_{1:r-1})$.

To generate knockoffs, we use a VAE trained on the original dataset as proposed in \cite{duarte2020knockoffinspired}, from which we obtain the necessary conditional probability distributions for the generation process. The generation process starts by marginalizing feature $j$ by setting it to 0. Then, the modified image is passed through the VAE encoder to infer the latent variables $z^*$, which are then decoded into a new image $x^*$. The knockoff value for feature $j$ is then set to the value of the feature from the decoded image $x^*_j$. This procedure is repeated for all features in the image, each time starting from the last obtained knockoff image.

\begin{figure*}[]
    \centering\includegraphics[width=15cm]{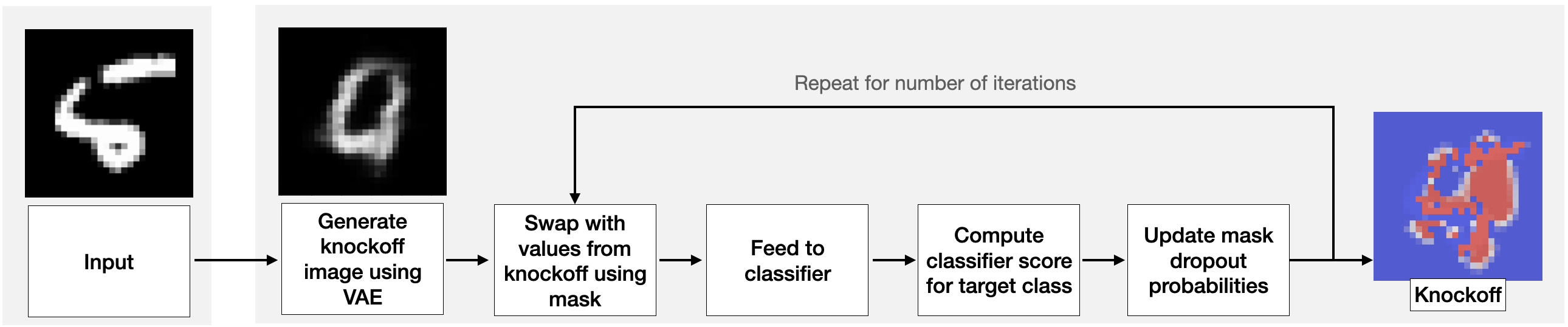}
    \caption{In-filling using knockoffs.}

    \label{fig:infill_knockoff_flow}
\end{figure*}

To generate counterfactuals, we use the knockoff image to replace the original with knockoff values as previously described in Equation \ref{eq:infill}. This process is also illustrated in Figure \ref{fig:infill_knockoff_flow}. The corrupted image is then fed to the classifier to obtain the classifier score, and the log-odds score is computed. Then, the objective is optimized, and the dropout probabilities of the mask are updated. To avoid computational overhead, we only generate one knockoff image, which we repeatedly use.

\paragraph{Flip}

One of the heuristic methods used in FIDO is the mean value of the dataset. Due to the binary nature of MNIST, using the mean would lead to counterfactuals signalizing the entire background as important. We are rather interested in important regions within the number. Therefore, our heuristic flips the bit value feature of interest $r$ to the dataset's background value, as defined by the function $\hat{x}_r = 0$.

\paragraph{Generative}

For the generative in-filling function, we use a convolutional VAE, which we train to do in-filling by randomly masking $7 \times 7$ pixel squares of the image. We then use it to replaces feature of interest $r$ by drawing possible values from its probability distribution conditioned on the remaining pixels $\hat{x}_r \sim p(x_r|x_{-r})$. From now on, we denote this in-filling function as VAE.

\subsection{Evaluation metrics}
\label{sec:eval}

\paragraph{Knockoffs quality evaluation} The quality of the knockoffs depends on the quality of the approximation of the data generating model. This translates to the VAE being able to generate new, diverse samples of good quality; otherwise, the knockoffs will be copies of the original images. To test the knockoffs' quality and compare them with the other in-filling methods, we assess how well the perturbation decorrelates the image from the original while still keeping the data in-distribution, similarly to \cite{chang18}. We quantify the correlation using the multi-structural similarity index (MS-SSIM) \cite{msssim} between original images and perturbed images, and the in-distribution-ness as the classifier target probability. We perturb the images by randomly replacing pixels with the different in-filling methods and compare the trade-off between correlation and in-distribution-ness for the different in-filling methods. Ideally, MS-SSIM should be as low as possible, while the target class probability should remain high.

\paragraph{Saliency map evaluation}To evaluate the SSR and SDR saliency maps (see definition in Section \ref{subsec:cf_generation}), we are mainly interested in the saliency metric (SM), which quantifies the information content and compactness of the salient region \cite{dabkowski17}. SM is computed by cropping the image to the saliency bounding box and upscaling the cropped image to the original size. The saliency bounding box is obtained by thresholding the relevance values at a certain level and taking the minimum bounding box that contains those pixels. The upscaled image is then fed to the classifier, and the SM is computed for image $x$ with saliency bounding box $b$ as $\log \max(\frac{area(b)}{area(x)}, 0.05) - \log p(c_{target}|CropAndUpscale(x, b))$. A negative score indicates that the saliency map has high information content in a small image area. 

To make the comparison with other methods more thorough, we also compute the weakly supervised location (WSL) metric. WSL quantifies how well the saliency maps find the object of interest as the percentage of correct saliency predictions \cite{dabkowski17}. A prediction is correct if the intersection over union (IOU) of the saliency bounding box with the ground truth bounding box is over 0.5.

However, our main metric remains the SM. With counterfactuals, we are rather interested in finding the causal regions that change the classifier's outcome, and consider that SM reveals the causal contribution: if the attributed region is not causal for the classifier outcome, it will return another label. Suppose the classifier learns the class 'car' only due to a confounding concept, e.g., the road. Then, the WSL score of an attribution method indicating the road would be near 0 (no attribution to the car). However, the SM score would be maximal because the confounding object was attributed - this is the correct causal explanation. which might be smaller than the object or even lie slightly outside of it. 

% -------------------------------------------------------------------

\section{Experimental results}

We evaluate and compare our proposed method on classifiers trained on the entire MNIST \cite{mnist10} dataset and two subsets derived from it, which will be described in the subsequent sections. We start with the presentation of the results on the entire MNIST dataset, for which we evaluate the metrics from Section \ref{sec:eval}. We then continue with a qualitative evaluation of the two subsets of the MNIST dataset. 

We explain pre-trained ResNet-18 classifiers, which were trained with a learning rate of 0.003 and a batch size of 256 for ten epochs. For the VAE in-filling models, we train convolutional VAEs with a learning rate of 0.0002 and a batch size of 256 for 200 epochs. For knockoff generation, we train VAE models with a learning rate of 0.0002 and a batch size of 128 for 500 epochs. 

To optimize the SSR and SDR objectives, we also use Adam \cite{kingma2014adam} with a learning rate of $\lambda=1-e^3$ unless otherwise specified. We sample batches of 8 masks and use a TV factor 0.01 for all datasets.

% -------------------------------------------------------------------

\subsection{Counterfactual explanations on MNIST}

We evaluate our approach on the benchmark dataset MNIST \cite{mnist10}, because its binary nature makes a clear delimitation between background and foreground, and the notion of counterfactual might be more intuitive on numbers. On MNIST, counterfactuals should indicate how a number could be modified to belong to a different class. An example is illustrated in Figure \ref{fig:38_counterfactual}, where pixels marked as red should be removed or added to transform a three into an eight and an eight into a three, respectively.

% -------------------------------------------------------------------
\paragraph{Knockoff evaluation}
\label{subsec:px_mnist_knockoff}

\begin{figure}
\centering
\begin{minipage}{.3\textwidth}
  \centering
     \includegraphics[width=4.5cm]{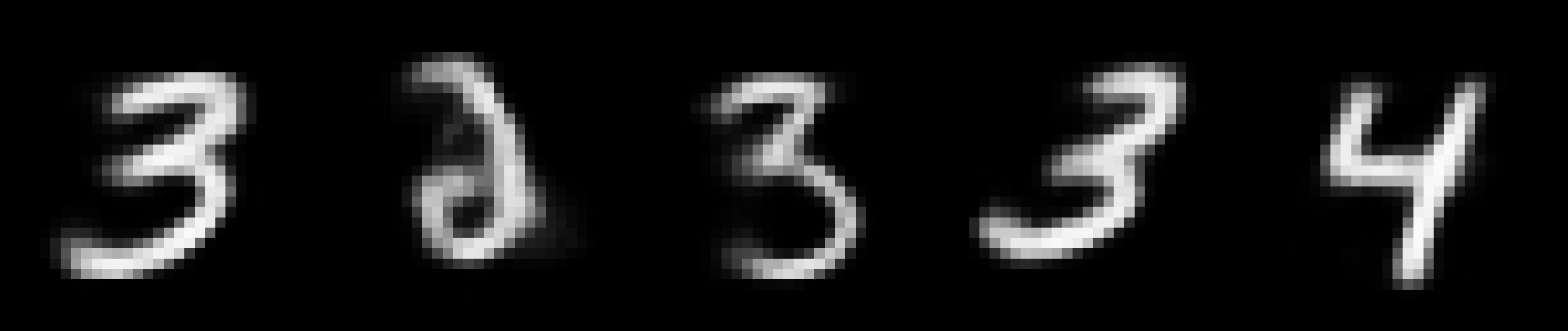}
    \includegraphics[width=4.5cm]{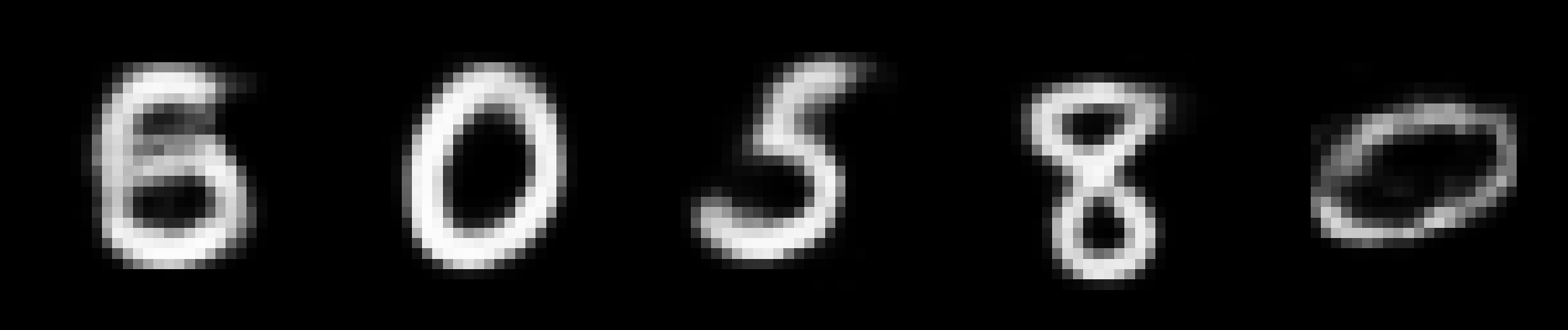}

    \includegraphics[width=4.5cm]{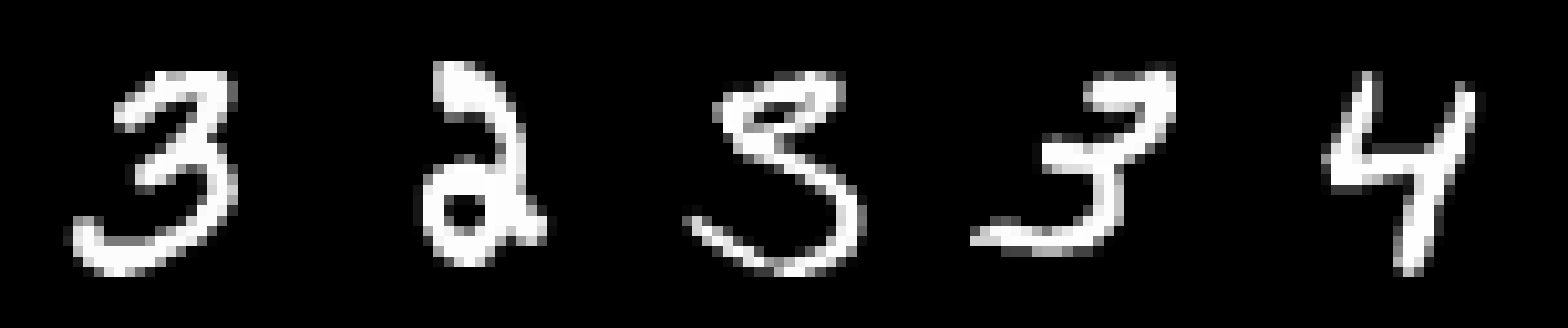}
    \includegraphics[width=4.5cm]{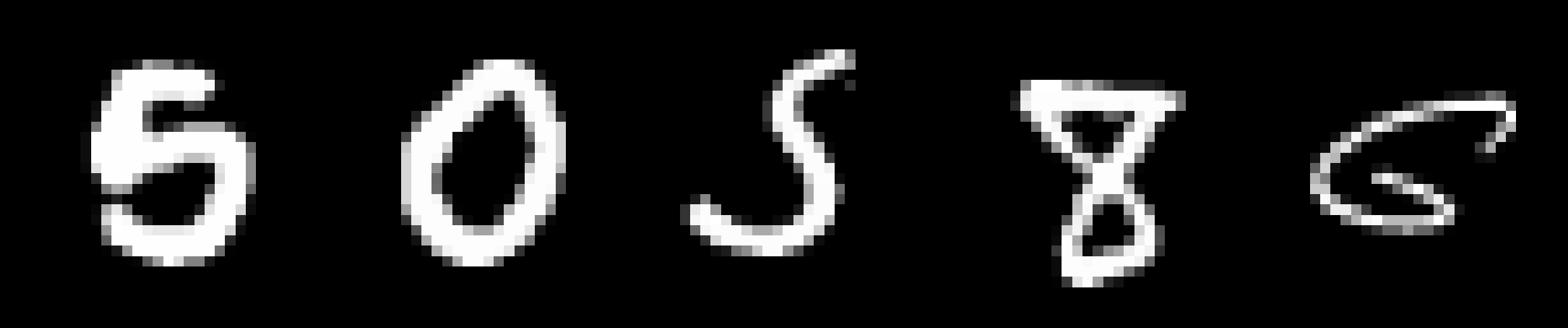}
    
    \caption{Examples of knockoffs from the VAE with latent code size 5. Upper row: knockoffs, lower row: original images.}
    \label{fig:mnist_knockoffs_5z}
\end{minipage} \qquad %
\begin{minipage}{.6\textwidth}
  \centering
  \includegraphics[width=6cm]{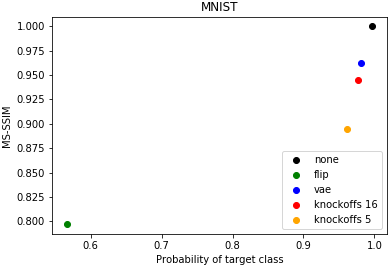}
    \caption{Structural similarity index (MS-SSIM) versus probability of target class for randomly corrupted pixels with the in-filling methods: without intervention (none), flip (flip), VAE (vae), knockoffs from a VAE with code size 5 (knockoff 5) and code size 16 (knockoff 16). Knockoffs from a VAE with latent code size 5 perform best.}
    \label{fig:mnist_corrupted}
\end{minipage}
\end{figure}

To generate knockoffs,  we use two VAEs with two latent code sizes: 5 and 16, and compare them with the flip and VAE in-fillings, as shown in Figure \ref{fig:mnist_corrupted}. We find that the VAE with code size 16 overfits, leading to a bad trade-off between in-distribution-ness and correlation. We, therefore, continue our experiments with the knockoffs from the VAE with code size 5, from which a few examples can be seen in Figure \ref{fig:mnist_knockoffs_5z}. In general, knockoffs tend to be a prototype of the class number or a number that is complementary to the original, similar to a distractor image.

% -------------------------------------------------------------------
\paragraph{Counterfactual generation}
\label{subsec:results1}

\begin{figure}[H]
    \centering\includegraphics[width=1.0cm]{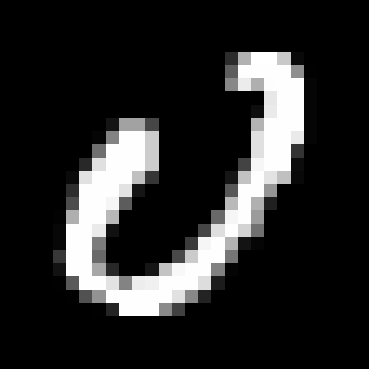}
    \centering\includegraphics[width=1.0cm]{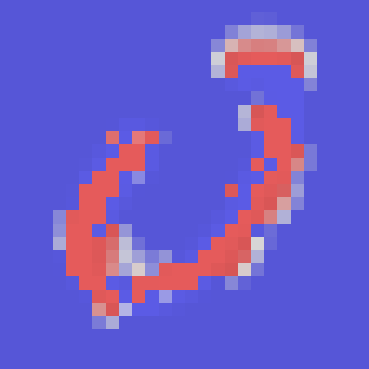}
    \centering\includegraphics[width=1.0cm]{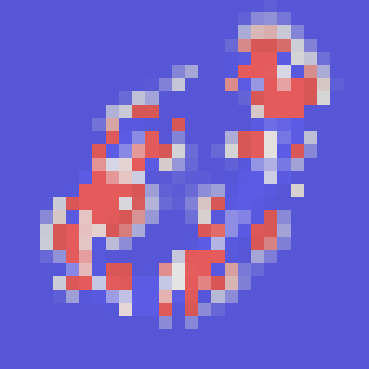}
    \centering\includegraphics[width=1.0cm]{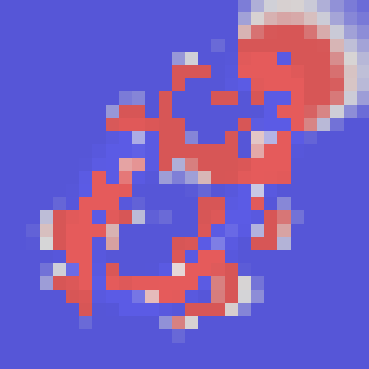}
    \centering\includegraphics[width=1.0cm]{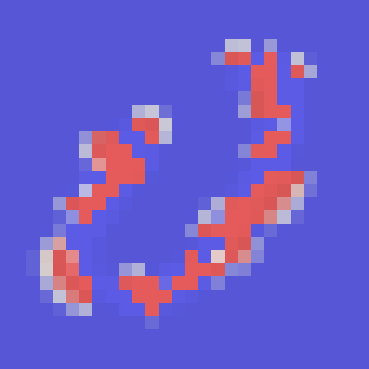}
    \centering\includegraphics[width=1.0cm]{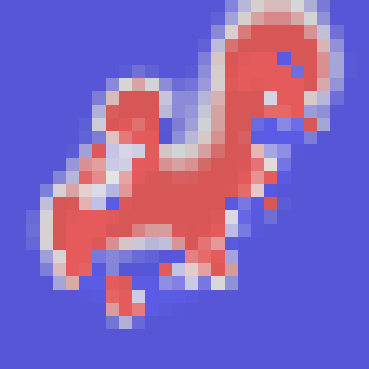}
    \centering\includegraphics[width=1.0cm]{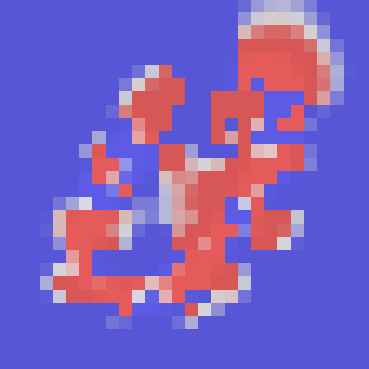}
    \centering\includegraphics[width=1.0cm]{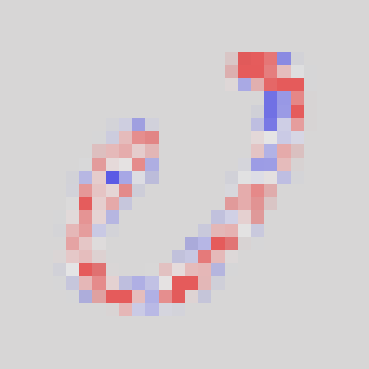}
    \centering\includegraphics[width=1.0cm]{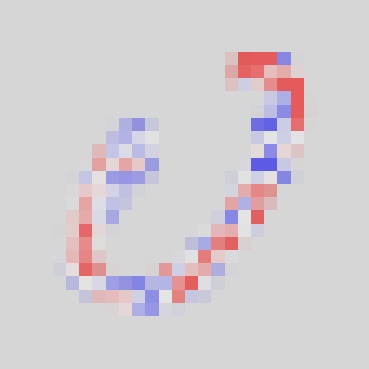}
    \centering\includegraphics[width=1.0cm]{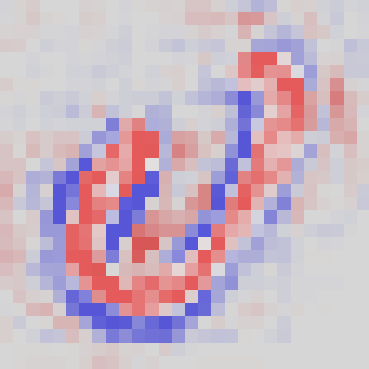}

    \centering\includegraphics[width=1.0cm]{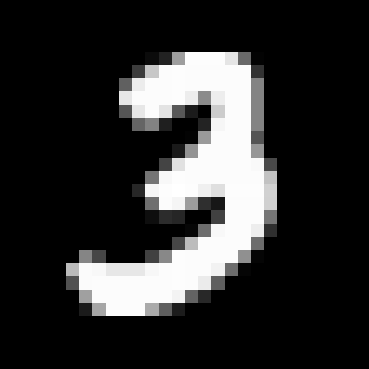}
    \centering\includegraphics[width=1.0cm]{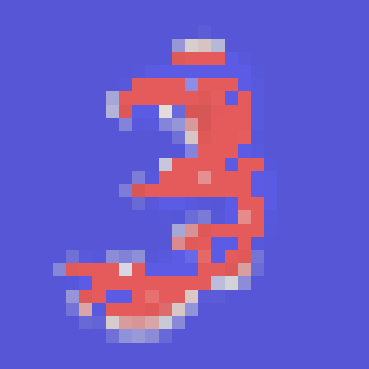}
    \centering\includegraphics[width=1.0cm]{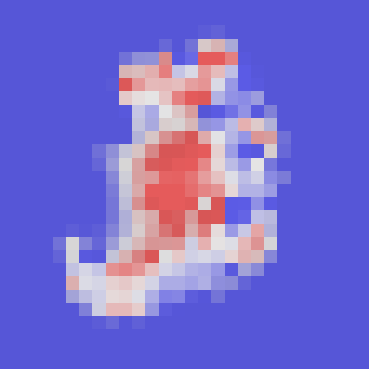}
    \centering\includegraphics[width=1.0cm]{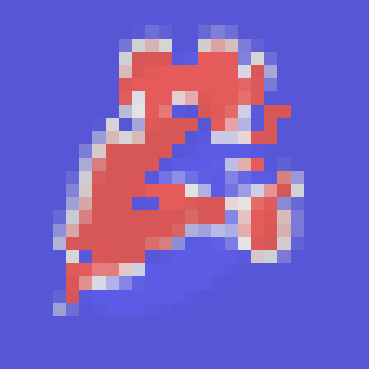}
    \centering\includegraphics[width=1.0cm]{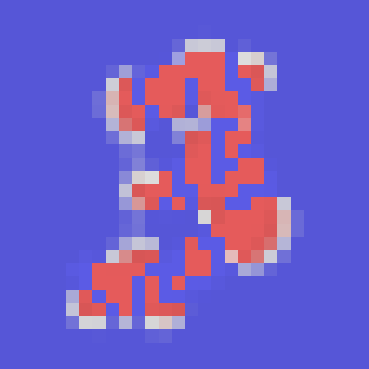}
    \centering\includegraphics[width=1.0cm]{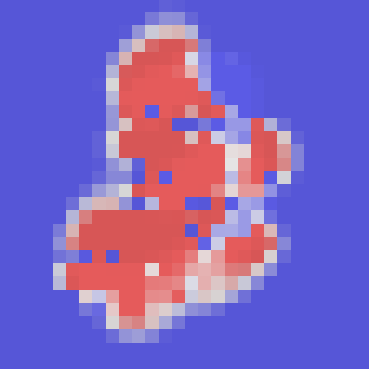}
    \centering\includegraphics[width=1.0cm]{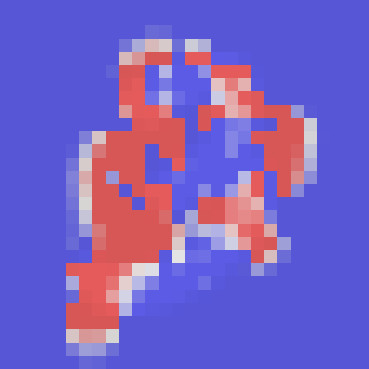}
    \centering\includegraphics[width=1.0cm]{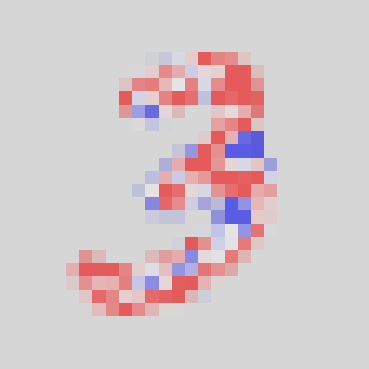}
    \centering\includegraphics[width=1.0cm]{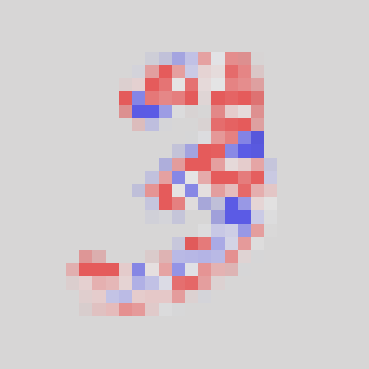}
    \centering\includegraphics[width=1.0cm]{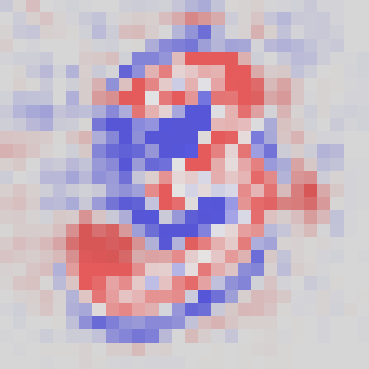}
    
    \centering\includegraphics[width=1.0cm]{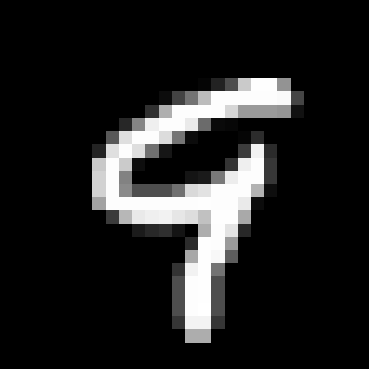}
    \centering\includegraphics[width=1.0cm]{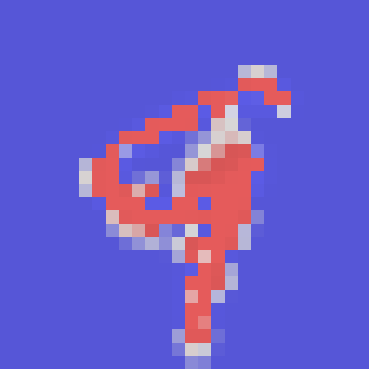}
    \centering\includegraphics[width=1.0cm]{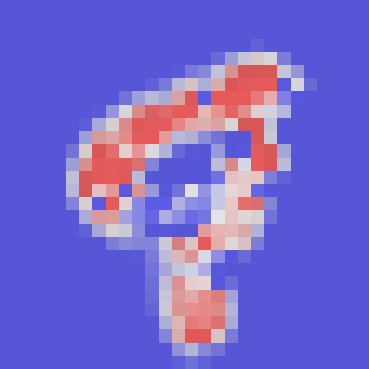}
    \centering\includegraphics[width=1.0cm]{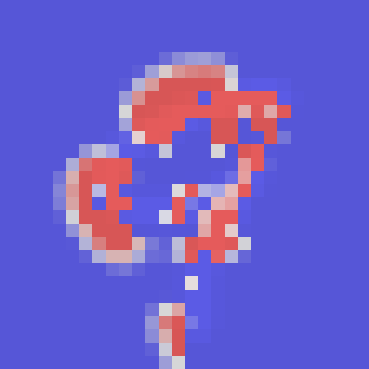}
    \centering\includegraphics[width=1.0cm]{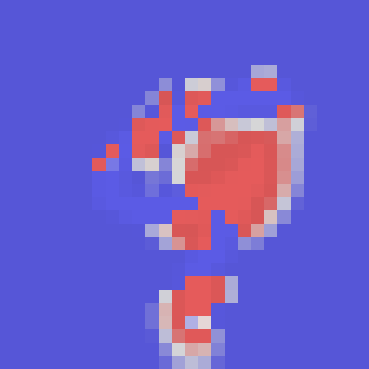}
    \centering\includegraphics[width=1.0cm]{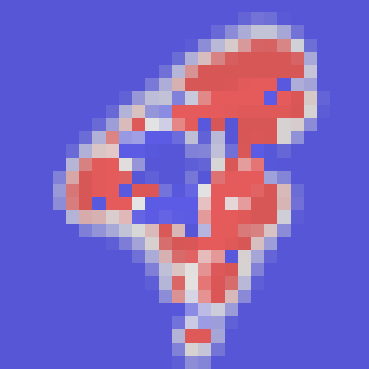}
    \centering\includegraphics[width=1.0cm]{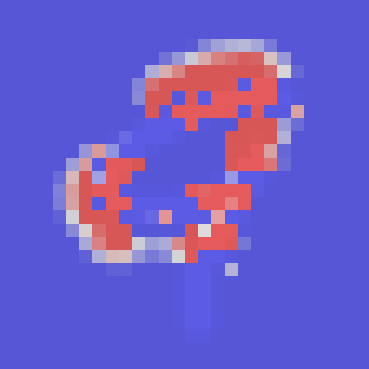}
    \centering\includegraphics[width=1.0cm]{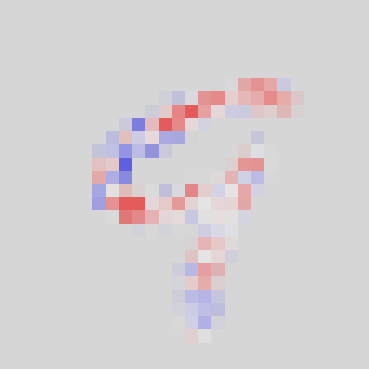}
    \centering\includegraphics[width=1.0cm]{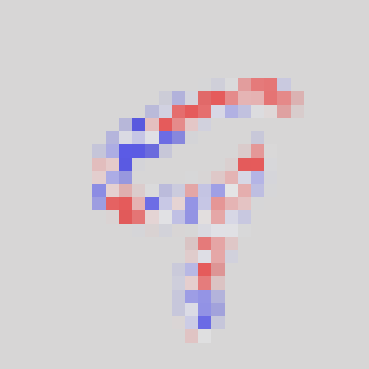}
    \centering\includegraphics[width=1.0cm]{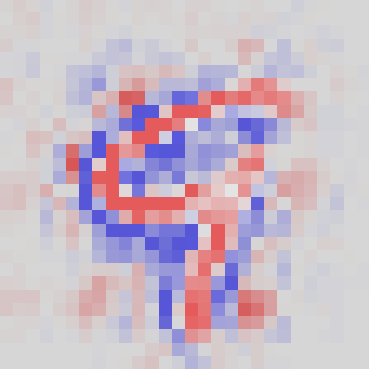}

    \centering\includegraphics[width=1.0cm]{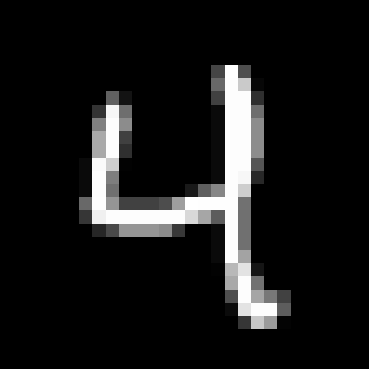}
    \centering\includegraphics[width=1.0cm]{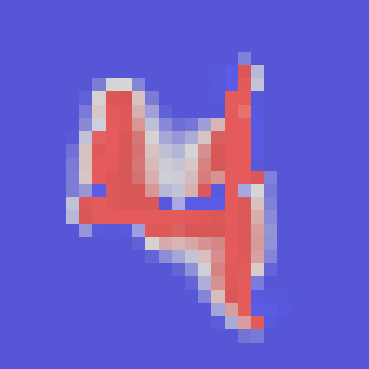}
    \centering\includegraphics[width=1.0cm]{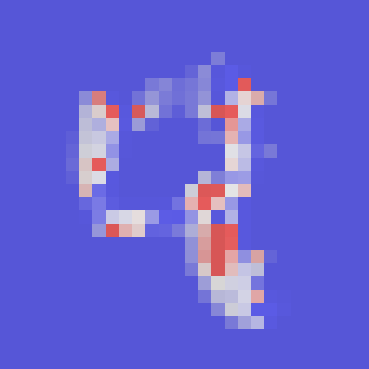}
    \centering\includegraphics[width=1.0cm]{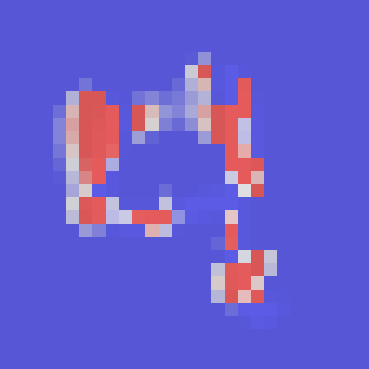}
    \centering\includegraphics[width=1.0cm]{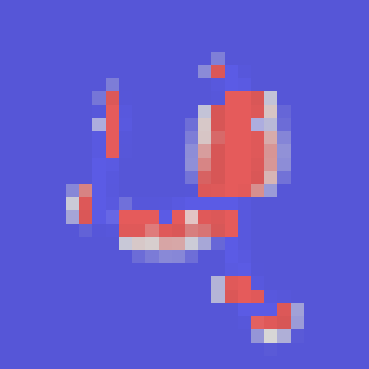}
    \centering\includegraphics[width=1.0cm]{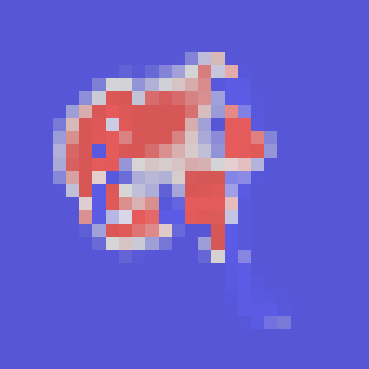}
    \centering\includegraphics[width=1.0cm]{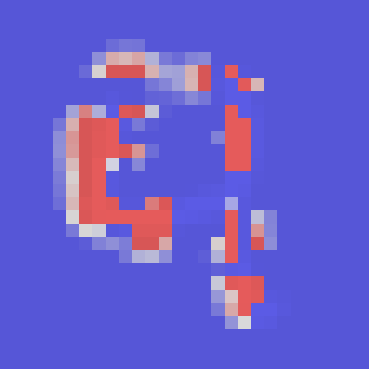}
    \centering\includegraphics[width=1.0cm]{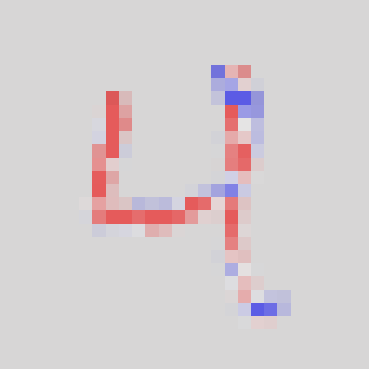}
    \centering\includegraphics[width=1.0cm]{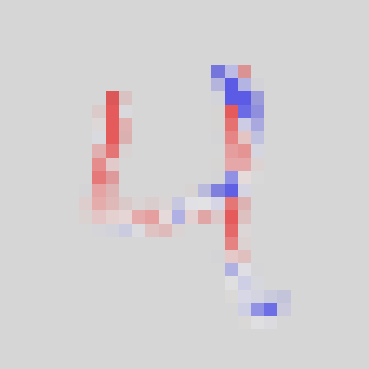}
    \centering\includegraphics[width=1.0cm]{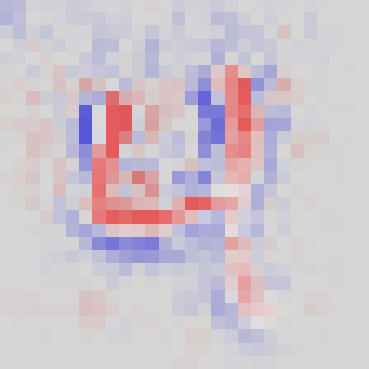}

    \caption{Examples of counterfactuals on MNIST. Left to right: original image, SSR flip, SSR VAE, SSR knockoff, SDR flip, SDR VAE, SDR knockoff, Integrated Gradients, Gradients $\times$ Input, Guided GradCAM. Flip and gradient-based methods highlight almost the entire number. VAE and knockoff in-filling methods focus on similar regions. However, knockoff counterfactuals highlight more areas outside the numbers that would transform each of them into another number. The generated saliency maps have different colors for the background pixels due to the different mapping of the attribution values. 
}
    \label{fig:mnist_maps}
\end{figure}

We generate counterfactuals to explain the classifier model with an accuracy score of 96.650\% on 2966 correctly classified samples with both SSR and SDR objectives. We visualize counterfactual explanations as divergent saliency maps by mapping the counterfactual ones to -0.5 to 0.5 and the gradient-based attributions to -1 and 1. The generated explanations for four selected samples can be found in \ref{fig:mnist_maps}.

To compute the metrics, we use three different threshold levels for the perturbation-based saliency maps: 0.4, 0.5, and 0.6.  For the gradient-based maps, we use  the following thresholds: the mean of the saliency map, 0.0, and 0.2 to account for the different scaling between the maps.  Since the MNIST dataset does not have ground-truth boxes, we threshold the images at the value 0.1 and take the smallest bounding box that encompasses the remaining pixels. We use these ground truth boxes to also compute a maximum baseline value, which we call MAX. We also compute a minimum baseline by defining the saliency box as the entire image.

Visual inspection of the results, some of which can be seen in Figure \ref{fig:mnist_maps}, reveal that the SSR flip counterfactuals and gradient-based methods generally highlight the entire number. In contrast, VAE counterfactuals attribute importance to fewer pixels within the number. Both VAE and knockoff counterfactuals highlight similar regions, and these are often the curves and angles that are unique to the number classes. Compared to VAE counterfactuals, knockoff counterfactuals also highlight regions that lie outside the numbers and could be modified to change the class of the number, as can be seen in the examples from Figure \ref{fig:mnist_maps}. VAE counterfactuals do not signalize these pixels due to the powerful regularization effect of the in-filling function. 

The quantitative results shown in Table \ref{tab:mnist_seed_res} support the qualitative observations. All counterfactual saliency maps give good results, with high WSL and negative SM values. The flip counterfactuals consistently give the best WSL values but fluctuate around the 0.0 value for the SM because they highlight almost the entire number. Among the gradient methods, Integrated Gradients (IG) and Input $\times$ Gradients (I $\times$ G) perform similarly well. Guided GradCAM (GGC) performs poorly, as it attributes pixels across the entire image.

As expected, the VAE and knockoff counterfactuals have a good SM score and a lower WSL score. Averaged over all classes and both objectives, knockoff counterfactuals obtain the best SM score. While this might indicate that they define the most compact relevant regions, the per-class evaluation presented in the Supplementary Material reveals additional insights: knockoff counterfactuals consistently perform better for classes 0, 3, and 8. VAE counterfactuals perform significantly worse in these classes, which explains the average outcome. For the other classes, VAE counterfactuals generally perform slightly better.  

A possible explanation for this outcome is that, for classes 0, 3, and 8, the generated knockoffs also belong to these same three classes with high probability. When used to corrupt the original image, the knockoffs displace critical lines such as the middle line of 0 or the lower and upper left lines of the 8. This displacement is always within the number and thus leads to a compact counterfactual. For the other classes, the generated knockoffs belong to classes that result in a displacement of lines further outside the object. Since the VAE rarely generates different values than the originals for the background pixels, the VAE counterfactuals do not highlight these regions. Another reason for failure to deliver explanations that indicate which regions should be changed is that some knockoffs are prototypes of the number. In those cases, the explanation does not indicate which regions should be modified and might stretch along with the entire number.  

Another important insight is that knockoffs also offer the most visually consistent results across both objectives. This is probably due to the reduced degrees of freedom of the knockoffs compared to the VAE approach: For the knockoff counterfactuals, we only use one knockoff image to obtain reference values, while  the VAE infers missing values at each iteration of the procedure. 

\begin{table*}[t]
\centering
\caption{Results of the weakly supervised location (WSL) and saliency metric (SM) for perturbation-based and gradient-based attributions. The WSL score is in \% (higher is better). For the SM, lower is better, and a negative value indicates a good saliency detector. Best scores are in bold.}
\resizebox{\textwidth}{!}{
\begin{tabular}{|l|l|l|l|l|l|l|l|l|l|l|l|}
\hline
\multirow{2}{*}{} & \multicolumn{7}{l|}{\textbf{Perturbation-based attribution}} & \multicolumn{4}{l|}{\textbf{Gradient-based attribution}} \\ \cline{2-12} 
 & \multirow{2}{*}{\textbf{Threshold}} & \multicolumn{3}{l|}{\textbf{SSR}} & \multicolumn{3}{l|}{\textbf{SDR}} & \multirow{2}{*}{\textbf{Threshold}} & \multirow{2}{*}{\textbf{IG}} &\multirow{2}{*}{\textbf{GGC}}&\multirow{2}{*}{\textbf{I $\times$ G}} \\ \cline{3-8}
\textbf{Metric} &  & \textbf{Flip} & \textbf{VAE} & \textbf{Knockoff} & \textbf{Flip} & \textbf{VAE} & \textbf{Knockoff} &  & &  &  \\ \hline
\multirow{3}{*}{\textbf{SM}} & 0.4 & -0.266 & \textbf{-0.440} & -0.418 & -0.319 & -0.385 & \textbf{-0.411} & \begin{tabular}[c]{@{}l@{}}saliency map \\ mean\end{tabular} & 0.055 & 0.050 & \textbf{0.002} \\ \cline{2-12} 
 & 0.5 & -0.228 & -0.334 & \textbf{-0.391} & -0.349 & -0.374 & \textbf{-0.398} & 0.0 & \textbf{-0.082} & -0.074 & 0.019 \\ \cline{2-12} 
 & 0.6 & -0.154 & -0.167 & \textbf{-0.363} & -0.150 & -0.354 & \textbf{-0.383} & 0.2 & 0.962 & 1.091 & \textbf{-0.037} \\ \hline
\multirow{3}{*}{\textbf{WSL}} & 0.4 & \textbf{99.804} & 98.335 & 97.970 & \textbf{99.906} & 97.390 & 96.723 & \begin{tabular}[c]{@{}l@{}}saliency map \\ mean\end{tabular} & \textbf{89.979} & 82.097 & 13.587 \\ \cline{2-12} 
 & 0.5 & \textbf{99.852} & 99.292 & 98.463 & \textbf{99.926} & 98.254 & 97.465 & 0.0 & \textbf{99.966} & 99.932 & 11.631 \\ \cline{2-12} 
 & 0.6 & \textbf{99.831} & 99.494 & 98.780 & \textbf{99.946} & 98.968 & 97.970 & 0.2 & \textbf{51.876} & 40.708 & 42.118 \\ \hline
\end{tabular}
}
\label{tab:mnist_seed_res}
\end{table*}

% -------------------------------------------------------------------
\subsection{Counterfactual explanations on complementary MNIST}
\label{sec:complementary_mnist}

\begin{figure}[H]
    \centering\includegraphics[width=1.0cm]{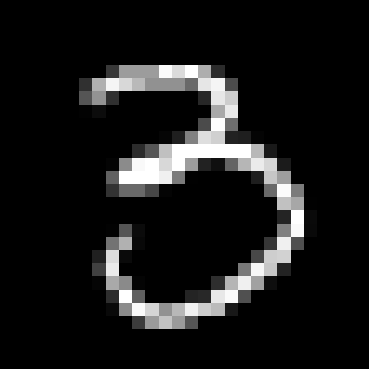}
    \centering\includegraphics[width=1.0cm]{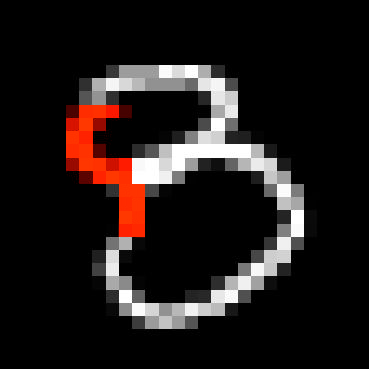}
    \centering\includegraphics[width=1.0cm]{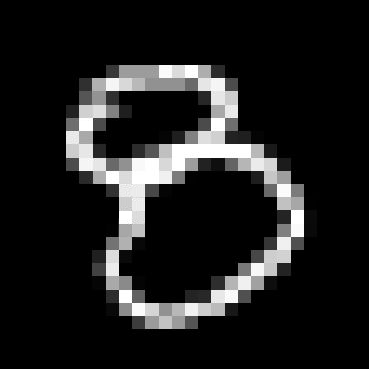}
    \centering\includegraphics[width=1.0cm]{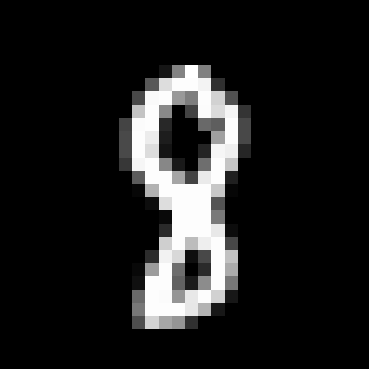}
    \centering\includegraphics[width=1.0cm]{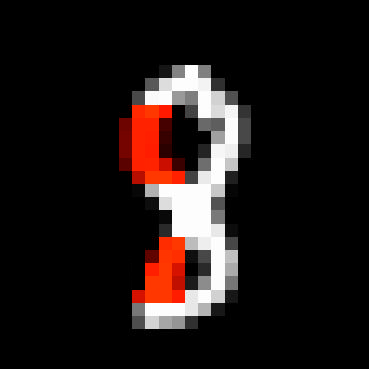}
    \centering\includegraphics[width=1.0cm]{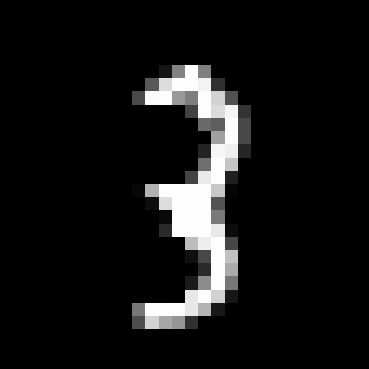}
    \caption{Examples of counterfactuals that result from the perturbation of numbers 3 and 8. By adding, in the case of 3, or removing, in the case of 8, the pixels highlighted in red, the numbers can be transformed into the other.}
    \label{fig:38_counterfactual}
\end{figure}

To better understand the ability of our in-filling method to generate counterfactuals in the causal sense, i.e., indicate which regions should be modified to change one number into the other, we derive two subsets from MNIST and generate counterfactuals for them. We select the pairs of numbers 3 and 8 and 5 and 6, consisting of 11982 training and 1984 testing images, and 11339 training and 1850 testing images, respectively. We name these subsets complementary MNIST 3/8 and 5/6. We consider these pairs of numbers complementary because they can easily be transformed into another by removing or adding parts, as highlighted for 3 and 8 in Figure \ref{fig:38_counterfactual}. Therefore, the evaluation of these datasets should better indicate which in-filling method displays the best counterfactual explanation.

Our trained classifiers obtain the following accuracy scores: 96.580\% on complementary MNIST 3/8 scores, and 90.140\% on complementary MNIST 5/6. We use a learning rate of $\lambda=5-e^4$ for the perturbation-based counterfactual generation, while all other parameters remain unchanged.

\paragraph{Counterfactual explanations  for complementary MNIST 3/8}

\begin{figure}
\centering
\begin{minipage}[t]{.48\textwidth}
  \centering
   \includegraphics[width=1.0cm]{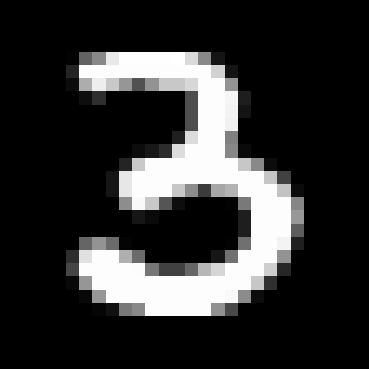}
   \includegraphics[width=1.0cm]{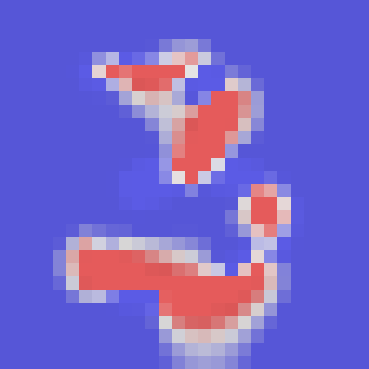}
   \includegraphics[width=1.0cm]{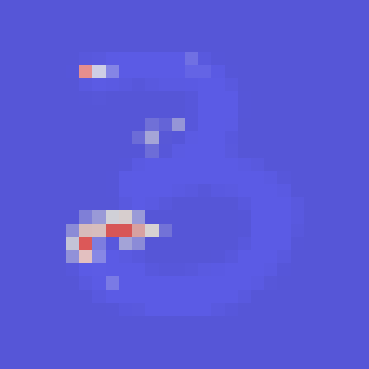}
   \includegraphics[width=1.0cm]{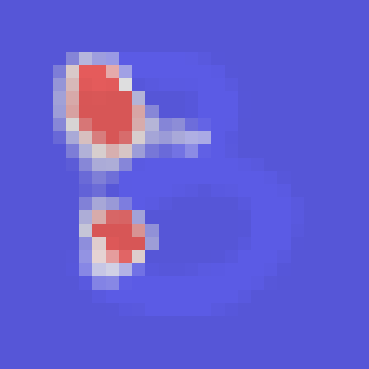}
   \includegraphics[width=1.0cm]{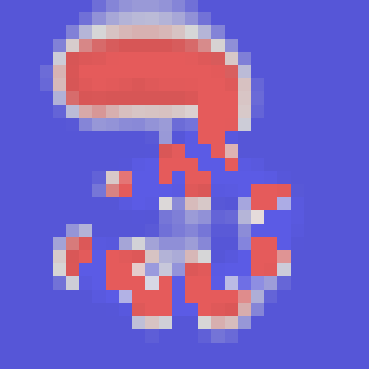}
   \includegraphics[width=1.0cm]{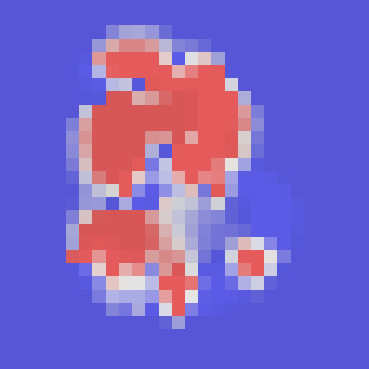}
   \includegraphics[width=1.0cm]{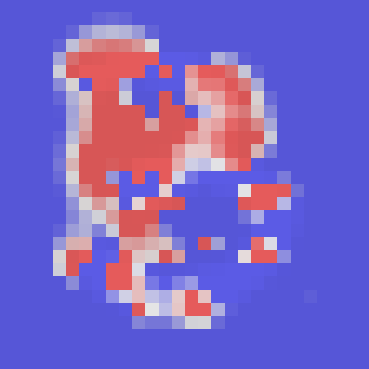}
   
   \includegraphics[width=1.0cm]{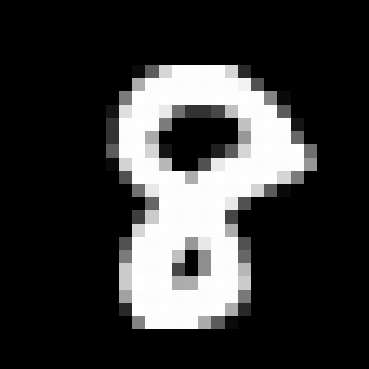}
   \includegraphics[width=1.0cm]{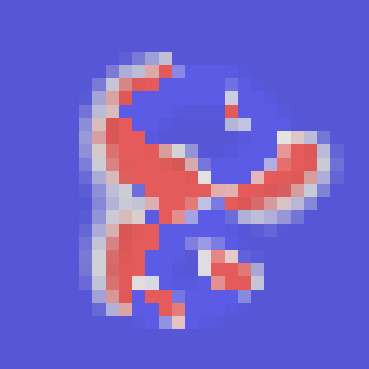}
   \includegraphics[width=1.0cm]{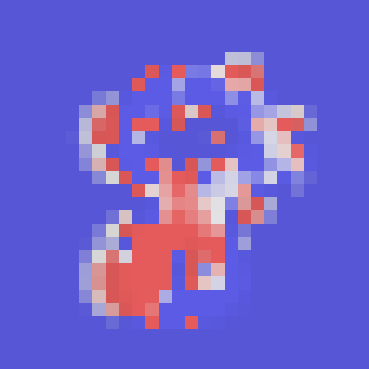}
   \includegraphics[width=1.0cm]{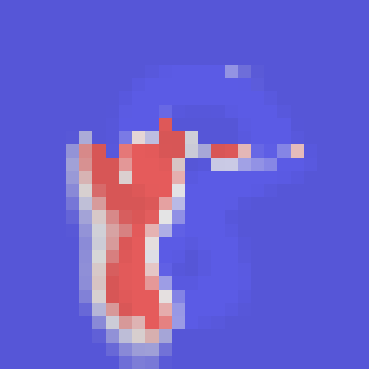}
   \includegraphics[width=1.0cm]{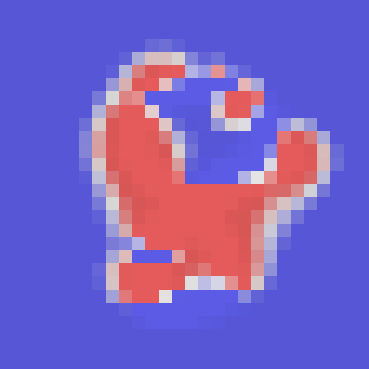}
   \includegraphics[width=1.0cm]{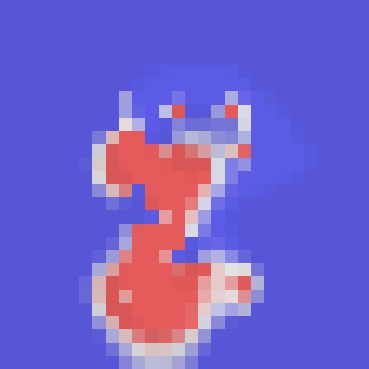}
   \includegraphics[width=1.0cm]{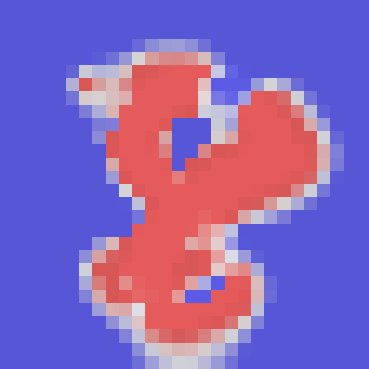}
   
   \includegraphics[width=1.0cm]{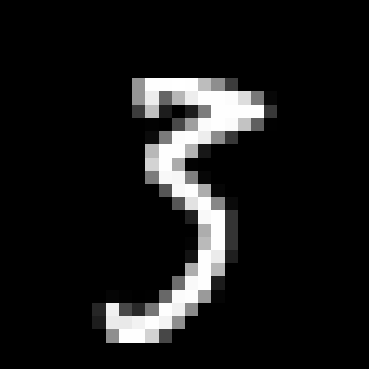}
   \includegraphics[width=1.0cm]{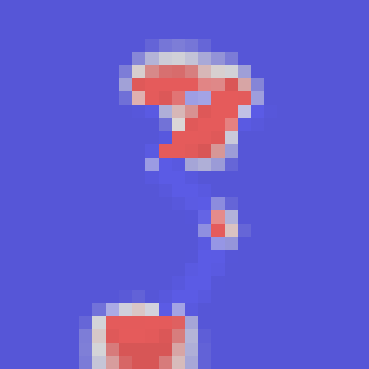}
   \includegraphics[width=1.0cm]{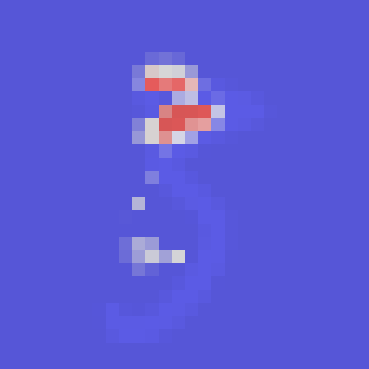}
   \includegraphics[width=1.0cm]{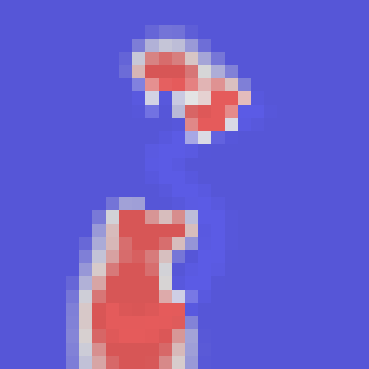}
   \includegraphics[width=1.0cm]{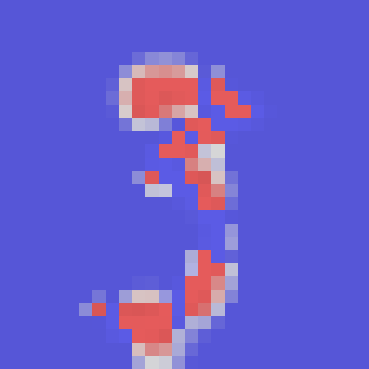}
   \includegraphics[width=1.0cm]{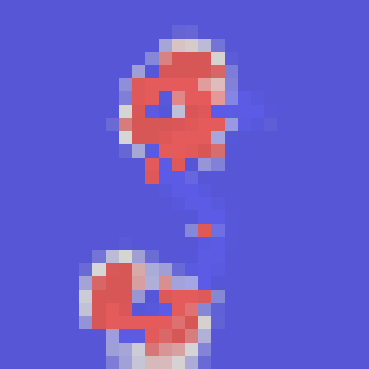}
   \includegraphics[width=1.0cm]{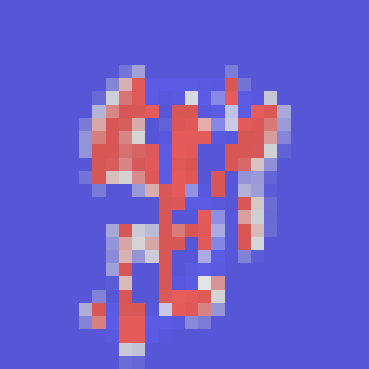}
   
   \includegraphics[width=1.0cm]{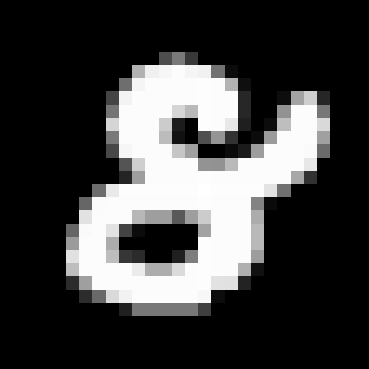}
   \includegraphics[width=1.0cm]{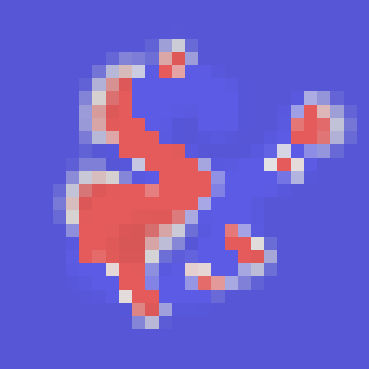}
   \includegraphics[width=1.0cm]{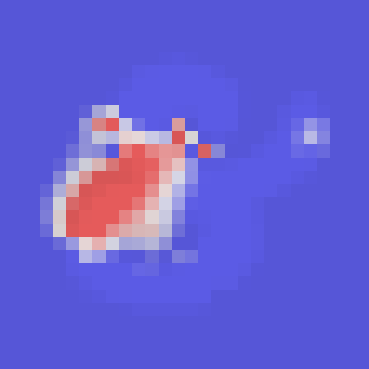}
   \includegraphics[width=1.0cm]{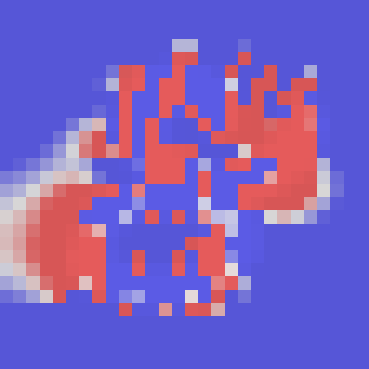}
   \includegraphics[width=1.0cm]{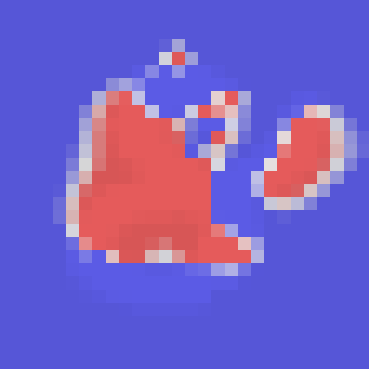}
   \includegraphics[width=1.0cm]{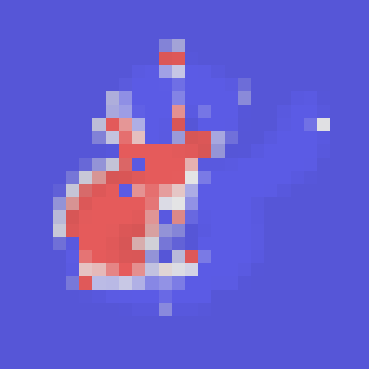}
   \includegraphics[width=1.0cm]{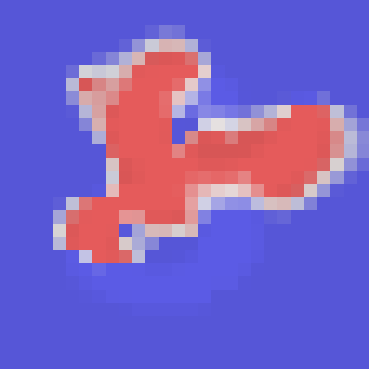}

    \caption{Counterfactuals for complementary MNIST 3/8. Left to right: original image, SSR flip, SSR VAE, SSR knockoff, SDR flip, SDR VAE, SDR knockoff. Upper two rows: samples for which SSR knockoff counterfactuals perform well compared to VAE counterfactuals. Lower two rows: samples for which SSR knockoff counterfactuals perform worse than VAE counterfactuals.}
    \label{fig:38}
\end{minipage} \hfill %
\begin{minipage}[t]{.48\textwidth}
    \centering
    \includegraphics[width=1.0cm]{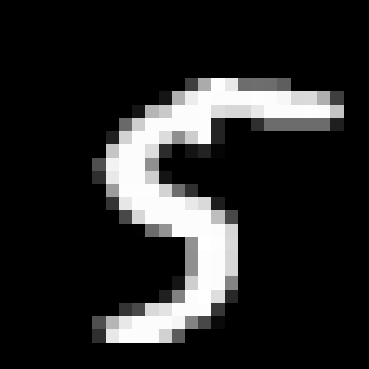}
    \includegraphics[width=1.0cm]{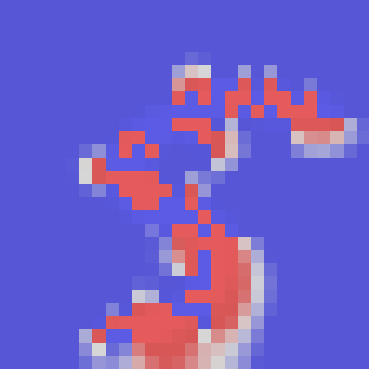}
    \includegraphics[width=1.0cm]{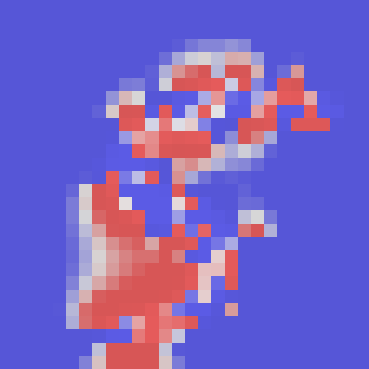}
    \includegraphics[width=1.0cm]{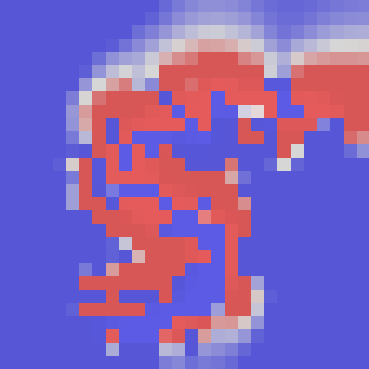}
    \includegraphics[width=1.0cm]{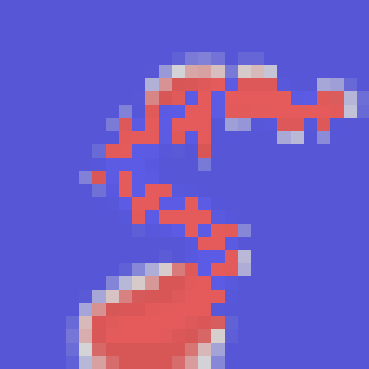}
    \includegraphics[width=1.0cm]{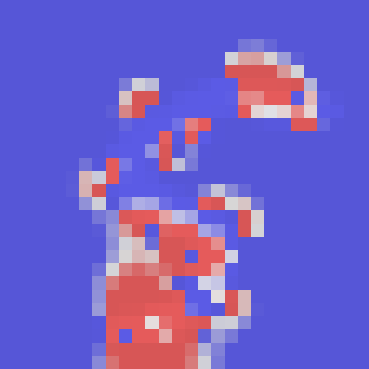}
    \includegraphics[width=1.0cm]{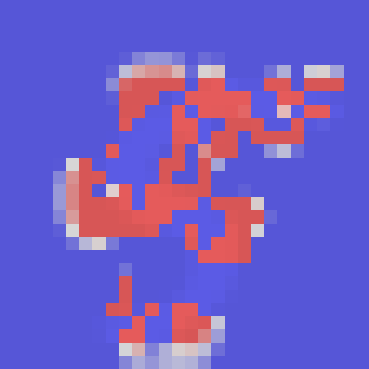}
   
    \includegraphics[width=1.0cm]{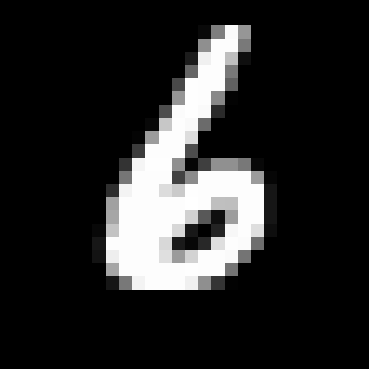}
    \includegraphics[width=1.0cm]{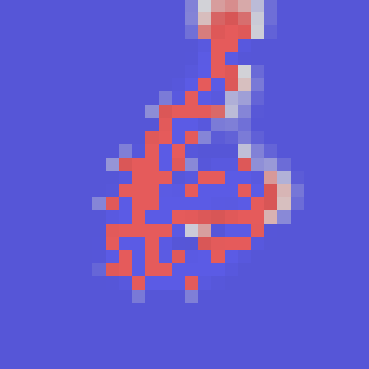}
    \includegraphics[width=1.0cm]{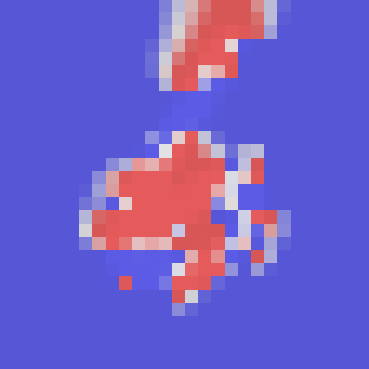}
    \includegraphics[width=1.0cm]{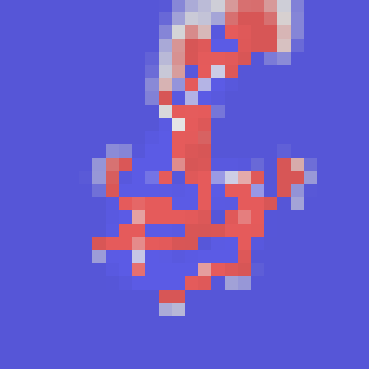}
    \includegraphics[width=1.0cm]{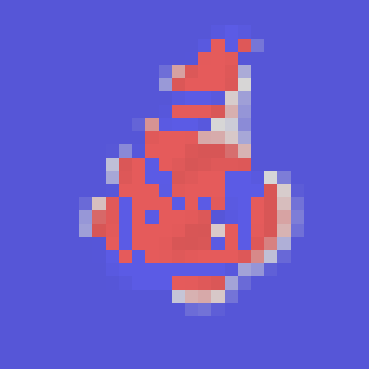}
    \includegraphics[width=1.0cm]{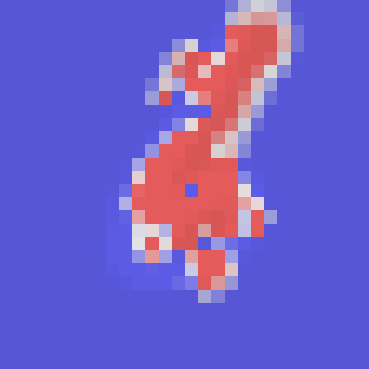}
    \includegraphics[width=1.0cm]{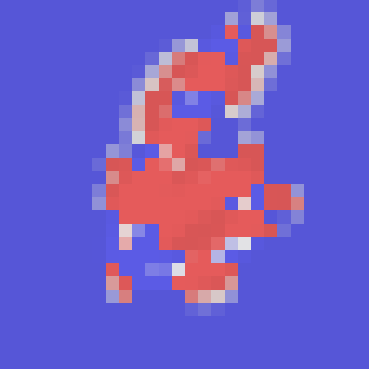}
    
    \includegraphics[width=1.0cm]{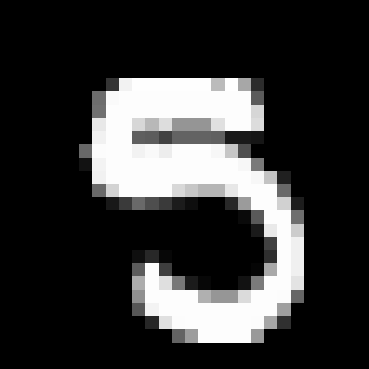}
    \includegraphics[width=1.0cm]{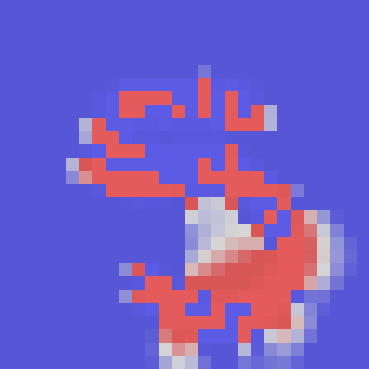}
    \includegraphics[width=1.0cm]{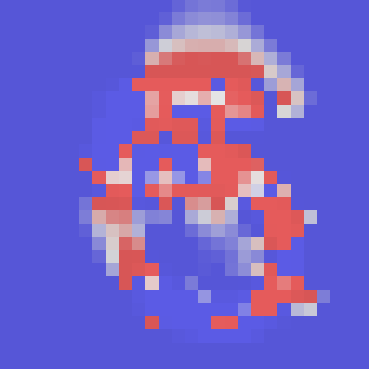}
    \includegraphics[width=1.0cm]{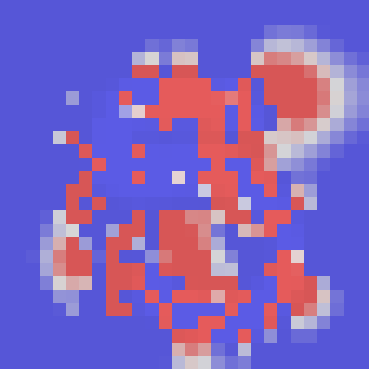}
    \includegraphics[width=1.0cm]{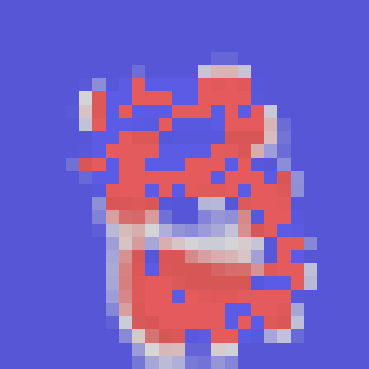}
    \includegraphics[width=1.0cm]{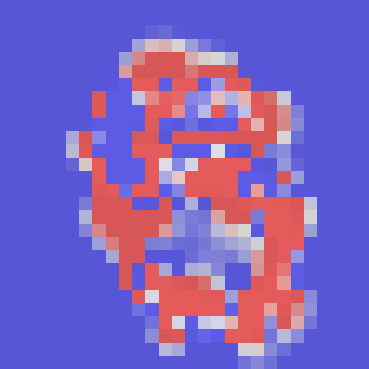}
    \includegraphics[width=1.0cm]{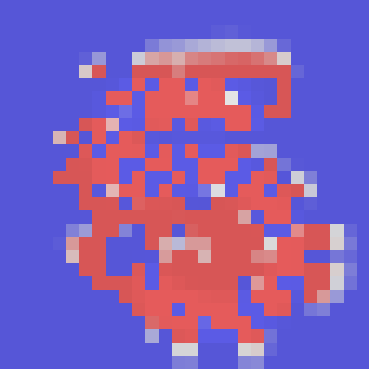}

    \includegraphics[width=1.0cm]{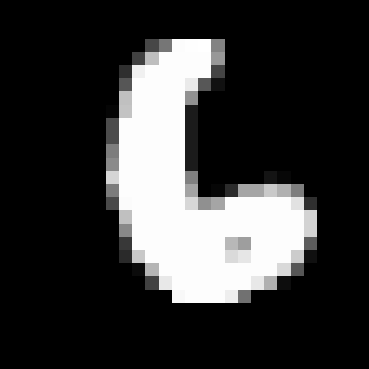}
    \includegraphics[width=1.0cm]{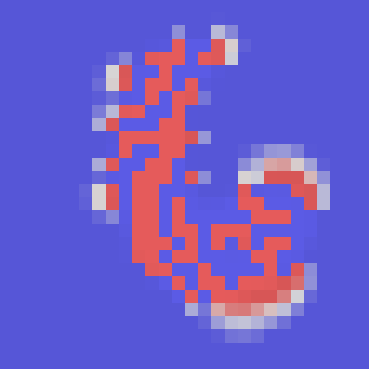}
    \includegraphics[width=1.0cm]{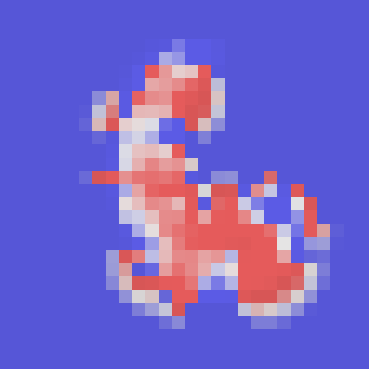}
    \includegraphics[width=1.0cm]{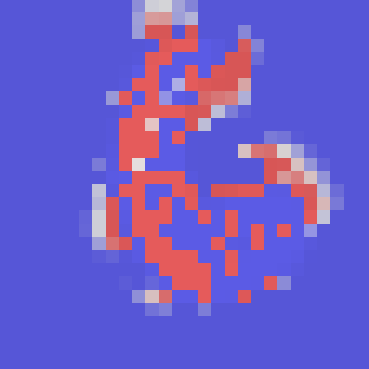}
    \includegraphics[width=1.0cm]{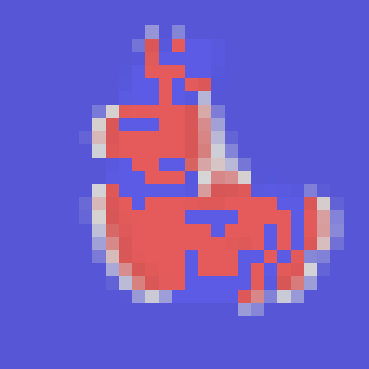}
    \includegraphics[width=1.0cm]{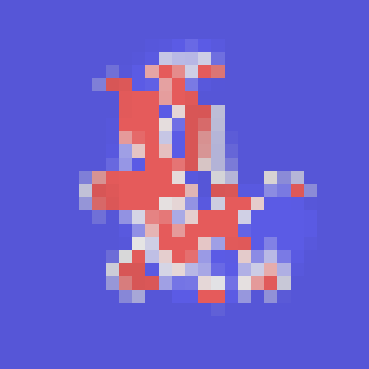}
    \includegraphics[width=1.0cm]{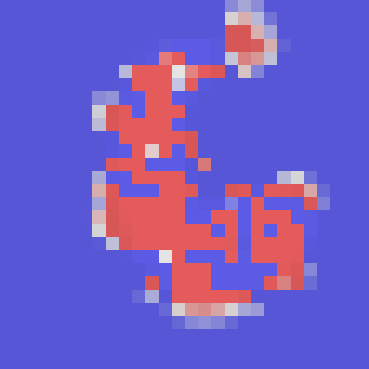}

    \caption{Counterfactuals for complementary MNIST 5/6. Left to right: original image, SSR flip, SSR VAE, SSR knockoff, SDR flip, SDR VAE, SDR knockoff. Upper two rows: samples for which SSR knockoff counterfactuals on the complementary MNIST 5/6 perform well compared to VAE counterfactuals. Lower two rows: samples for which SSR knockoff counterfactuals perform worse than VAE counterfactuals.}
    \label{fig:56}
\end{minipage}
\end{figure}

Since this subset contains only two classes, the flip in-filling generates better counterfactuals than for MNIST but still highlights pixels that cover a large surface of the number, as shown in the second row of Figure \ref{fig:38}. On the other extreme, VAE counterfactuals often select few pixels at the numbers' ends or curves, rather than the entire regions. Knockoff counterfactuals highlight whole areas that should be removed or added to a number to transform them into the other but generally restrict to these regions only. These observations hold only for the SSR objective. For the SDR objective, all counterfactuals cover large parts of the numbers, probably due to the higher degrees of freedom. It is also important to note that the knockoff in-filling sometimes fails. This happens when the generated knockoff is a prototype of the class, rather than a complementary number, as shown in the last row of Figure \ref{fig:38}. 

\paragraph{Counterfactual explanations for complementary MNIST 5/6}

For the complementary MNIST 5/6, flip SSR counterfactuals signalize the whole number as important. VAE counterfactuals signalize more compact regions that should be modified to transform one number into the other, as can be seen in Figure \ref{fig:56}. Knockoff counterfactuals highlight similar areas to the VAE yet tend to attribute more pixels. For example, for the numbers 5 in Figure \ref{fig:56}, importance is attributed to the upper and lower left parts of the numbers. However, some artifacts can be observed in the upper central parts of the numbers. Bad performing saliency maps, such as the last two rows of Figure \ref{fig:56}, reveal the downside of reduced regularization that comes with knockoffs. It leads to artifacts in the knockoff counterfactuals that are difficult to interpret. Although these maps still signalize some important regions, they are much noisier compared to the VAE maps. This might be again because, for the 5/6 dataset, the generated knockoffs are more often prototypes of the class number.

As for the 3/8 subset, SDR counterfactuals generally signalize large areas of the number. Knockoffs and VAE SDR counterfactuals do focus on similar regions, but knockoff counterfactuals tend to signalize more pixels than the VAE ones.

% \subsection{Discussion}

% Our results showed that on average, knockoff counterfactuals score best with respect to SM. However, class-wise inspection shows that knockoff counterfactuals consistently score best for three classes: 0, 3, and 8, for which the VAE counterfactuals perform significantly worse. One reason for this is that the knockoffs for these classes still belong to one of the classes. Thus, the resulting images have displaced pixels that lie within the number and result in compact explanation. For other classes, the altered pixels lie more outside the number because the knockoff is not complementary in shape. Generally, knockoff counterfactuals highlight entire regions to be perturbed to change the number into another, while VAE counterfactuals mark as relevant key pixels of the number. Since knockoffs respect the in-distributioni-ness condition, we believe that these pixels are indeed important to the classifier. Furthermore, the consistency of the explanation might better fulfill expectations from causal and human perspectives. 

\section{Conclusions}

Counterfactual explanations are highly influenced by the reference values used to perturb the features. Heuristically generated values falsely attribute pixels due to the out-of-distribution data after perturbation, which biases the classifier, and, implicitly, the explanation \cite{chang18}. Perturbations with generative in-fillings alleviate this problem since they are in-distribution. They also regularize counterfactual explanations since generative models predict redundant pixels well and thus introduce no change for these pixels. However, generative methods may also over-regularize since they can also perform well on less redundant pixels. We hypothesized that this might lead to an increased number of false negatives and showed that this is indeed the case.

To alleviate the over-regularization, we proposed to use the Knockoffs framework of \cite{Barber_2015} to generate perturbation values. We compared our developed counterfactual explanation method with heuristic and generative in-fillings and gradient-based methods for CNNs classifiers. Here we could demonstrate that knockoff counterfactuals highlight the entire region to be perturbed to change the number into another, while VAE counterfactuals highlight the number's relevant key pixels. On average, knockoff counterfactuals score best with respect to the compactness of the salient area. Yet, a class-wise inspection shows that knockoff counterfactuals consistently score best for the classes for which perturbed pixels are still within the number. When knockoffs are not complementary in shape with the original, pixels signalized as important lie outside the number. However, these indicate which regions could be modified to change the classifier output. Furthermore, since knockoffs respect the in-distribution-ness condition, we believe that these pixels are important to the classifier.

Although we obtained promising results in our proof-of-concept study, our current approach needs further refinement. First, an extensive evaluation of knockoff generation methods for image data is necessary. This step would also allow us to test our method on natural image data. We also faced the question of whether evaluating the regions' compactness is the right approach for counterfactual evaluation. A consistent explanation might better fulfill expectations from causal and human perspective. We believe that for the future, taking a more causal approach towards evaluation metrics would be interesting and could create a bridge between human and machine reasoning.

\bibliographystyle{unsrt}
\bibliography{references}

\begin{thebibliography}{10}

\bibitem{ancona17}
Marco Ancona, Enea Ceolini, Cengiz Öztireli, and Markus Gross.
\newblock A unified view of gradient-based attribution methods for deep neural
  networks.
\newblock 11 2017.

\bibitem{chang18}
Chun{-}Hao Chang, Elliot Creager, Anna Goldenberg, and David Duvenaud.
\newblock Explaining image classifiers by adaptive dropout and generative
  in-filling.
\newblock {\em CoRR}, abs/1807.08024, 2018.

\bibitem{Barber_2015}
Rina~Foygel Barber and Emmanuel~J. Candès.
\newblock Controlling the false discovery rate via knockoffs.
\newblock {\em The Annals of Statistics}, 43(5):2055–2085, Oct 2015.

\bibitem{candes16}
Emmanuel Candes, Yingying Fan, Lucas Janson, and Jinchi Lv.
\newblock Panning for gold: Model-x knockoffs for high-dimensional controlled
  variable selection, 2016.

\bibitem{watson2019testing}
David~S. Watson and Marvin~N. Wright.
\newblock Testing conditional independence in supervised learning algorithms,
  2019.

\bibitem{mnist10}
Yann LeCun and Corinna Cortes.
\newblock {MNIST} handwritten digit database.
\newblock 2010.

\bibitem{Ancona2019GradientBasedAM}
Marco~B Ancona, Enea Ceolini, Cengiz {\"O}ztireli, and Markus~H. Gross.
\newblock Gradient-based attribution methods.
\newblock In {\em Explainable AI}, 2019.

\bibitem{dabkowski17}
Piotr Dabkowski and Yarin Gal.
\newblock Real time image saliency for black box classifiers, 2017.

\bibitem{fong17}
Ruth Fong and Andrea Vedaldi.
\newblock Interpretable explanations of black boxes by meaningful perturbation.
\newblock {\em CoRR}, abs/1704.03296, 2017.

\bibitem{zeiler13}
Matthew~D. Zeiler and Rob Fergus.
\newblock Visualizing and understanding convolutional networks.
\newblock {\em CoRR}, abs/1311.2901, 2013.

\bibitem{kingma13}
Diederik~P Kingma and Max Welling.
\newblock Auto-encoding variational bayes, 2013.

\bibitem{gan}
Ian Goodfellow, Jean Pouget-Abadie, Mehdi Mirza, Bing Xu, David Warde-Farley,
  Sherjil Ozair, Aaron Courville, and Yoshua Bengio.
\newblock Generative adversarial nets.
\newblock In Z.~Ghahramani, M.~Welling, C.~Cortes, N.~Lawrence, and K.~Q.
  Weinberger, editors, {\em Advances in Neural Information Processing Systems},
  volume~27, pages 2672--2680. Curran Associates, Inc., 2014.

\bibitem{sundararajan17}
Mukund Sundararajan, Ankur Taly, and Qiqi Yan.
\newblock Axiomatic attribution for deep networks.
\newblock {\em CoRR}, abs/1703.01365, 2017.

\bibitem{Binder2016LayerWiseRP}
Alexander Binder, Sebastian Bach, Gr{\'e}goire Montavon, Klaus-Robert
  M{\"u}ller, and Wojciech Samek.
\newblock Layer-wise relevance propagation for deep neural network
  architectures.
\newblock 2016.

\bibitem{DBLP:journals/corr/deeplift}
Avanti Shrikumar, Peyton Greenside, and Anshul Kundaje.
\newblock Learning important features through propagating activation
  differences.
\newblock {\em CoRR}, abs/1704.02685, 2017.

\bibitem{smoothgrad}
Daniel Smilkov, Nikhil Thorat, Been Kim, Fernanda~B. Vi{\'{e}}gas, and Martin
  Wattenberg.
\newblock Smoothgrad: removing noise by adding noise.
\newblock {\em CoRR}, abs/1706.03825, 2017.

\bibitem{ZhouKLOT15}
Bolei Zhou, Aditya Khosla, {\`{A}}gata Lapedriza, Aude Oliva, and Antonio
  Torralba.
\newblock Learning deep features for discriminative localization.
\newblock {\em CoRR}, abs/1512.04150, 2015.

\bibitem{grad_cam}
Ramprasaath~R. Selvaraju, Abhishek Das, Ramakrishna Vedantam, Michael Cogswell,
  Devi Parikh, and Dhruv Batra.
\newblock Grad-cam: Why did you say that? visual explanations from deep
  networks via gradient-based localization.
\newblock {\em CoRR}, abs/1610.02391, 2016.

\bibitem{Romano19}
Yaniv Romano, Matteo Sesia, and Emmanuel Candès.
\newblock Deep knockoffs.
\newblock {\em Journal of the American Statistical Association}, page 1–12,
  Oct 2019.

\bibitem{duarte2020knockoffinspired}
Marco~F. Duarte and Siwei Feng.
\newblock Knockoff-inspired feature selection via generative models, 2020.

\bibitem{gal2017concrete}
Yarin Gal, Jiri Hron, and Alex Kendall.
\newblock Concrete dropout, 2017.

\bibitem{msssim}
Z.~{Wang}, E.~P. {Simoncelli}, and A.~C. {Bovik}.
\newblock Multiscale structural similarity for image quality assessment.
\newblock In {\em The Thrity-Seventh Asilomar Conference on Signals, Systems
  Computers, 2003}, volume~2, pages 1398--1402 Vol.2, 2003.

\bibitem{kingma2014adam}
Diederik~P. Kingma and Jimmy Ba.
\newblock Adam: A method for stochastic optimization, 2014.

\end{thebibliography}

\newcommand{\beginsupplement}{%
        \setcounter{table}{0}
        \renewcommand{\thetable}{S\arabic{table}}%
        \setcounter{figure}{0}
        \renewcommand{\thefigure}{S\arabic{figure}}%
     }
     
\section{Supplementary Material}
\beginsupplement

\subsection{Class-wise evaluation}

\begin{table}[H]
\centering
\begin{minipage}[t]{.48\textwidth}
\centering
\caption{Results of the class-wise evaluation of the weakly supervised location (WSL) and saliency metric (SM) on MNIST counterfactuals with 0.4 threshold for the \textbf{SSR} objective. Flip scores best for class 1, VAE scores best for classes: 2, 4, 5, 6, 7, 9, and knockoff score best for classes: 0, 3, 8. Best scores are in bold.}
\resizebox{\textwidth}{!}{
\renewcommand{\arraystretch}{0.5}
\begin{tabular}{|l|l|l|l|l|l|l|}
\hline
 & \multicolumn{3}{l|}{\textbf{WSL}} & \multicolumn{3}{l|}{\textbf{SM}} \\ \hline
\textbf{Class} & \textbf{Flip} & \textbf{VAE} & \textbf{Knockoff} & \textbf{Flip} & \textbf{VAE} & \textbf{Knockoff} \\ \hline
\textbf{0} & \textbf{100.000} & \textbf{100.000} & 99.267 & 0.603 & 0.056 & \textbf{-0.164} \\ \hline
\textbf{1} & \textbf{98.237} & 85.593 & 85.897 & \textbf{-1.113} & -1.104 & -1.104 \\ \hline
\textbf{2} & \textbf{100.000} & \textbf{100.000} & 99.934 & -0.480 & \textbf{-0.525} & -0.279 \\ \hline
\textbf{3} & \textbf{100.000} & \textbf{100.000} & 99.781 & -0.333 & 0.072 & \textbf{-0.371} \\ \hline
\textbf{4} & \textbf{100.000} & \textbf{100.000} & 99.262 & -0.647 & \textbf{-0.743} & -0.532 \\ \hline
\textbf{5} & \textbf{100.000} & \textbf{100.000} & 100.000 & -0.633 & \textbf{-0.739} & -0.585 \\ \hline
\textbf{6} & \textbf{100.000} & 99.189 & 98.986 & -0.552 & \textbf{-0.683} & -0.474 \\ \hline
\textbf{7} & \textbf{100.000} & \textbf{100.000} & 99.744 & -0.487 & \textbf{-0.760} & -0.661 \\ \hline
\textbf{8} & \textbf{100.000} & \textbf{100.000} & 99.241 & 0.145 & 0.240 & \textbf{-0.010} \\ \hline
\textbf{9} & \textbf{100.000} & \textbf{100.000} & 99.419 & 0.942 & \textbf{-0.058} & 0.091 \\ \hline
\end{tabular}
}
\end{minipage} \hfill %
\begin{minipage}[t]{.48\textwidth}
\centering
\caption{Results of the class-wise evaluation of weakly supervised location (WSL) and saliency metric (SM) on MNIST counterfactuals with 0.4 threshold for the \textbf{SDR} objective. Flip scores best for classes: 1, 2, 4, 5, 6, VAE scores best for class 7, and knockoff scores best for classes: 0, 3, 8, 9. Best scores are in bold.}
\resizebox{\textwidth}{!}{
\renewcommand{\arraystretch}{0.5}
\begin{tabular}{|l|l|l|l|l|l|l|}
\hline
 & \multicolumn{3}{l|}{\textbf{WSL}} & \multicolumn{3}{l|}{\textbf{SM}} \\ \hline
\textbf{Class} & \textbf{Flip} & \textbf{VAE} & \textbf{Knockoff} & \textbf{Flip} & \textbf{VAE} & \textbf{Knockoff} \\ \hline
\textbf{0} & \textbf{100.000} & 99.414 & \textbf{100.000} & 0.345 & -0.007 & \textbf{{-0.019}} \\ \hline
\textbf{1} & \textbf{99.271} & 81.824 & 76.231 & \textbf{-1.136} & -1.067 & -1.069 \\ \hline
\textbf{2} & \textbf{100.000} & \textbf{100.000} & 99.934 & \textbf{-0.461} & -0.386 & -0.390 \\ \hline
\textbf{3} & \textbf{100.000} & 98.759 & 99.124 & -0.354 & -0.318 & \textbf{-0.366} \\ \hline
\textbf{4} & \textbf{100.000} & 99.530 & 99.597 & \textbf{-0.601} & -0.489 & -0.547 \\ \hline
\textbf{5} & \textbf{100.000} & \textbf{100.000} & 99.787 & \textbf{-0.615} & -0.563 & -0.544 \\ \hline
\textbf{6} & \textbf{100.000} & \textbf{100.000} & 98.378 & { \textbf{-0.616}} & -0.558 & -0.483 \\ \hline
\textbf{7} & \textbf{100.000} & 99.297 & 99.361 & -0.608 & { \textbf{-0.629}} & -0.553 \\ \hline
\textbf{8} & \textbf{100.000} & 99.379 & 99.241 & 0.283 & 0.152 & \textbf{{-0.057}} \\ \hline
\textbf{9} & 99.871 & 97.677 & 98.194 & 0.686 & 0.121 & \textbf{0.013} \\ \hline
\end{tabular}
}
\end{minipage}
\end{table}

\vspace{-2em}

\begin{table}[H]
\centering
\begin{minipage}[t]{.48\textwidth}
\centering
\caption{Results of the class-wise evaluation of weakly supervised location (WSL) and saliency metric (SM) on MNIST counterfactuals with 0.5 threshold for the \textbf{SSR} objective. Flip scores best for class: 2, VAE scores best for classes: 4, 5, 6, 7, 9 and knockoff scores best for classes: 0, 1, 3, 8. Best scores are in bold.}
\resizebox{\textwidth}{!}{
\renewcommand{\arraystretch}{0.5}
\begin{tabular}{|l|l|l|l|l|l|l|}
\hline
 & \multicolumn{3}{l|}{\textbf{WSL}} & \multicolumn{3}{l|}{\textbf{SM}} \\ \hline
\textbf{Class} & \textbf{Flip} & \textbf{VAE} & \textbf{Knockoff} & \textbf{Flip} & \textbf{VAE} & \textbf{Knockoff} \\ \hline
\textbf{0} & \textbf{100.000} & \textbf{100.000} & \textbf{100.000} & 1.005 & 0.449 & \textbf{{-0.097}}\\ \hline
\textbf{1} & \textbf{98.646} & 93.975 & 88.624 & -0.956 & -0.783 & \textbf{{-0.987}} \\ \hline
\textbf{2} & \textbf{100.000} & \textbf{100.000} & \textbf{100.000} & \textbf{{-0.556}} & {-0.544} & -0.296 \\ \hline
\textbf{3} & \textbf{100.000} & \textbf{100.000} & \textbf{100.000} & -0.210 & 0.739 & \textbf{{-0.340}} \\ \hline
\textbf{4} & \textbf{100.000} & \textbf{100.000} & 99.256 & -0.699 & \textbf{{-0.733}} & -0.562 \\ \hline
\textbf{5} & \textbf{100.000} & \textbf{100.000} & \textbf{100.000} & -0.669 & \textbf{{-0.769}} & -0.575 \\ \hline
\textbf{6} & \textbf{100.000} & 99.392 & 99.172 & -0.576 & \textbf{{-0.721}} & -0.510 \\ \hline
\textbf{7} & \textbf{100.000} & \textbf{100.000} & 99.920 & -0.554 & \textbf{{-0.779}} & -0.699 \\ \hline
\textbf{8} & \textbf{100.000} & \textbf{100.000} & 99.448 & 0.256 & 0.292 & \textbf{{0.027}}\\ \hline
\textbf{9} & \textbf{100.000} & \textbf{100.000} & 99.612 & 0.994 & \textbf{{0.143}} & {0.276} \\ \hline
\end{tabular}
}
\end{minipage} \hfill %
\begin{minipage}[t]{.48\textwidth}
\centering
\caption{Results of the class-wise evaluation of weakly supervised location (WSL) and saliency metric (SM) on MNIST counterfactuals with 0.5 threshold for the \textbf{SDR} objective. Flip scores best for classes: 1, 2, 4, 5, 6, VAE scores best for classes: 1, 7 and knockoff scores best for classes: 3, 8, 9. Best scores are in bold.}
\resizebox{\textwidth}{!}{
\renewcommand{\arraystretch}{0.5}
\begin{tabular}{|l|l|l|l|l|l|l|}
\hline
 & \multicolumn{3}{l|}{\textbf{WSL}} & \multicolumn{3}{l|}{\textbf{SM}} \\ \hline
\textbf{Class} & \textbf{Flip} & \textbf{VAE} & \textbf{Knockoff} & \textbf{Flip} & \textbf{VAE} & \textbf{Knockoff} \\ \hline
\textbf{0} & \textbf{100.000} & \textbf{100.000} & \textbf{100.000} & 0.891 & \textbf{{0.084}} & {0.128} \\ \hline
\textbf{1} & \textbf{99.453} & 88.389 & 82.249 & \textbf{{-1.195}} & -1.131 & -1.118 \\ \hline
\textbf{2} & \textbf{100.000} & \textbf{100.000} & \textbf{100.000} & \textbf{{-0.461}} & -0.435 & -0.414 \\ \hline
\textbf{3} & \textbf{100.000} & 98.686 & 99.635 & -0.263 & -0.250 & \textbf{{-0.299}} \\ \hline
\textbf{4} & \textbf{100.000} & 99.732 & 98.523 & \textbf{{-0.639}} & -0.495 & -0.588 \\ \hline
\textbf{5} & \textbf{100.000} & 99.858 & 99.858 & \textbf{{-0.664}} & -0.589 & -0.570 \\ \hline
\textbf{6} & \textbf{100.000} & \textbf{100.000} & 98.784 & \textbf{-0.624} & -0.610 & -0.538 \\ \hline
\textbf{7} & \textbf{100.000} & 99.553 & 99.617 & -0.626 & \textbf{{-0.655}} & -0.577 \\ \hline
\textbf{8} & \textbf{100.000} & \textbf{100.000} & 99.310 & 0.333 & 0.253 & \textbf{{-0.036}} \\ \hline
\textbf{9} & \textbf{99.871} & 97.613 & 98.645 & 0.916 & 0.209 & \textbf{{0.142}} \\ \hline
\end{tabular}
}
\end{minipage}
\end{table}

\vspace{-2em}

\begin{table}[H]
\centering
\begin{minipage}[t]{.48\textwidth}
\centering
\caption{Results of the class-wise evaluation of weakly supervised location (WSL) and saliency metric (SM) on MNIST counterfactuals with 0.6 threshold for the \textbf{SSR} objective. Flip scores best for classes: 1, 2, 4, VAE scores best for classes: 5, 6, 7, 9 and knockoff scores best for classes: 0, 3, and 8. Best scores are in bold.}
\resizebox{\textwidth}{!}{
\renewcommand{\arraystretch}{0.5}
\begin{tabular}{|l|l|l|l|l|l|l|}
\hline
 & \multicolumn{3}{l|}{\textbf{WSL}} & \multicolumn{3}{l|}{\textbf{SM}} \\ \hline
\textbf{Class} & \textbf{Flip} & \textbf{VAE} & \textbf{Knockoff} & \textbf{Flip} & \textbf{VAE} & \textbf{Knockoff} \\ \hline
\textbf{0} & \textbf{100.000} & \textbf{100.000} & \textbf{100.000} & 1.469 & 0.885 & \textbf{{0.118}}\\ \hline
\textbf{1} & \textbf{98.480} & 97.751 & 91.489 & \textbf{{-1.175}} & -1.165 & {-1.170} \\ \hline
\textbf{2} & \textbf{100.000} & \textbf{100.000} & \textbf{100.000} & \textbf{{-0.538}} & {-0.432} & -0.278 \\ \hline
\textbf{3} & \textbf{100.000} & \textbf{100.000} & 99.197 & -0.051 & 1.490 & \textbf{{-0.197}} \\ \hline
\textbf{4} & \textbf{100.000} & 99.463 & 98.993 & \textbf{{-0.753}} & {-0.720} & -0.565 \\ \hline
\textbf{5} & \textbf{100.000} & \textbf{100.000} & 98.936 & -0.708 & \textbf{ {-0.760}} & -0.632 \\ \hline
\textbf{6} & \textbf{100.000} & 99.797 & 98.986 & -0.628 & \textbf{{-0.750}} & -0.563 \\ \hline
\textbf{7} & \textbf{100.000} & 99.936 & \textbf{100.000} & -0.586 & \textbf{{-0.758}} & -0.747 \\ \hline
\textbf{8} & \textbf{100.000} & \textbf{100.000} & 99.862 & 0.433 & 0.508 & \textbf{{0.072}}\\ \hline
\textbf{9} & \textbf{100.000} & \textbf{100.000}0 & 99.419 & 1.200 & \textbf{{0.349}} & {0.452} \\ \hline
\end{tabular}
}
\end{minipage} \hfill %
\begin{minipage}[t]{.48\textwidth}
\centering
\caption{Results of the class-wise evaluation of weakly supervised location (WSL) and saliency metric (SM) on MNIST counterfactuals with 0.6 threshold for the \textbf{SDR} objective. Flip scores best for classes: 4, 5, VAE scores best for classes: 0, 1, 6, 7 and knockoff scores best for classes: 2, 3, 8, 9. Best scores are in bold.}
\resizebox{\textwidth}{!}{
\renewcommand{\arraystretch}{0.5}
\begin{tabular}{|l|l|l|l|l|l|l|}
\hline
 & \multicolumn{3}{l|}{\textbf{WSL}} & \multicolumn{3}{l|}{\textbf{SM}} \\ \hline
\textbf{Class} & \textbf{Flip} & \textbf{VAE} & \textbf{Knockoff} & \textbf{Flip} & \textbf{VAE} & \textbf{Knockoff} \\ \hline
\textbf{0} & \textbf{100.000} & \textbf{100.000} & \textbf{100.000} & 1.405 & \textbf{{0.234}} & {0.411} \\ \hline
\textbf{1} & \textbf{99.635} & 94.043 & 85.410 & {-1.182} & \textbf{{-1.197}} & -1.148 \\ \hline
\textbf{2} & \textbf{100.000} & \textbf{100.000} & \textbf{100.000} & {-0.447} & -0.450 & \textbf{{-0.485}} \\ \hline
\textbf{3} & \textbf{100.000} & 98.540 & 99.854 & -0.082 & -0.190 & \textbf{{-0.237}} \\ \hline
\textbf{4} & \textbf{100.000} & 99.664 & 98.591 & \textbf{{-0.705}} & -0.519 & -0.631 \\ \hline
\textbf{5} & \textbf{100.000} & 99.929 & \textbf{100.000} & \textbf{{-0.737}} & -0.623 & -0.586 \\ \hline
\textbf{6} & \textbf{100.000} & \textbf{100.000} & 98.986 & {-0.599} & \textbf{{0.667}} & -0.603 \\ \hline
\textbf{7} & \textbf{100.000} & 99.872 & 99.681 & -0.678 & \textbf{{-0.697}} & -0.631 \\ \hline
\textbf{8} & \textbf{100.000} & \textbf{100.000} & \textbf{100.000} & 0.587 & 0.304 & \textbf{{-0.005}} \\ \hline
\textbf{9} & \textbf{99.871} & 98.258 & 98.839 & 1.147 & 0.404 & \textbf{{0.233}} \\ \hline
\end{tabular}
}
\end{minipage}
\end{table}

\subsection{Additional examples}

% -------------------------------------------------------------------
\subsubsection{MNIST counterfactuals}
\label{app:mnist}

\vspace{-1em}

\begin{figure}[H]
    \centering\includegraphics[width=1.0cm]{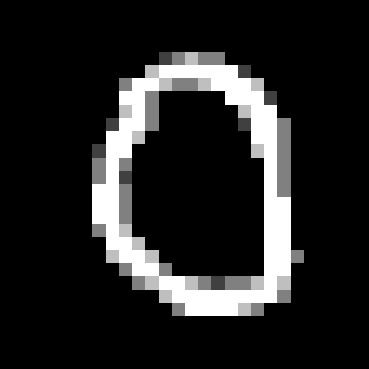}
    \centering\includegraphics[width=1.0cm]{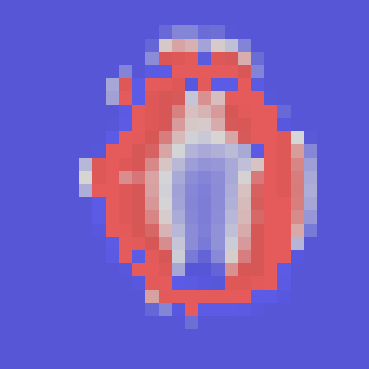}
    \centering\includegraphics[width=1.0cm]{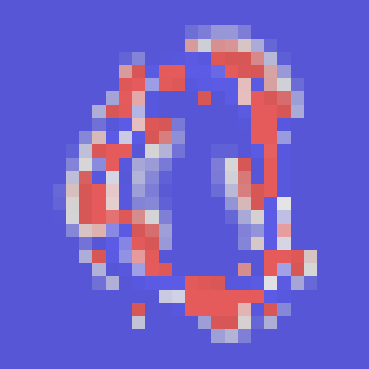}
    \centering\includegraphics[width=1.0cm]{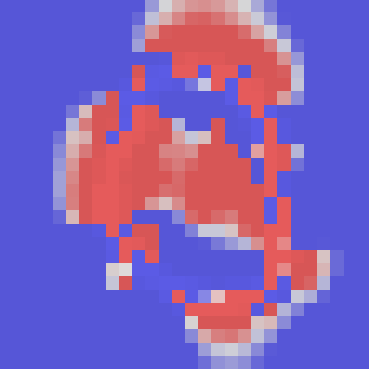}
    \centering\includegraphics[width=1.0cm]{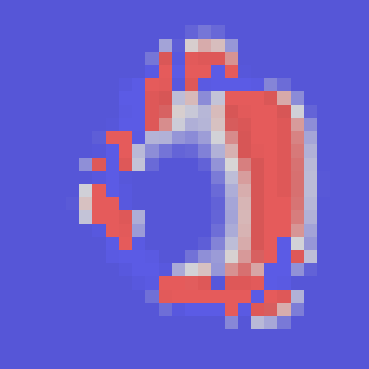}
    \centering\includegraphics[width=1.0cm]{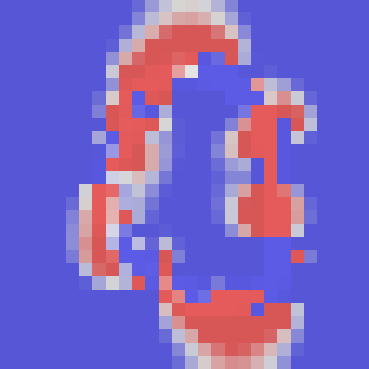}
    \centering\includegraphics[width=1.0cm]{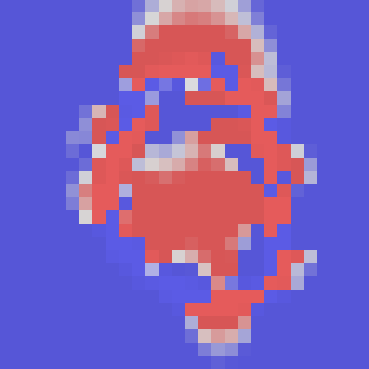}
    \centering\includegraphics[width=1.0cm]{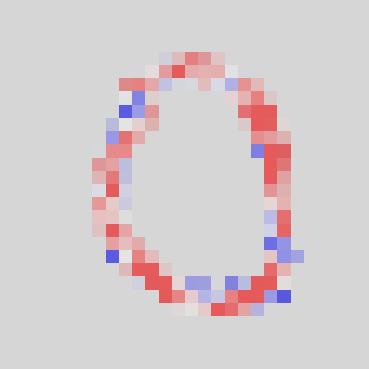}
    \centering\includegraphics[width=1.0cm]{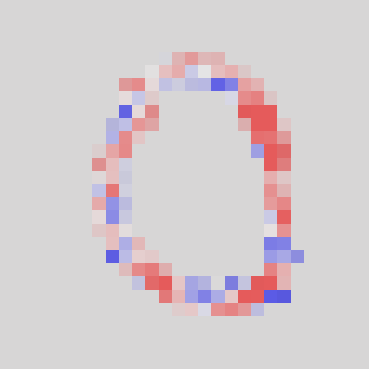}
    \centering\includegraphics[width=1.0cm]{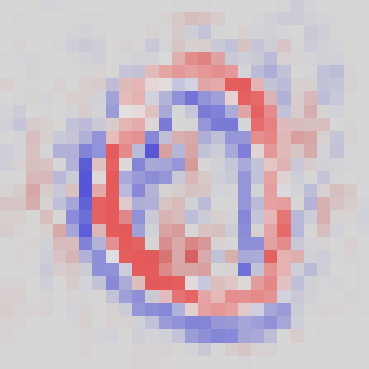}
    
    \centering\includegraphics[width=1.0cm]{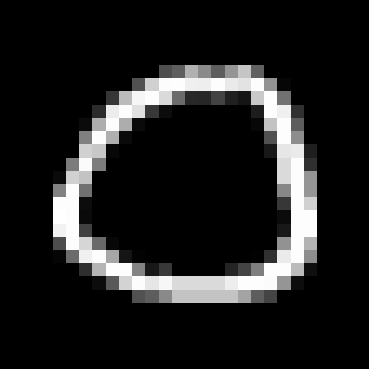}
    \centering\includegraphics[width=1.0cm]{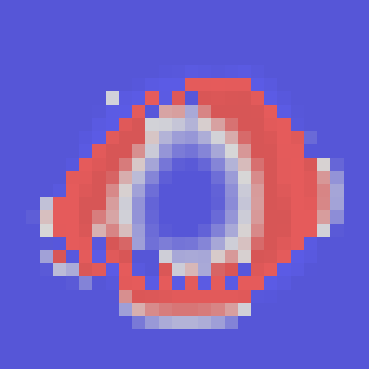}
    \centering\includegraphics[width=1.0cm]{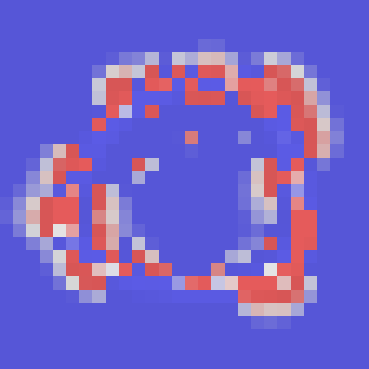}
    \centering\includegraphics[width=1.0cm]{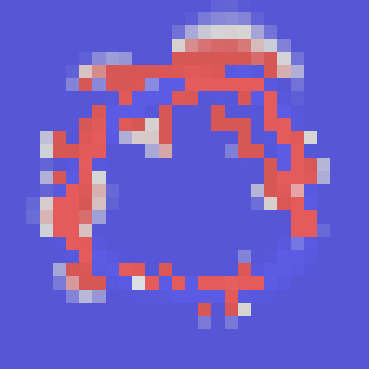}
    \centering\includegraphics[width=1.0cm]{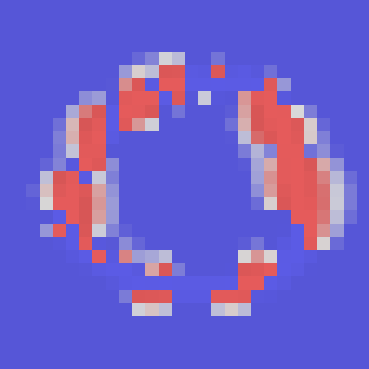}
    \centering\includegraphics[width=1.0cm]{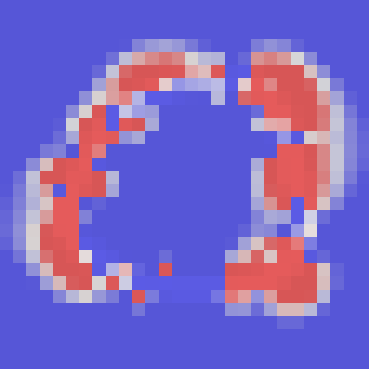}
    \centering\includegraphics[width=1.0cm]{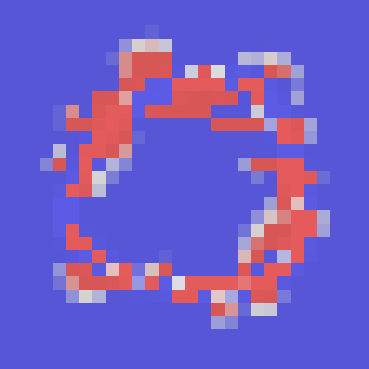}
    \centering\includegraphics[width=1.0cm]{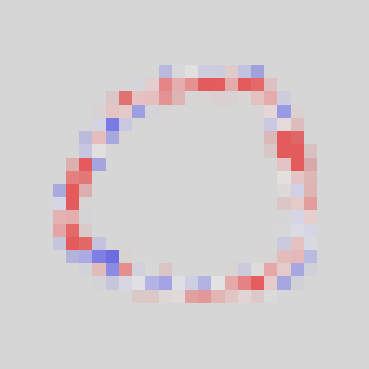}
    \centering\includegraphics[width=1.0cm]{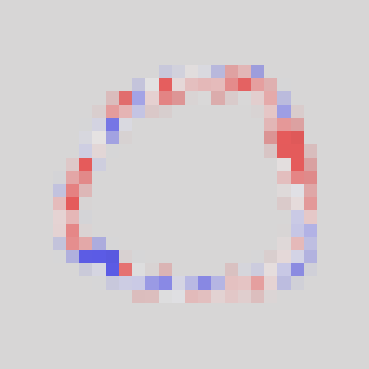}
    \centering\includegraphics[width=1.0cm]{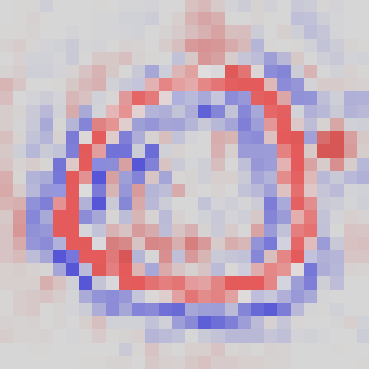}

    \centering\includegraphics[width=1.0cm]{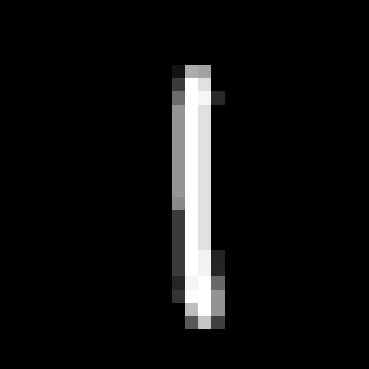}
    \centering\includegraphics[width=1.0cm]{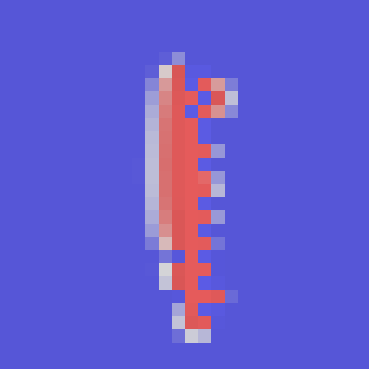}
    \centering\includegraphics[width=1.0cm]{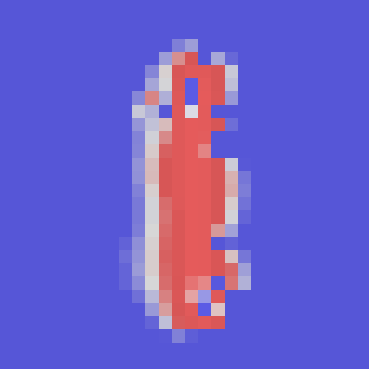}
    \centering\includegraphics[width=1.0cm]{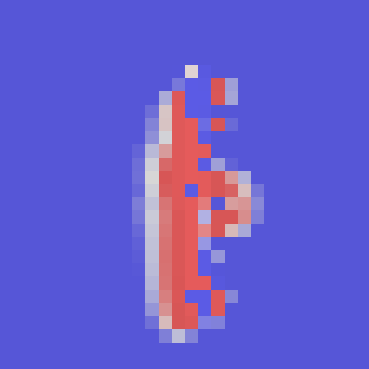}
    \centering\includegraphics[width=1.0cm]{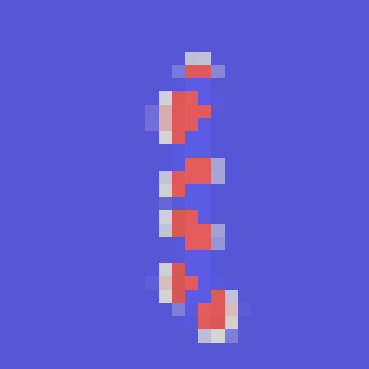}
    \centering\includegraphics[width=1.0cm]{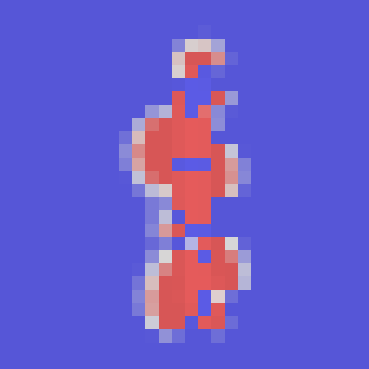}
    \centering\includegraphics[width=1.0cm]{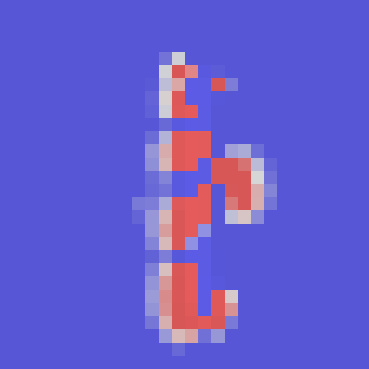}
    \centering\includegraphics[width=1.0cm]{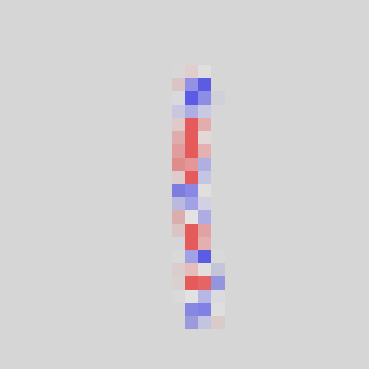}
    \centering\includegraphics[width=1.0cm]{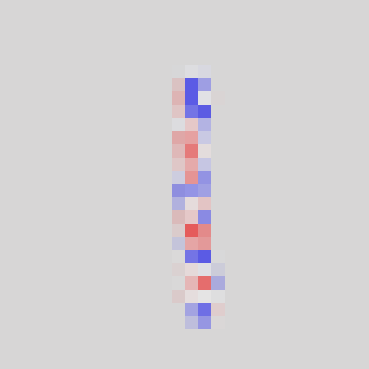}
    \centering\includegraphics[width=1.0cm]{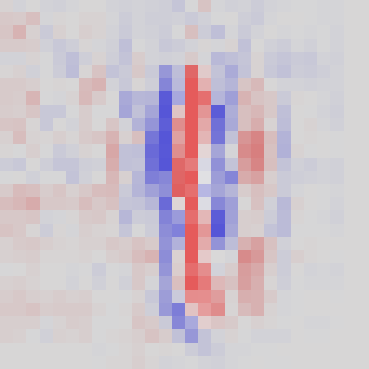}

    \centering\includegraphics[width=1.0cm]{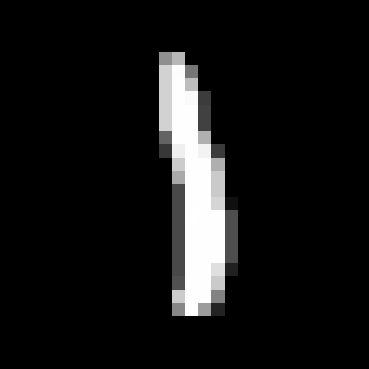}
    \centering\includegraphics[width=1.0cm]{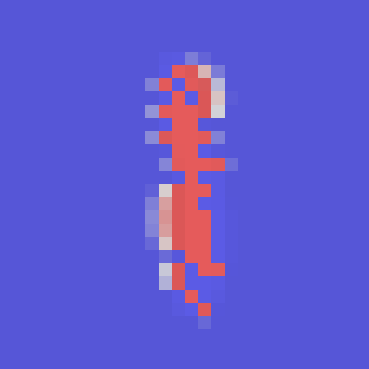}
    \centering\includegraphics[width=1.0cm]{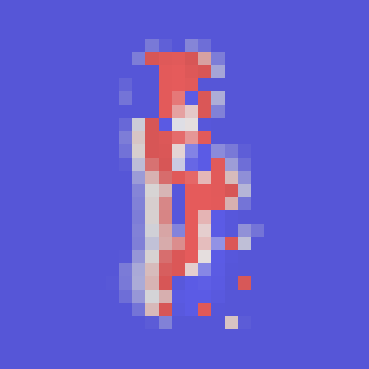}
    \centering\includegraphics[width=1.0cm]{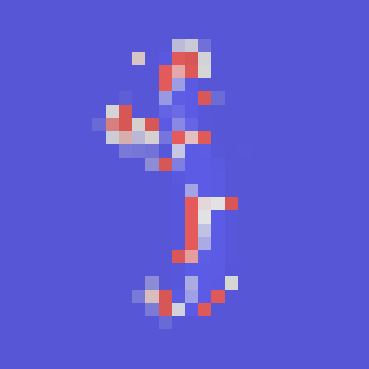}
    \centering\includegraphics[width=1.0cm]{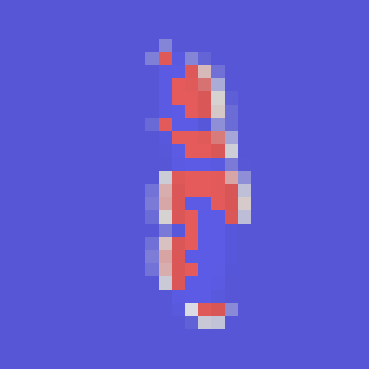}
    \centering\includegraphics[width=1.0cm]{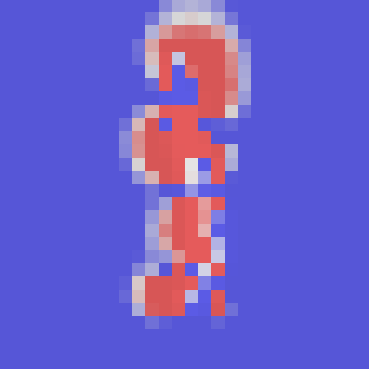}
    \centering\includegraphics[width=1.0cm]{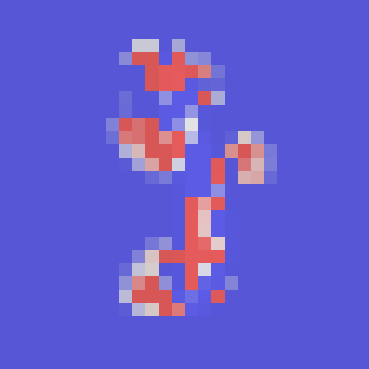}
    \centering\includegraphics[width=1.0cm]{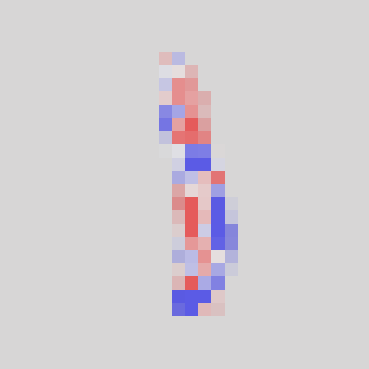}
    \centering\includegraphics[width=1.0cm]{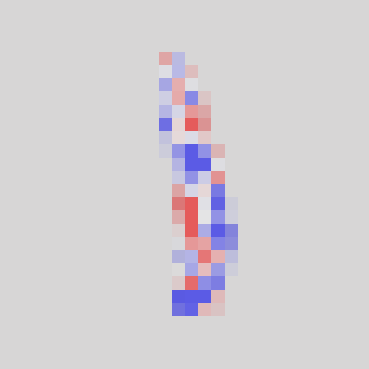}
    \centering\includegraphics[width=1.0cm]{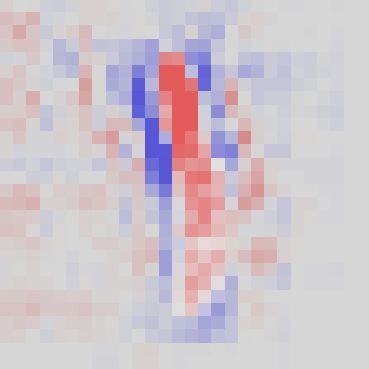}
    
    \centering\includegraphics[width=1.0cm]{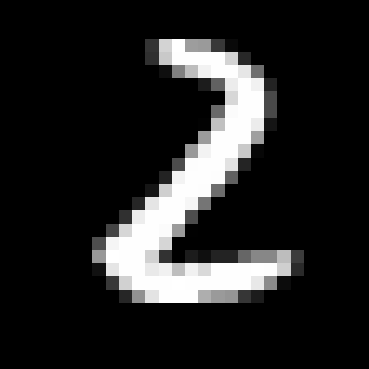}
    \centering\includegraphics[width=1.0cm]{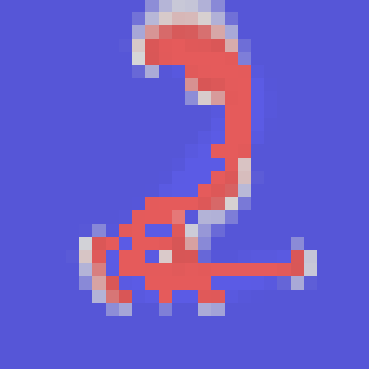}
    \centering\includegraphics[width=1.0cm]{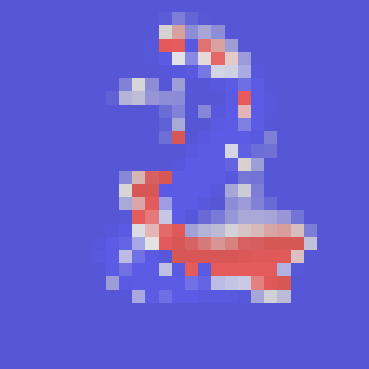}
    \centering\includegraphics[width=1.0cm]{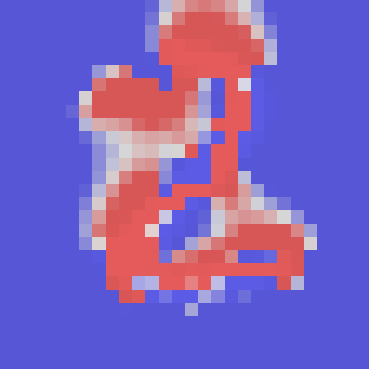}
    \centering\includegraphics[width=1.0cm]{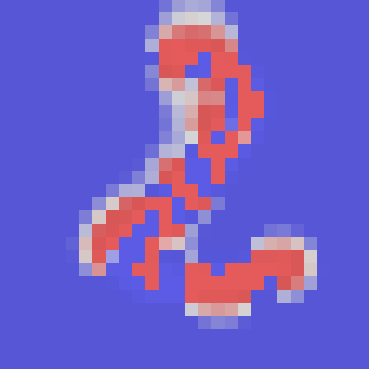}
    \centering\includegraphics[width=1.0cm]{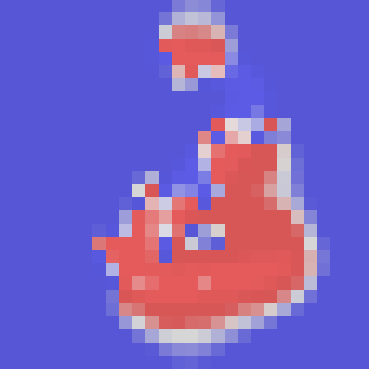}
    \centering\includegraphics[width=1.0cm]{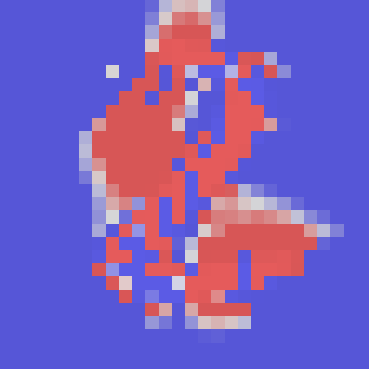}
    \centering\includegraphics[width=1.0cm]{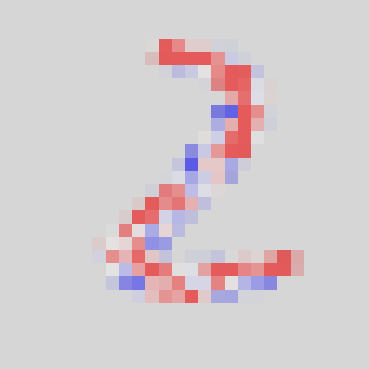}
    \centering\includegraphics[width=1.0cm]{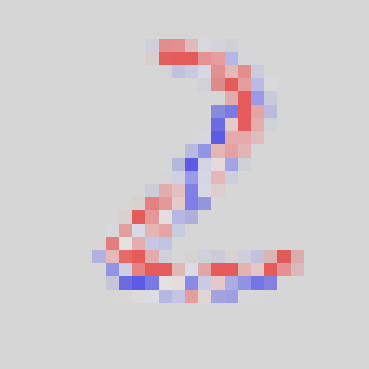}
    \centering\includegraphics[width=1.0cm]{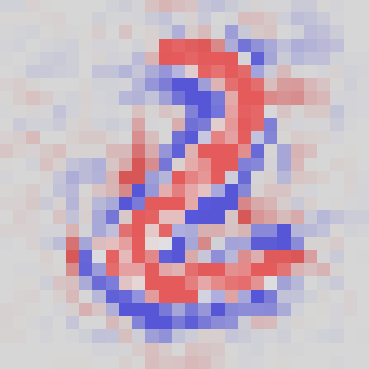}

    \centering\includegraphics[width=1.0cm]{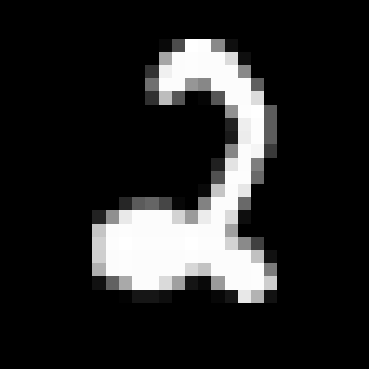}
    \centering\includegraphics[width=1.0cm]{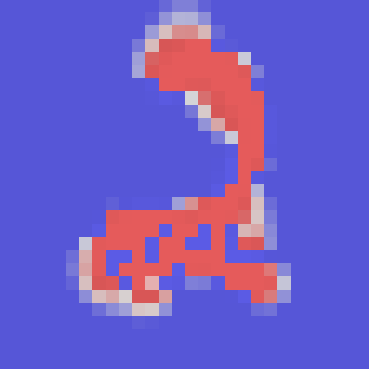}
    \centering\includegraphics[width=1.0cm]{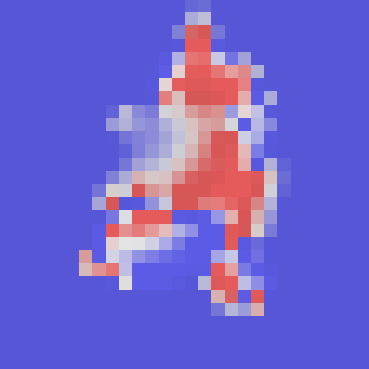}
    \centering\includegraphics[width=1.0cm]{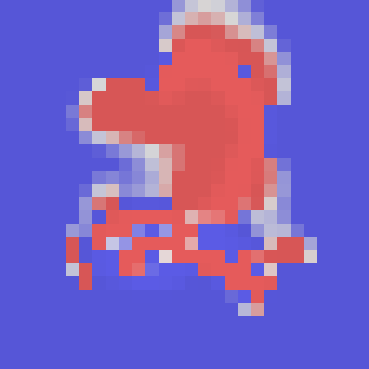}
    \centering\includegraphics[width=1.0cm]{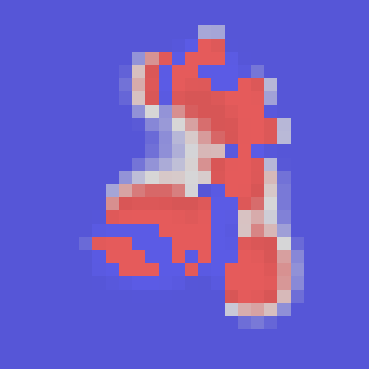}
    \centering\includegraphics[width=1.0cm]{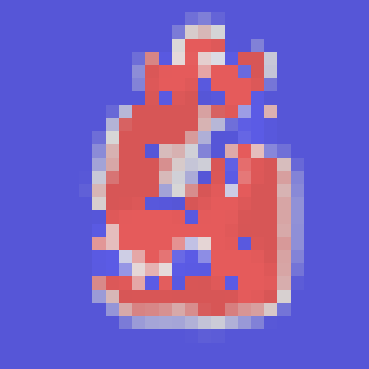}
    \centering\includegraphics[width=1.0cm]{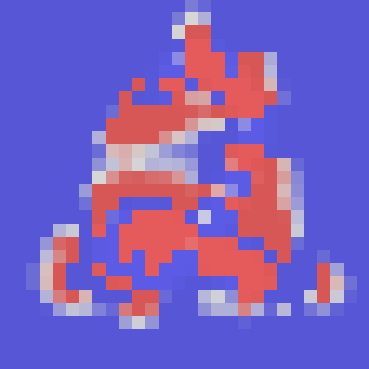}
    \centering\includegraphics[width=1.0cm]{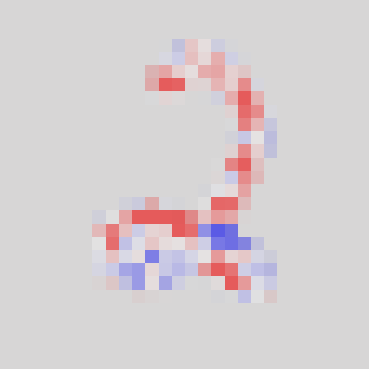}
    \centering\includegraphics[width=1.0cm]{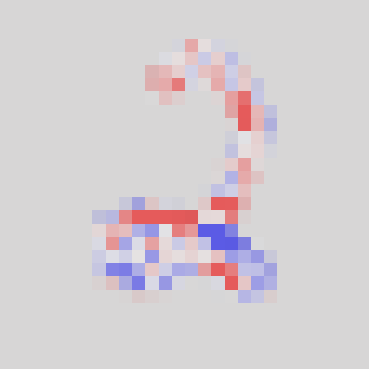}
    \centering\includegraphics[width=1.0cm]{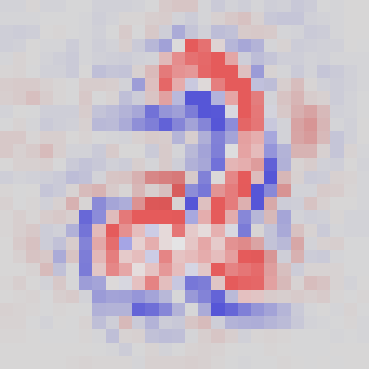}
    
    \centering\includegraphics[width=1.0cm]{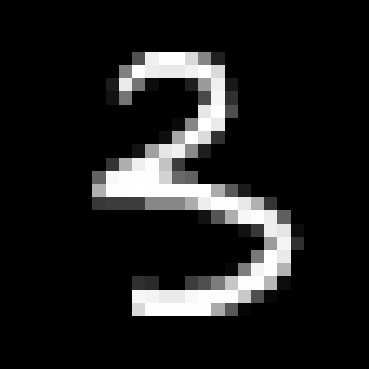}
    \centering\includegraphics[width=1.0cm]{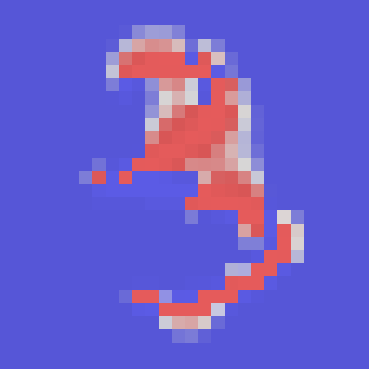}
    \centering\includegraphics[width=1.0cm]{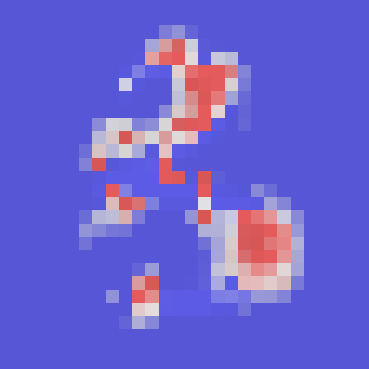}
    \centering\includegraphics[width=1.0cm]{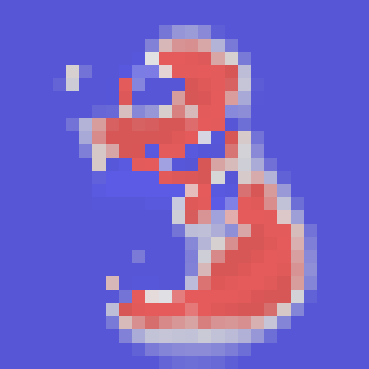}
    \centering\includegraphics[width=1.0cm]{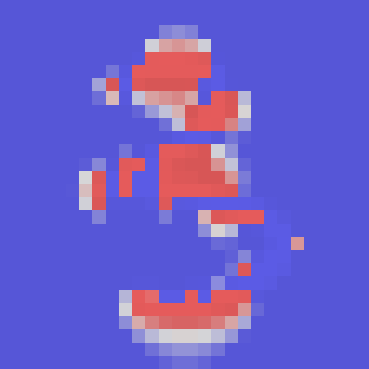}
    \centering\includegraphics[width=1.0cm]{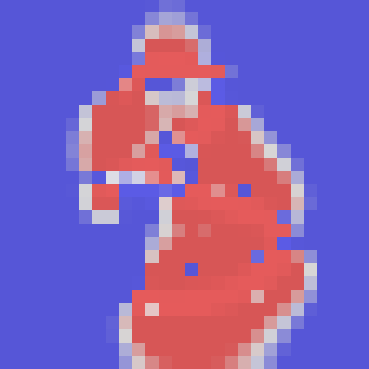}
    \centering\includegraphics[width=1.0cm]{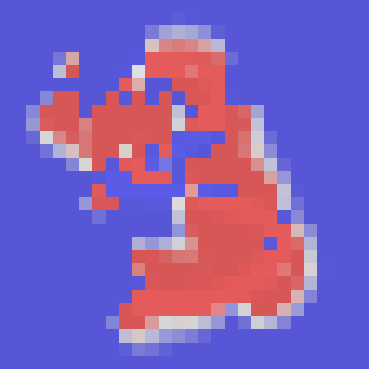}
    \centering\includegraphics[width=1.0cm]{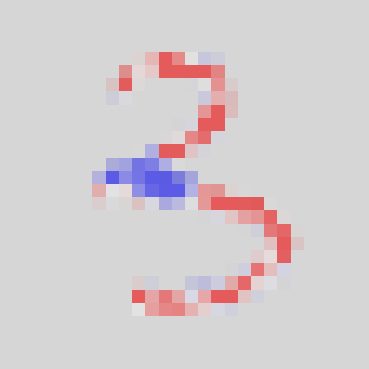}
    \centering\includegraphics[width=1.0cm]{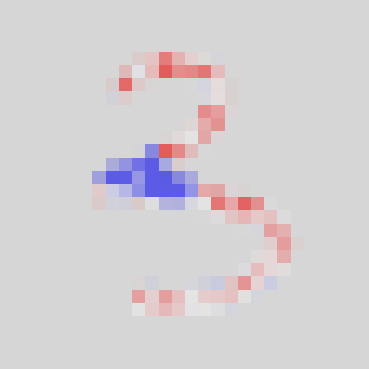}
    \centering\includegraphics[width=1.0cm]{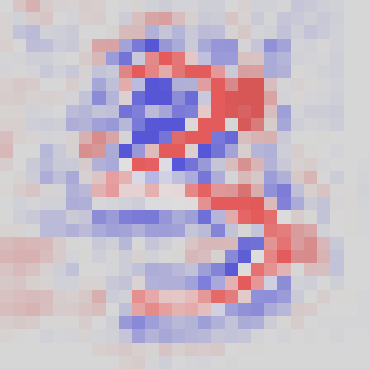}

    \centering\includegraphics[width=1.0cm]{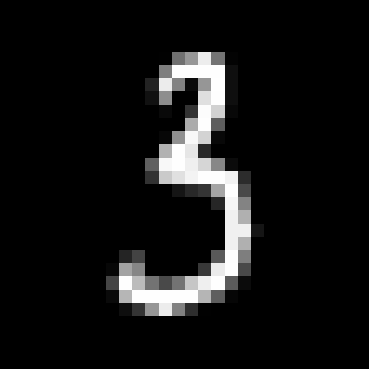}
    \centering\includegraphics[width=1.0cm]{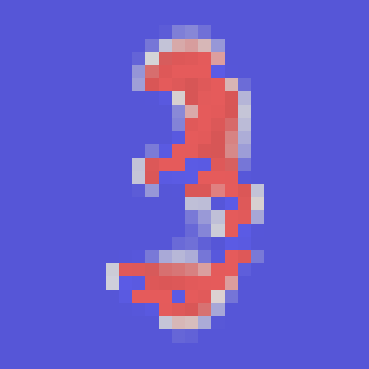}
    \centering\includegraphics[width=1.0cm]{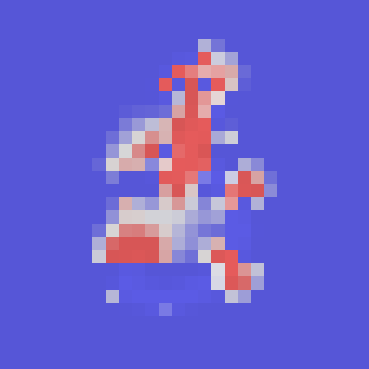}
    \centering\includegraphics[width=1.0cm]{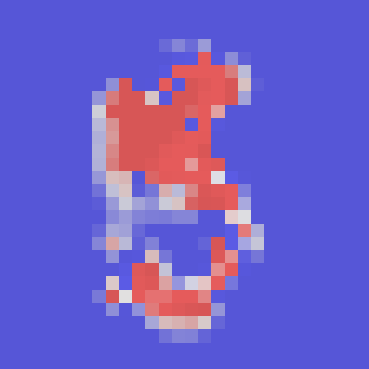}
    \centering\includegraphics[width=1.0cm]{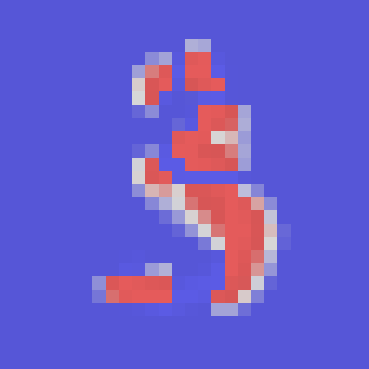}
    \centering\includegraphics[width=1.0cm]{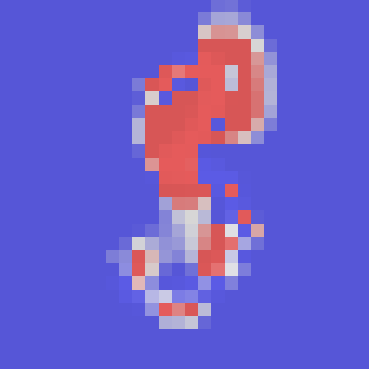}
    \centering\includegraphics[width=1.0cm]{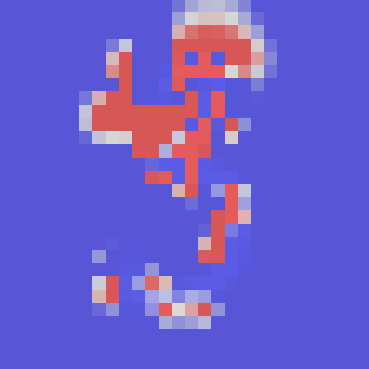}
    \centering\includegraphics[width=1.0cm]{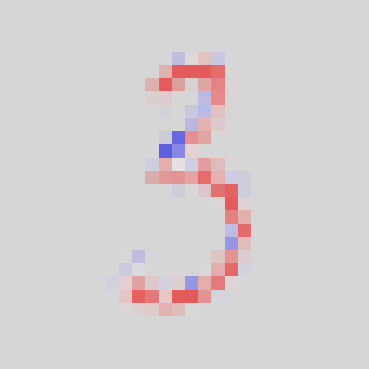}
    \centering\includegraphics[width=1.0cm]{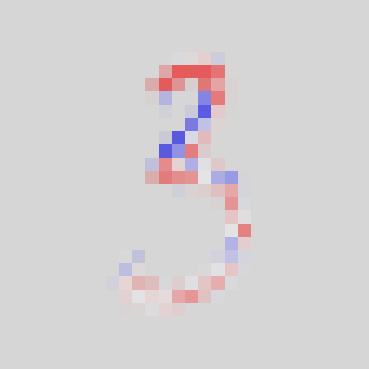}
    \centering\includegraphics[width=1.0cm]{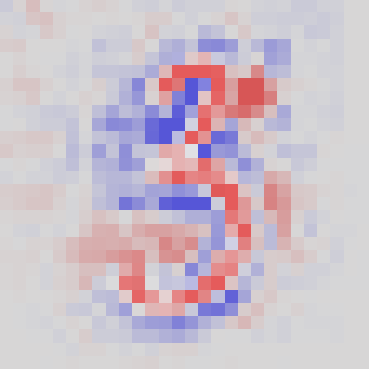}

    \centering\includegraphics[width=1.0cm]{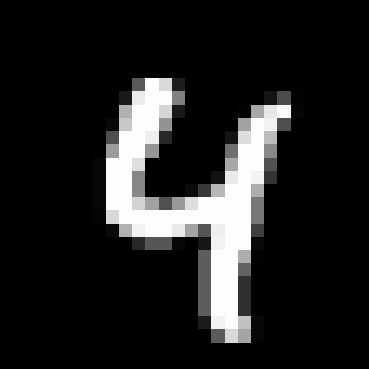}
    \centering\includegraphics[width=1.0cm]{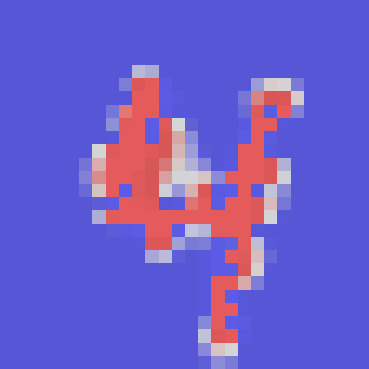}
    \centering\includegraphics[width=1.0cm]{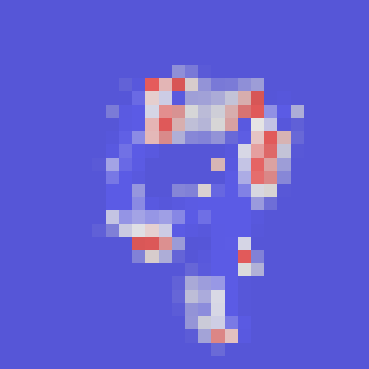}
    \centering\includegraphics[width=1.0cm]{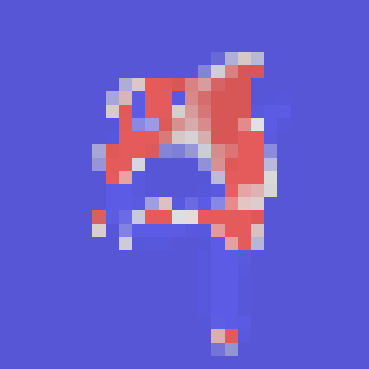}
    \centering\includegraphics[width=1.0cm]{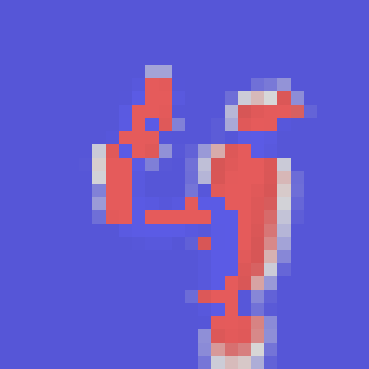}
    \centering\includegraphics[width=1.0cm]{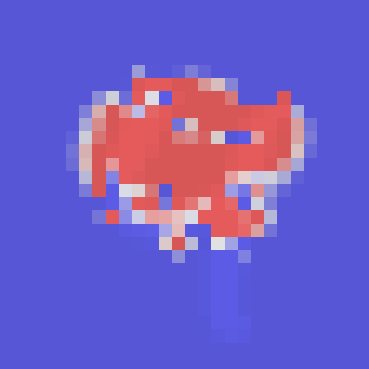}
    \centering\includegraphics[width=1.0cm]{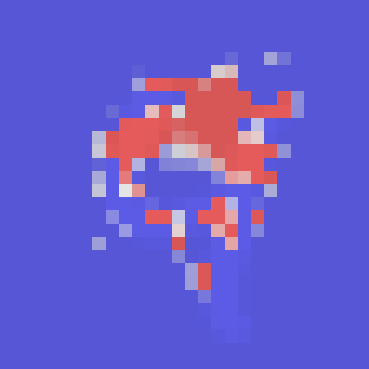}
    \centering\includegraphics[width=1.0cm]{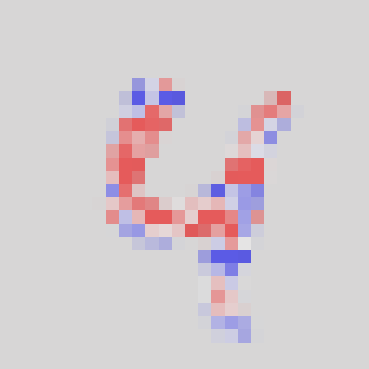}
    \centering\includegraphics[width=1.0cm]{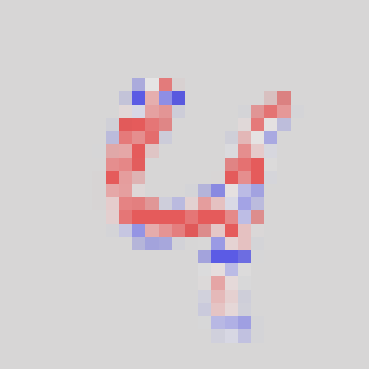}
    \centering\includegraphics[width=1.0cm]{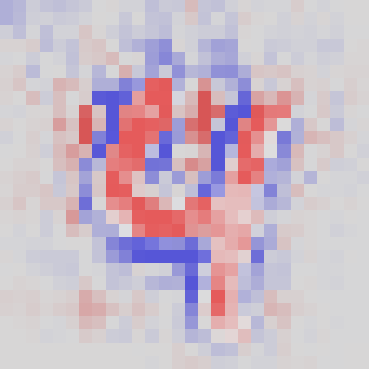}

    \centering\includegraphics[width=1.0cm]{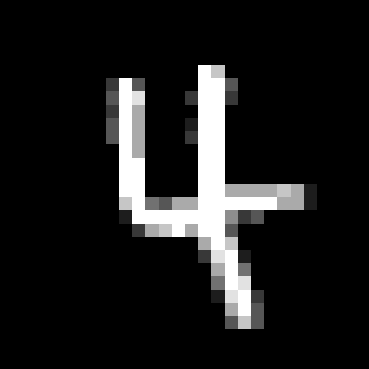}
    \centering\includegraphics[width=1.0cm]{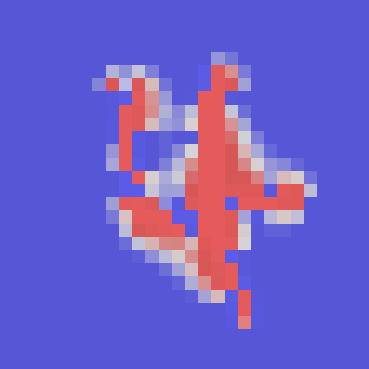}
    \centering\includegraphics[width=1.0cm]{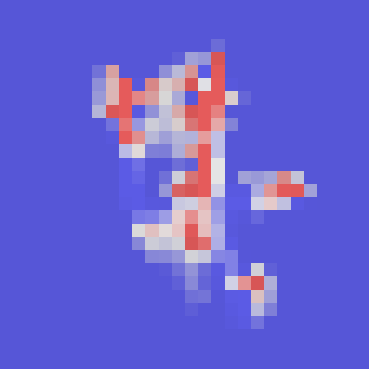}
    \centering\includegraphics[width=1.0cm]{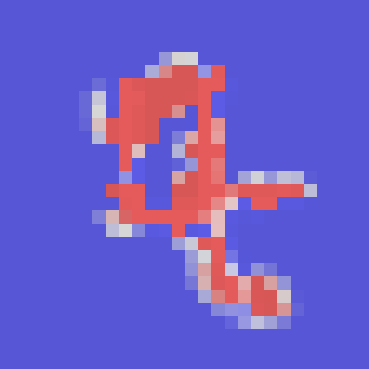}
    \centering\includegraphics[width=1.0cm]{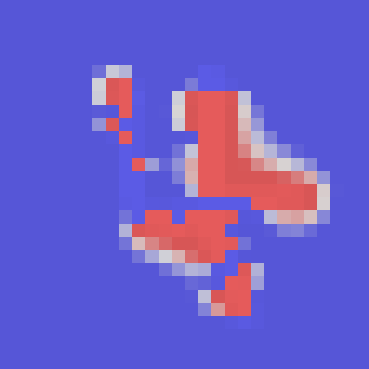}
    \centering\includegraphics[width=1.0cm]{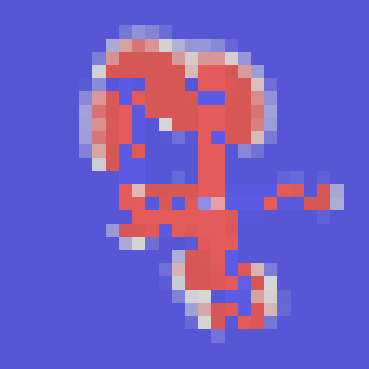}
    \centering\includegraphics[width=1.0cm]{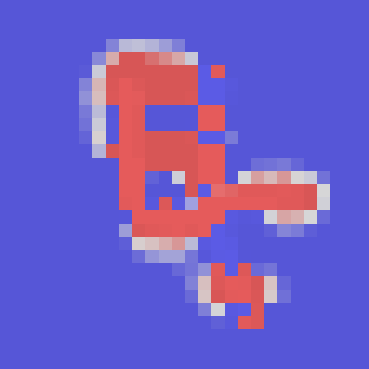}
    \centering\includegraphics[width=1.0cm]{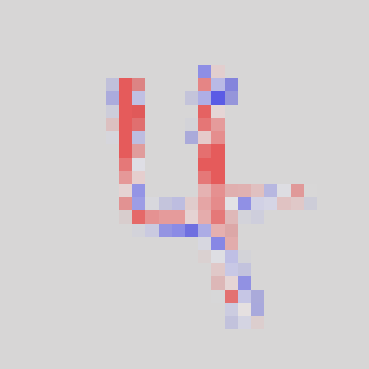}
    \centering\includegraphics[width=1.0cm]{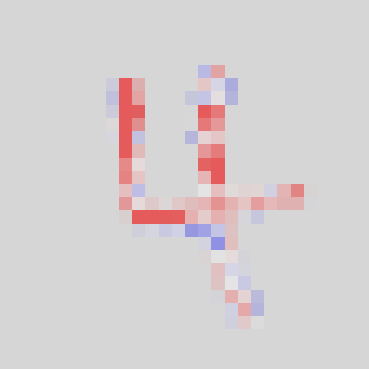}
    \centering\includegraphics[width=1.0cm]{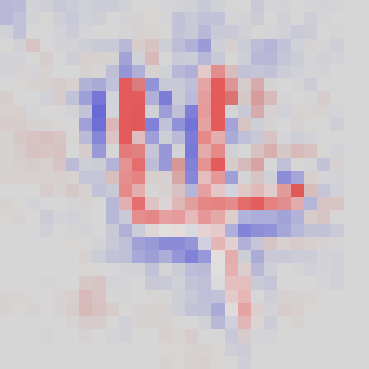}
    
    \centering\includegraphics[width=1.0cm]{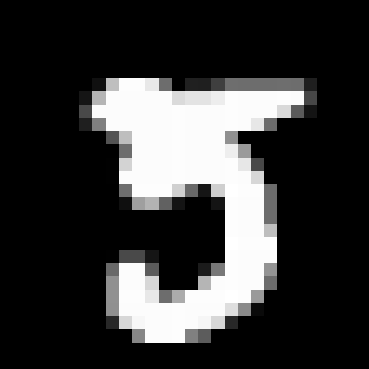}
    \centering\includegraphics[width=1.0cm]{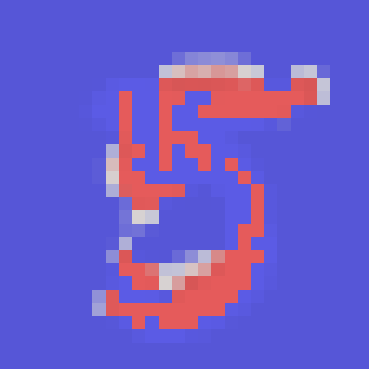}
    \centering\includegraphics[width=1.0cm]{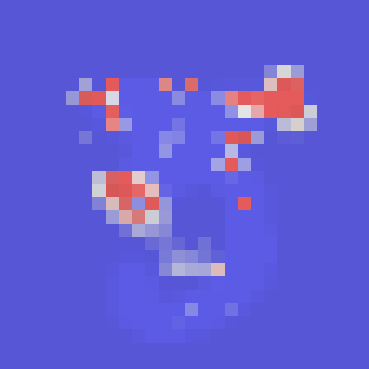}
    \centering\includegraphics[width=1.0cm]{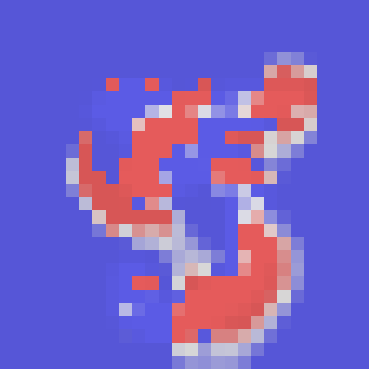}
    \centering\includegraphics[width=1.0cm]{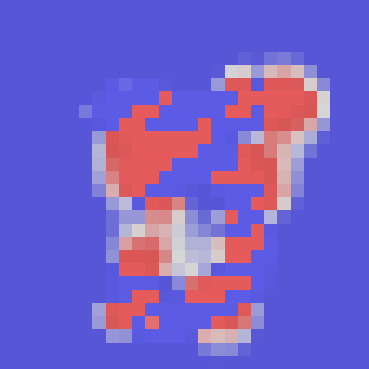}
    \centering\includegraphics[width=1.0cm]{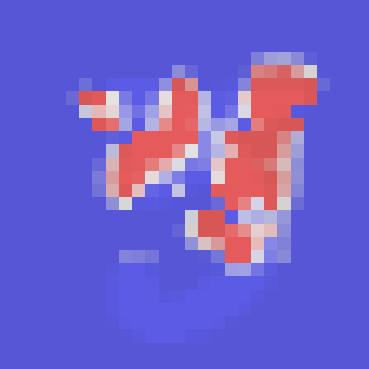}
    \centering\includegraphics[width=1.0cm]{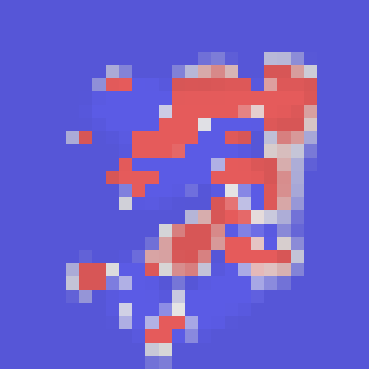}
    \centering\includegraphics[width=1.0cm]{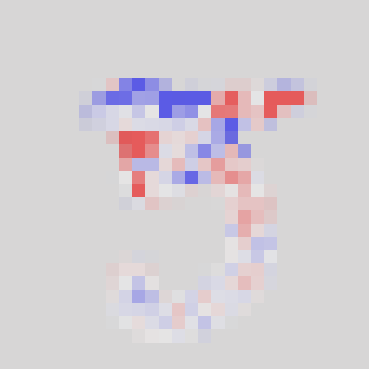}
    \centering\includegraphics[width=1.0cm]{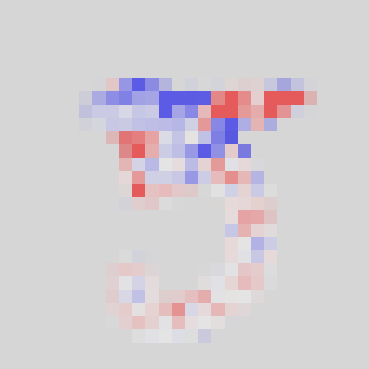}
    \centering\includegraphics[width=1.0cm]{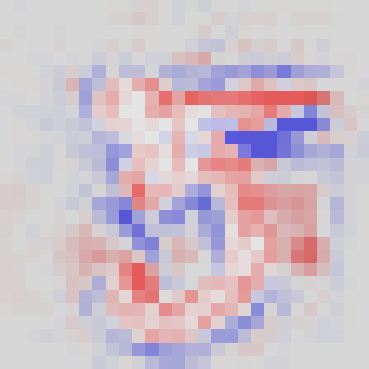}

    \centering\includegraphics[width=1.0cm]{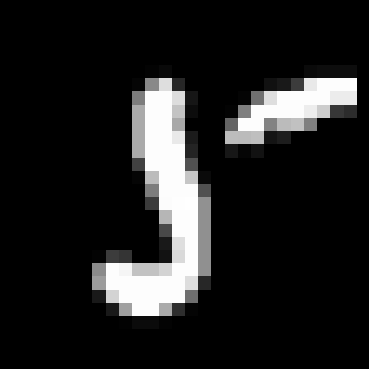}
    \centering\includegraphics[width=1.0cm]{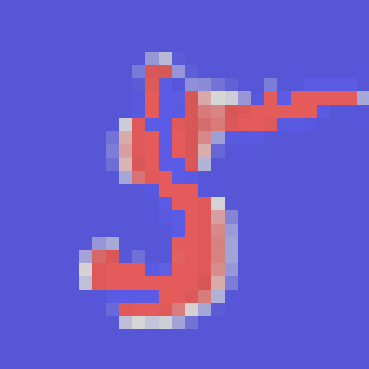}
    \centering\includegraphics[width=1.0cm]{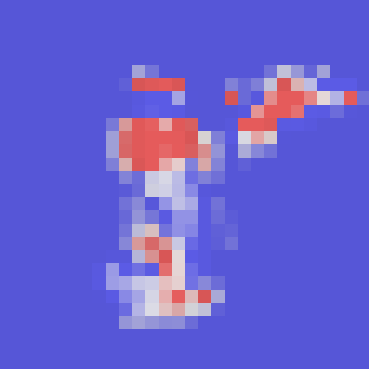}
    \centering\includegraphics[width=1.0cm]{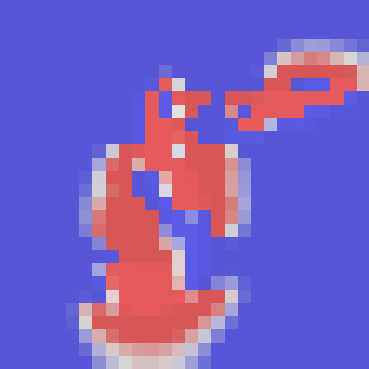}
    \centering\includegraphics[width=1.0cm]{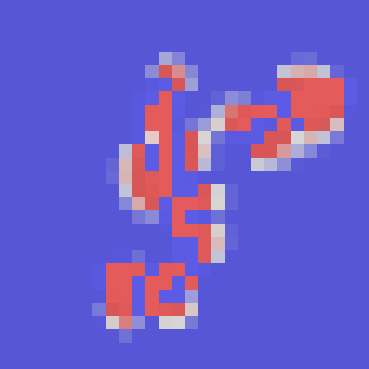}
    \centering\includegraphics[width=1.0cm]{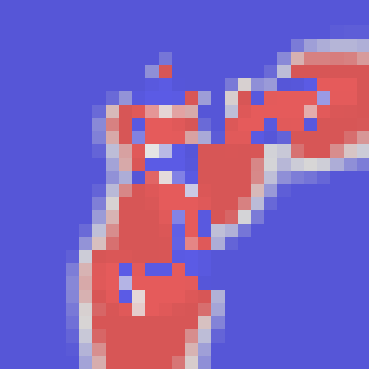}
    \centering\includegraphics[width=1.0cm]{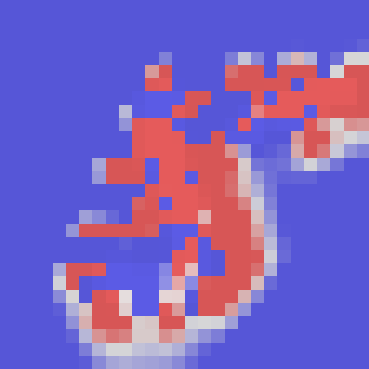}
    \centering\includegraphics[width=1.0cm]{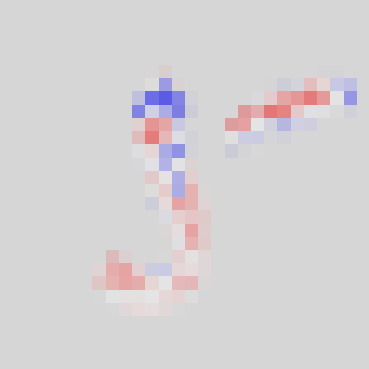}
    \centering\includegraphics[width=1.0cm]{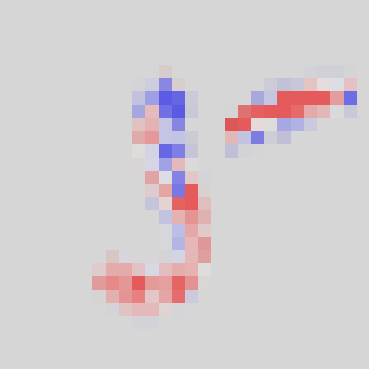}
    \centering\includegraphics[width=1.0cm]{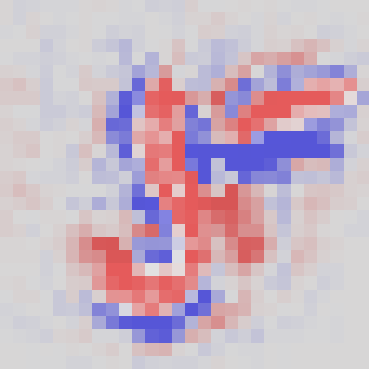}

    \centering\includegraphics[width=1.0cm]{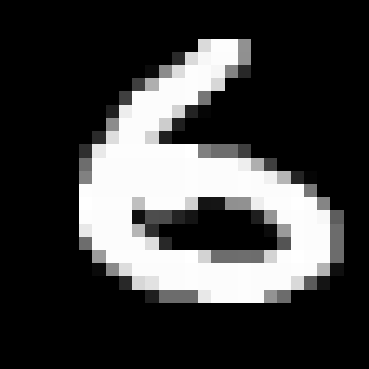}
    \centering\includegraphics[width=1.0cm]{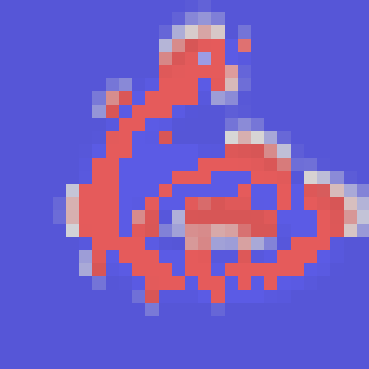}
    \centering\includegraphics[width=1.0cm]{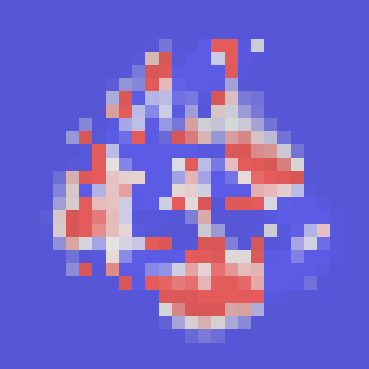}
    \centering\includegraphics[width=1.0cm]{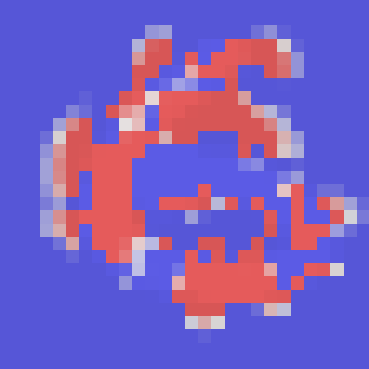}
    \centering\includegraphics[width=1.0cm]{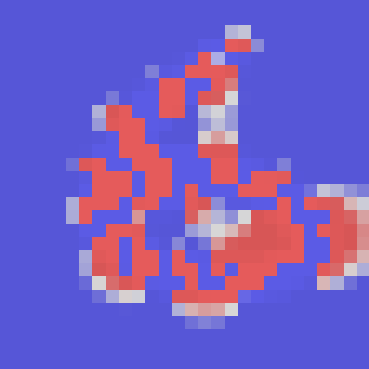}
    \centering\includegraphics[width=1.0cm]{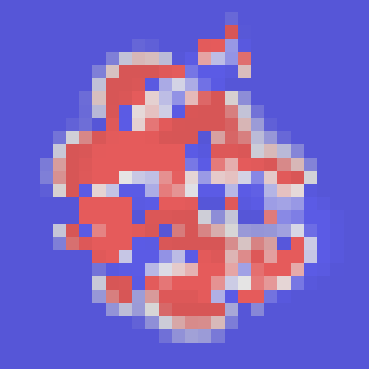}
    \centering\includegraphics[width=1.0cm]{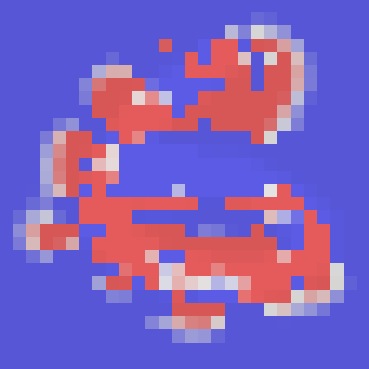}
    \centering\includegraphics[width=1.0cm]{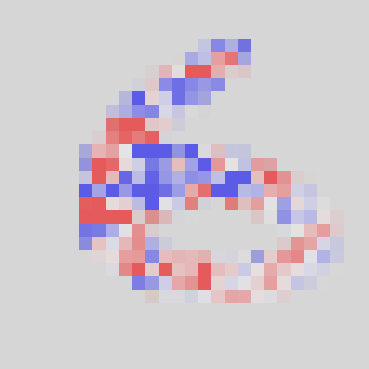}
    \centering\includegraphics[width=1.0cm]{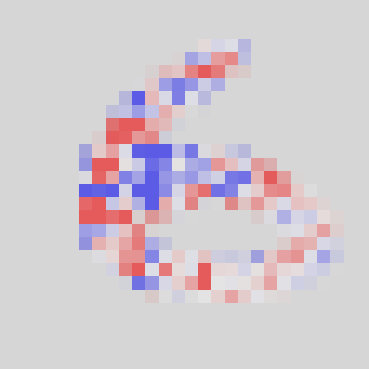}
    \centering\includegraphics[width=1.0cm]{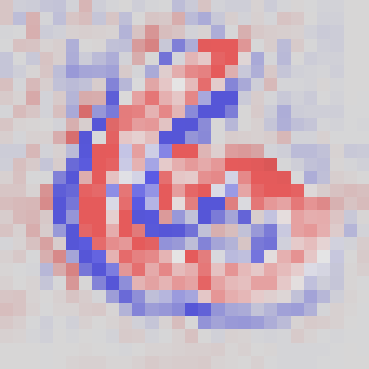}
    
    \centering\includegraphics[width=1.0cm]{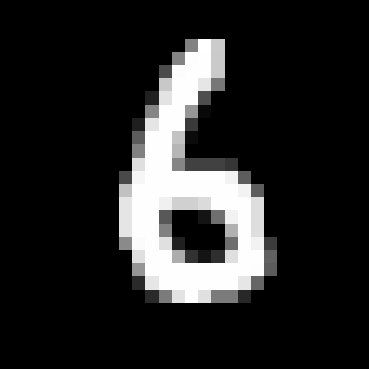}
    \centering\includegraphics[width=1.0cm]{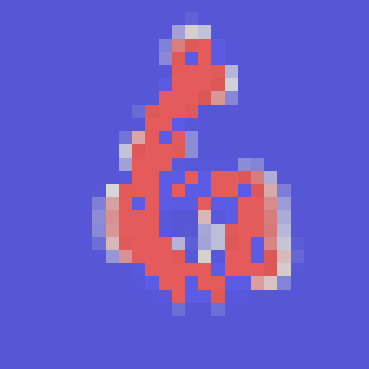}
    \centering\includegraphics[width=1.0cm]{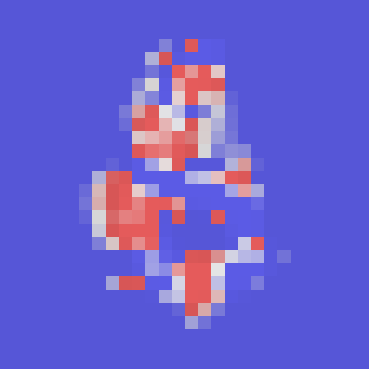}
    \centering\includegraphics[width=1.0cm]{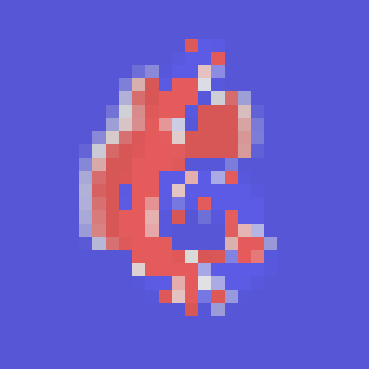}
    \centering\includegraphics[width=1.0cm]{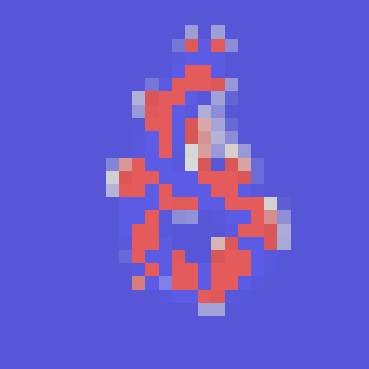}
    \centering\includegraphics[width=1.0cm]{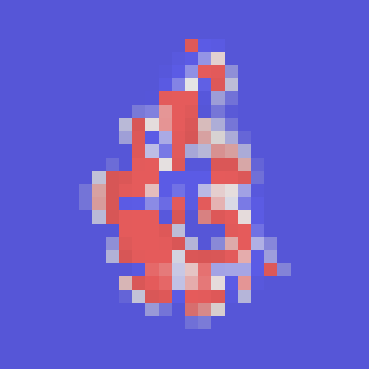}
    \centering\includegraphics[width=1.0cm]{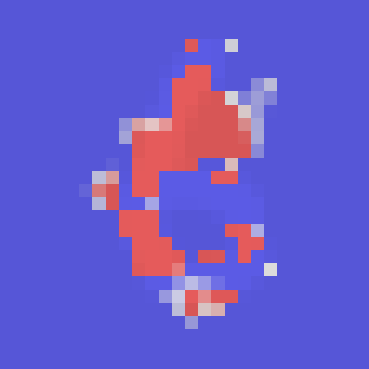}
    \centering\includegraphics[width=1.0cm]{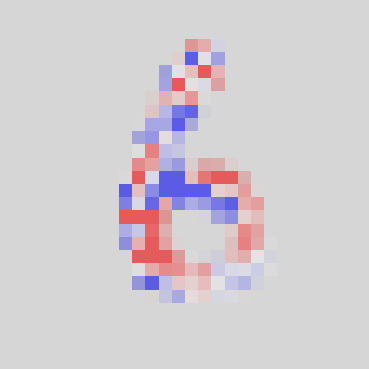}
    \centering\includegraphics[width=1.0cm]{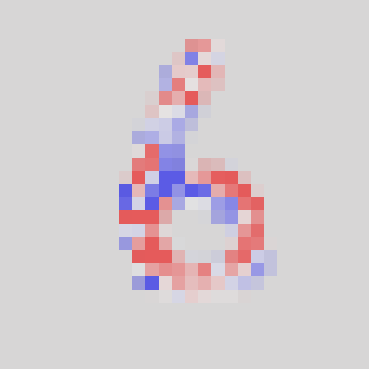}
    \centering\includegraphics[width=1.0cm]{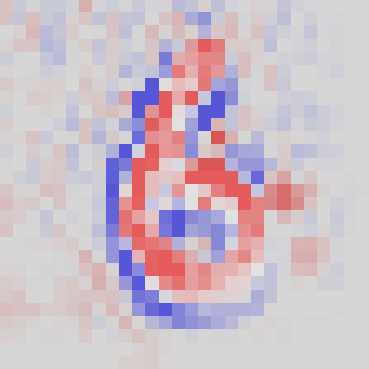}

%  \caption{Examples of counterfactuals for samples from MNIST. Left to right: original image, SSR flip, SSR VAE, SSR knockoff, SDR flip, SDR VAE, SDR knockoff, Gradients $\times$ Input, Integrated Gradients, Guided GradCAM.}
% \end{figure}

% \begin{figure}[H]

    \centering\includegraphics[width=1.0cm]{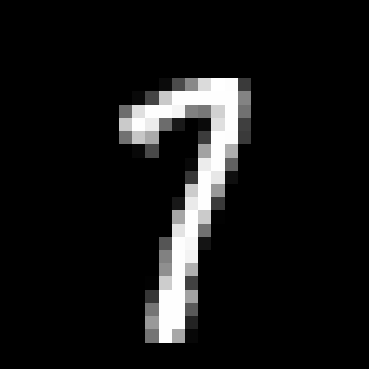}
    \centering\includegraphics[width=1.0cm]{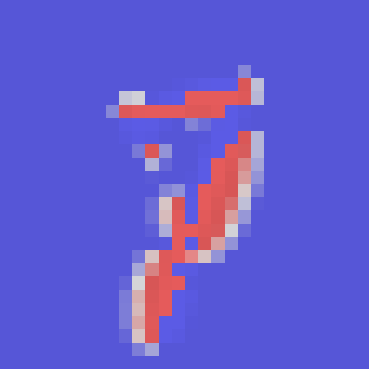}
    \centering\includegraphics[width=1.0cm]{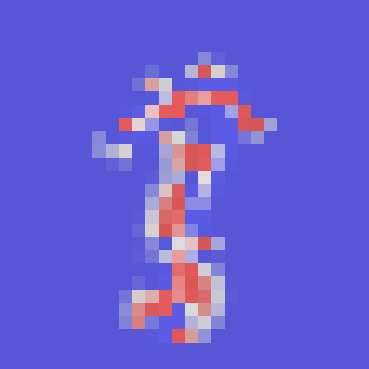}
    \centering\includegraphics[width=1.0cm]{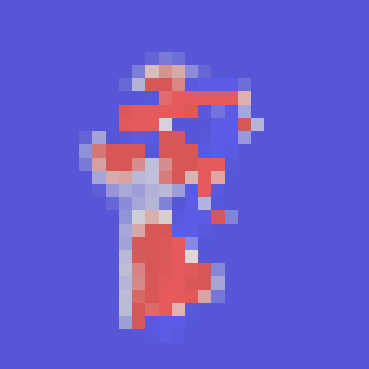}
    \centering\includegraphics[width=1.0cm]{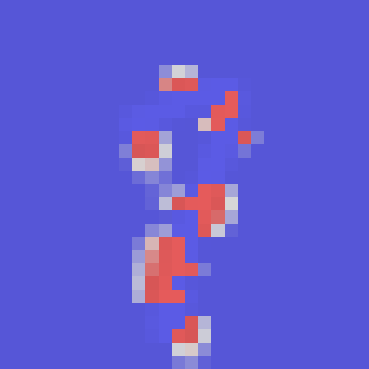}
    \centering\includegraphics[width=1.0cm]{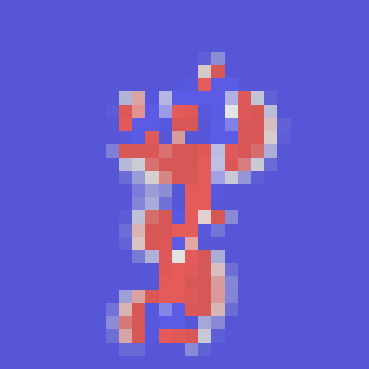}
    \centering\includegraphics[width=1.0cm]{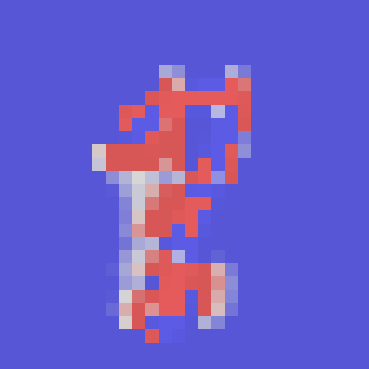}
    \centering\includegraphics[width=1.0cm]{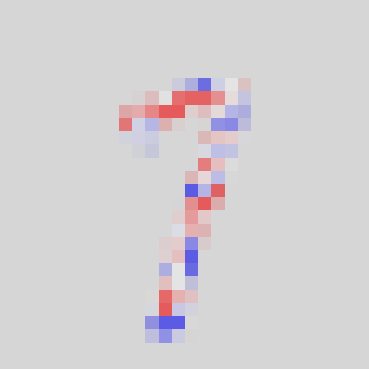}
    \centering\includegraphics[width=1.0cm]{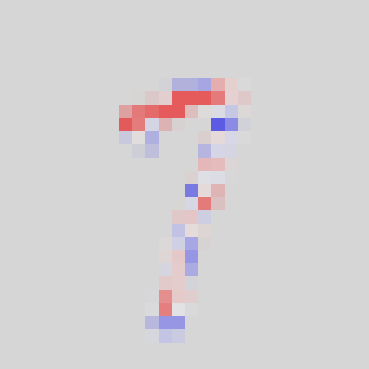}
    \centering\includegraphics[width=1.0cm]{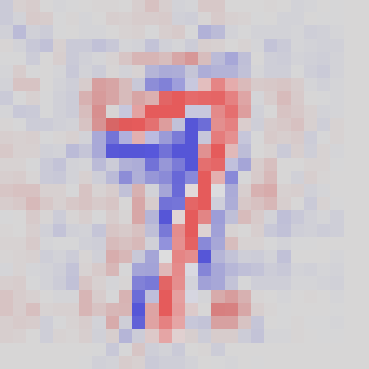}

    \centering\includegraphics[width=1.0cm]{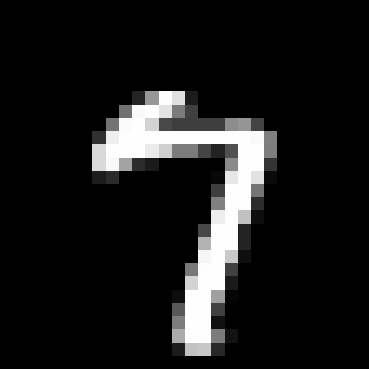}
    \centering\includegraphics[width=1.0cm]{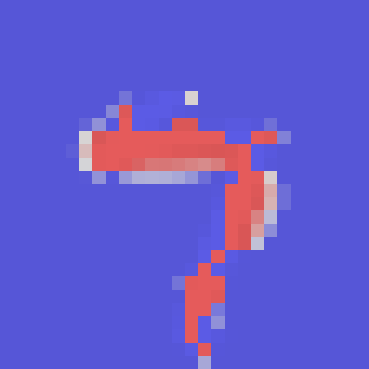}
    \centering\includegraphics[width=1.0cm]{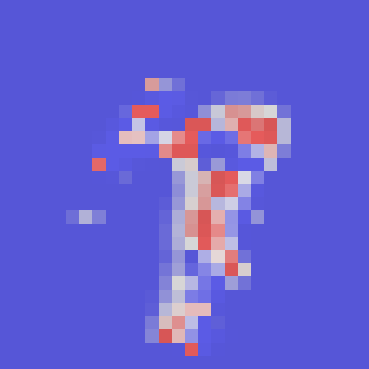}
    \centering\includegraphics[width=1.0cm]{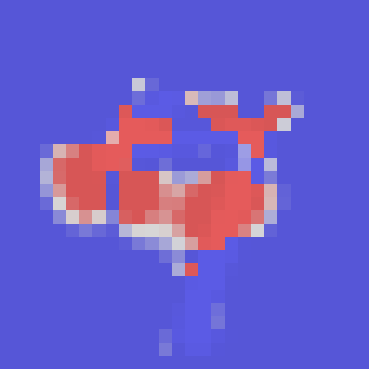}
    \centering\includegraphics[width=1.0cm]{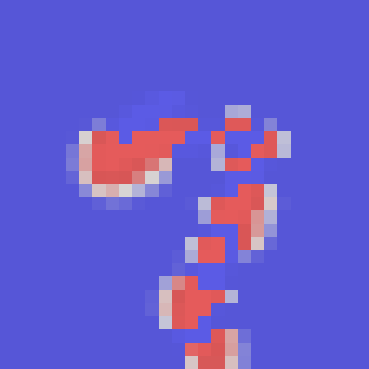}
    \centering\includegraphics[width=1.0cm]{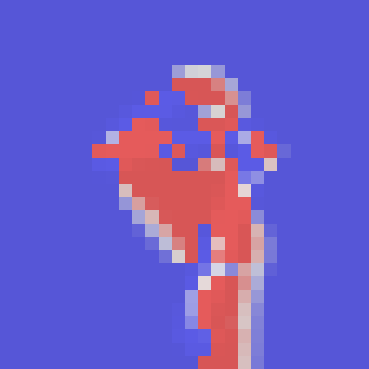}
    \centering\includegraphics[width=1.0cm]{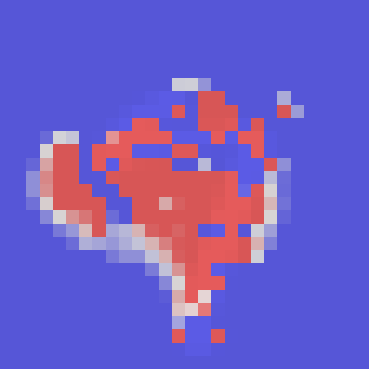}
    \centering\includegraphics[width=1.0cm]{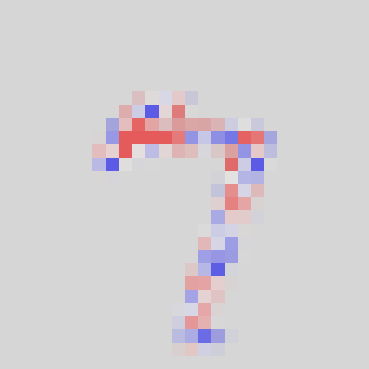}
    \centering\includegraphics[width=1.0cm]{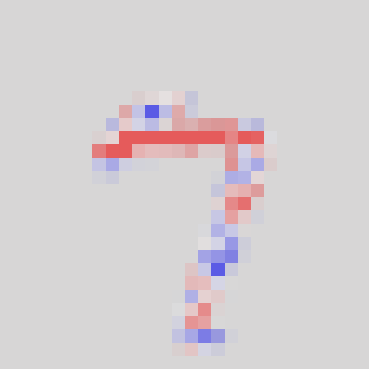}
    \centering\includegraphics[width=1.0cm]{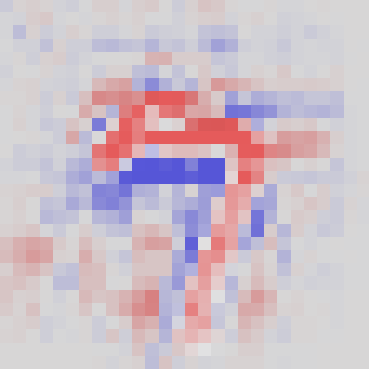}
    
    \centering\includegraphics[width=1.0cm]{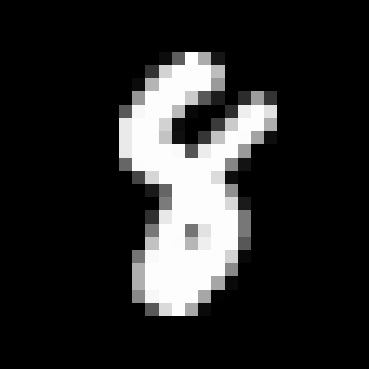}
    \centering\includegraphics[width=1.0cm]{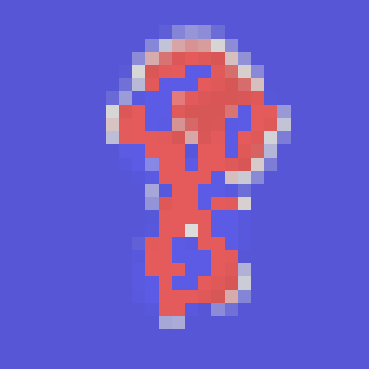}
    \centering\includegraphics[width=1.0cm]{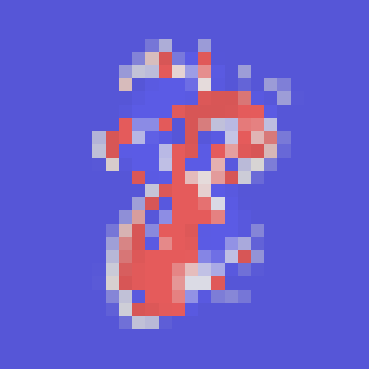}
    \centering\includegraphics[width=1.0cm]{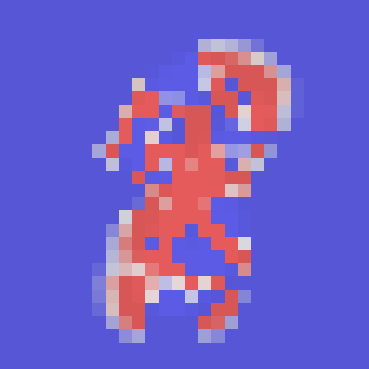}
    \centering\includegraphics[width=1.0cm]{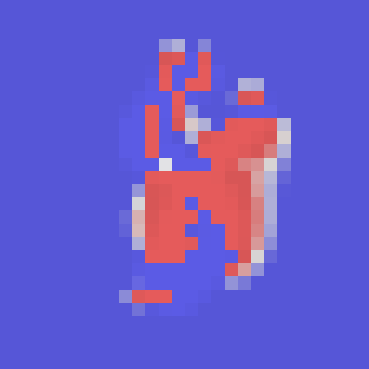}
    \centering\includegraphics[width=1.0cm]{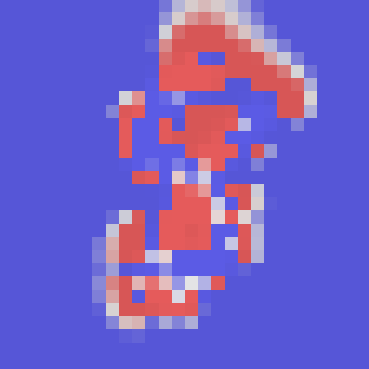}
    \centering\includegraphics[width=1.0cm]{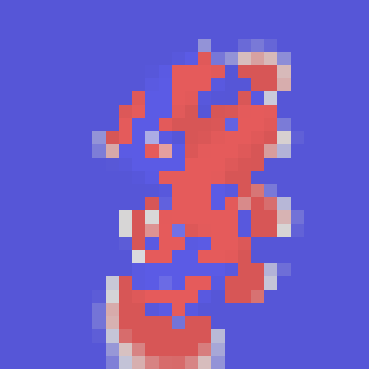}
    \centering\includegraphics[width=1.0cm]{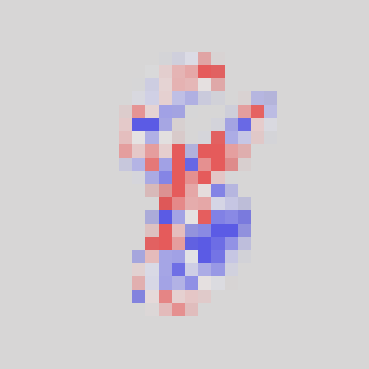}
    \centering\includegraphics[width=1.0cm]{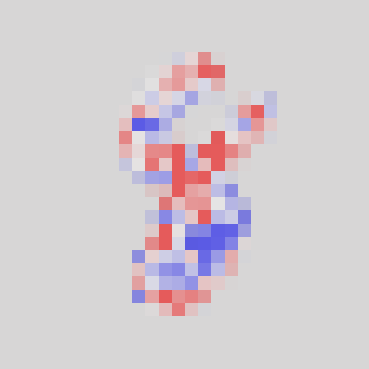}
    \centering\includegraphics[width=1.0cm]{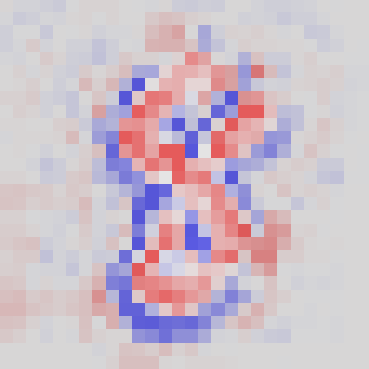}

    \centering\includegraphics[width=1.0cm]{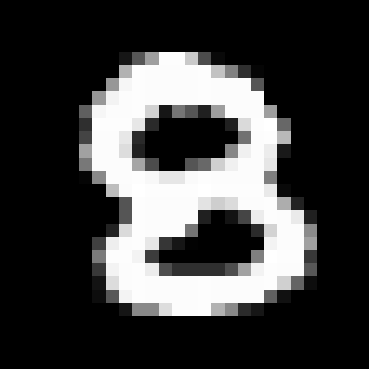}
    \centering\includegraphics[width=1.0cm]{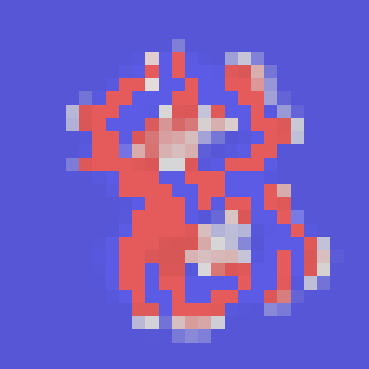}
    \centering\includegraphics[width=1.0cm]{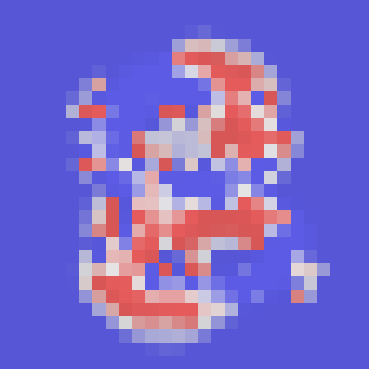}
    \centering\includegraphics[width=1.0cm]{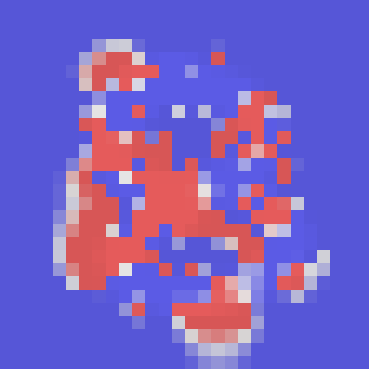}
    \centering\includegraphics[width=1.0cm]{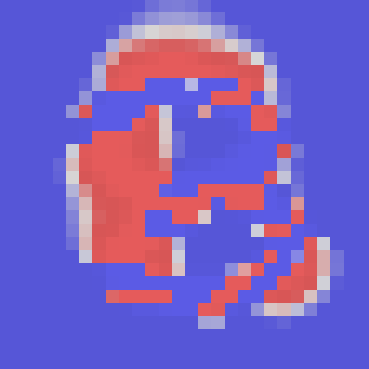}
    \centering\includegraphics[width=1.0cm]{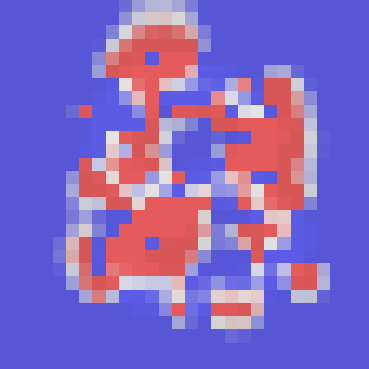}
    \centering\includegraphics[width=1.0cm]{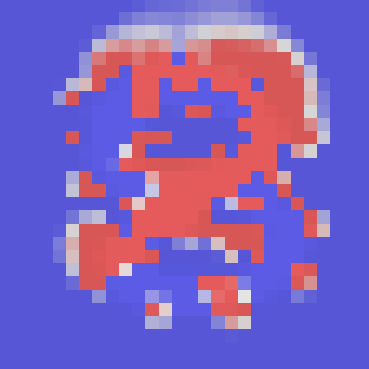}
    \centering\includegraphics[width=1.0cm]{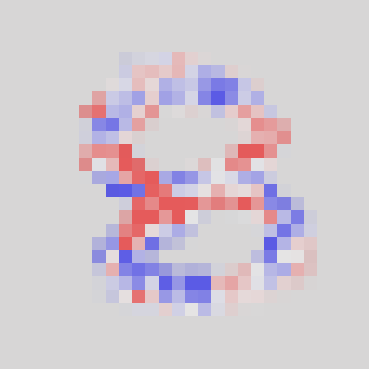}
    \centering\includegraphics[width=1.0cm]{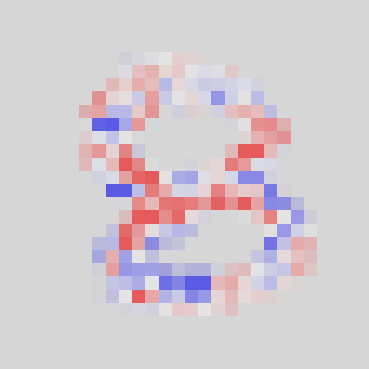}
    \centering\includegraphics[width=1.0cm]{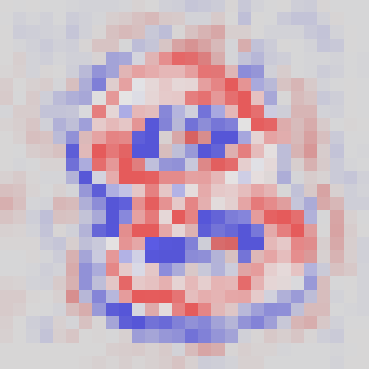}
    
    \centering\includegraphics[width=1.0cm]{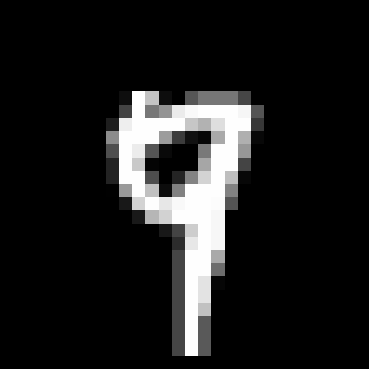}
    \centering\includegraphics[width=1.0cm]{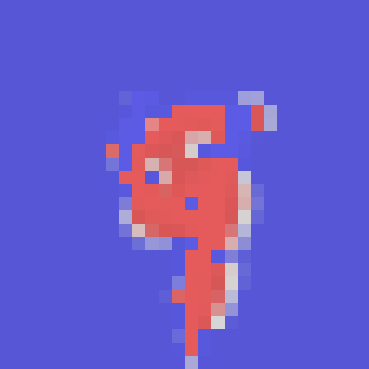}
    \centering\includegraphics[width=1.0cm]{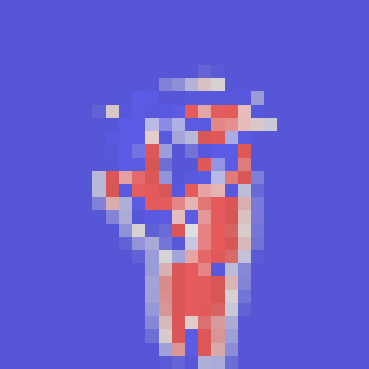}
    \centering\includegraphics[width=1.0cm]{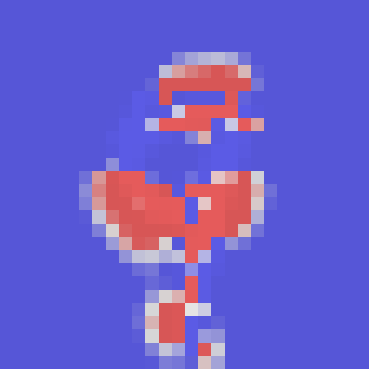}
    \centering\includegraphics[width=1.0cm]{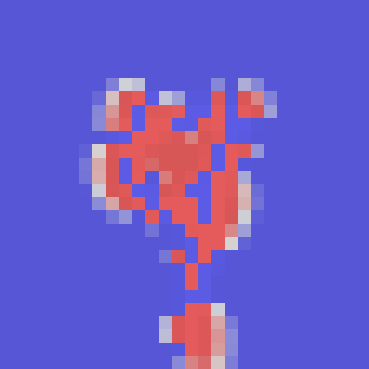}
    \centering\includegraphics[width=1.0cm]{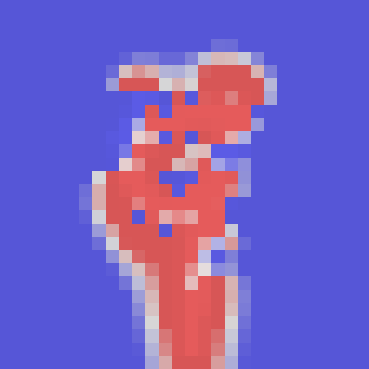}
    \centering\includegraphics[width=1.0cm]{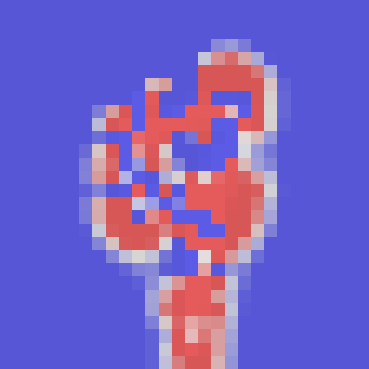}
    \centering\includegraphics[width=1.0cm]{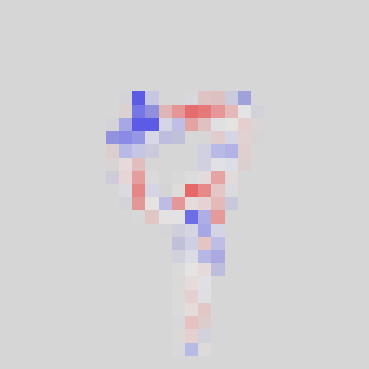}
    \centering\includegraphics[width=1.0cm]{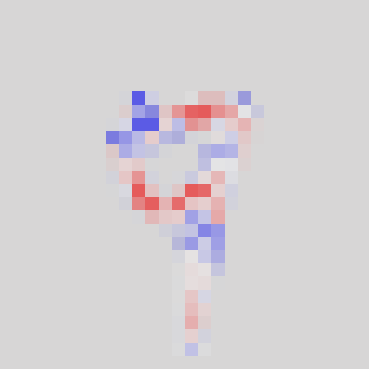}
    \centering\includegraphics[width=1.0cm]{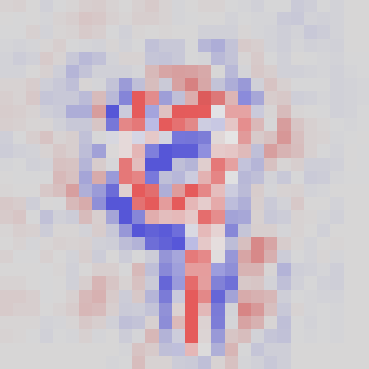}
    
    \centering\includegraphics[width=1.0cm]{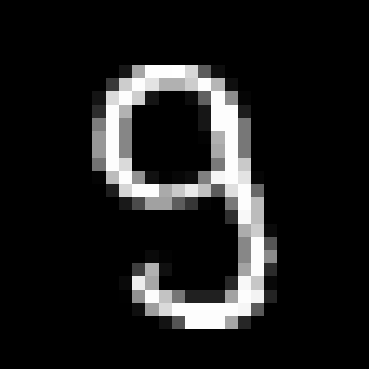}
    \centering\includegraphics[width=1.0cm]{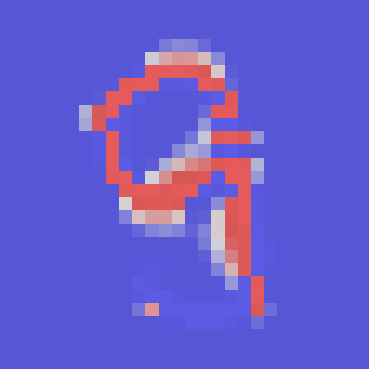}
    \centering\includegraphics[width=1.0cm]{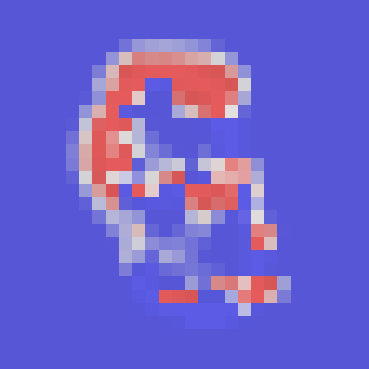}
    \centering\includegraphics[width=1.0cm]{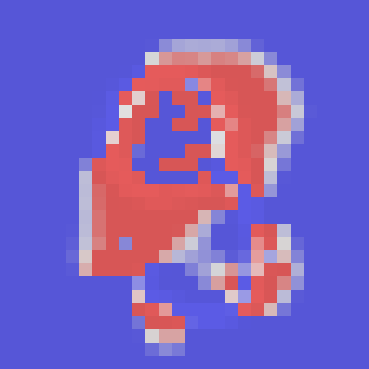}
    \centering\includegraphics[width=1.0cm]{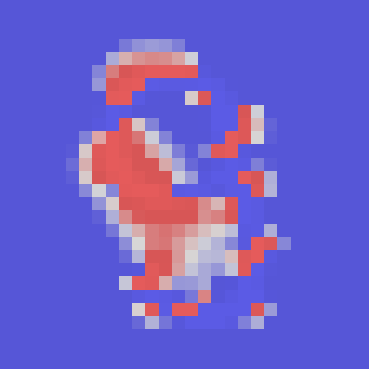}
    \centering\includegraphics[width=1.0cm]{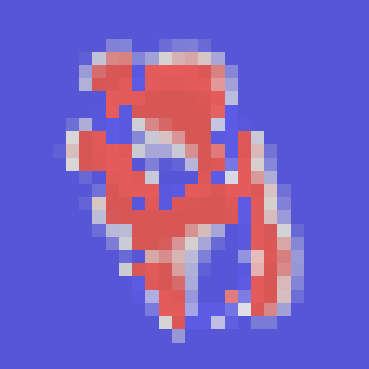}
    \centering\includegraphics[width=1.0cm]{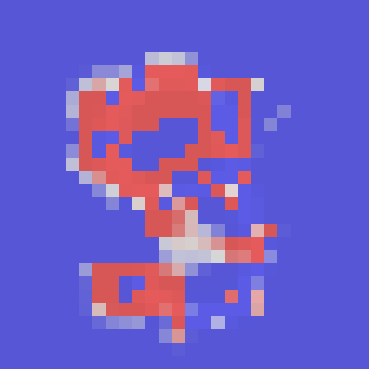}
    \centering\includegraphics[width=1.0cm]{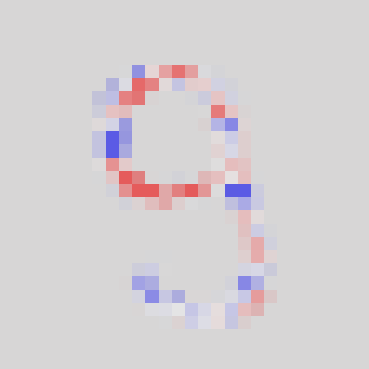}
    \centering\includegraphics[width=1.0cm]{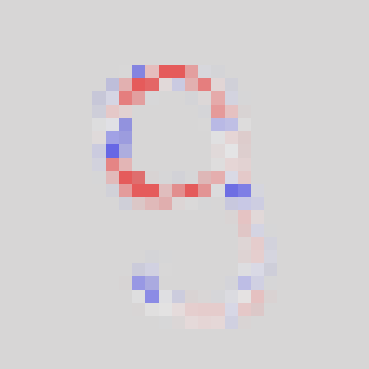}
    \centering\includegraphics[width=1.0cm]{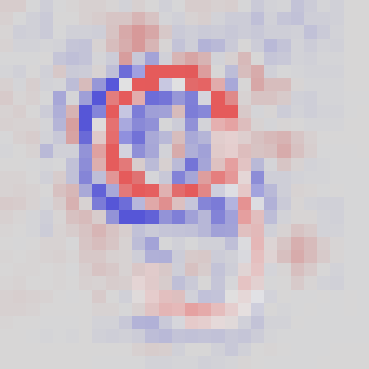}
    
    \caption{Examples of counterfactuals for samples from MNIST. Left to right: original image, SSR flip, SSR VAE, SSR knockoff, SDR flip, SDR VAE, SDR knockoff, Gradients $\times$ Input, Integrated Gradients, Guided GradCAM.}
\end{figure}

\hfill
% \vfill
% -------------------------------------------------------------------
\subsubsection{Complementary MNIST 3/8 and 5/6}
% \label{app:compl_38}

\begin{figure}[H]
\centering
\begin{minipage}[t]{.48\textwidth}
\centering
    \includegraphics[width=1.0cm]{images/complementary_mnist_38/1579/orig.png}
    \includegraphics[width=1.0cm]{images/complementary_mnist_38/1579/flip_ssr_0005_tv01.png}
    \includegraphics[width=1.0cm]{images/complementary_mnist_38/1579/vae_ssr_0005_tv01.png}
    \includegraphics[width=1.0cm]{images/complementary_mnist_38/1579/knockoff_seed_2_ssr_0005_tv01_k3z.png}
    \includegraphics[width=1.0cm]{images/complementary_mnist_38/1579/flip_sdr_0005_tv01.png}
    \includegraphics[width=1.0cm]{images/complementary_mnist_38/1579/vae_sdr_0005_tv01.png}
    \includegraphics[width=1.0cm]{images/complementary_mnist_38/1579/knockoff_seed_2_sdr_0005_tv01_k3z.png}

    \includegraphics[width=1.0cm]{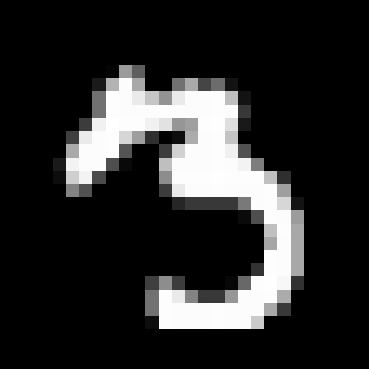}
    \includegraphics[width=1.0cm]{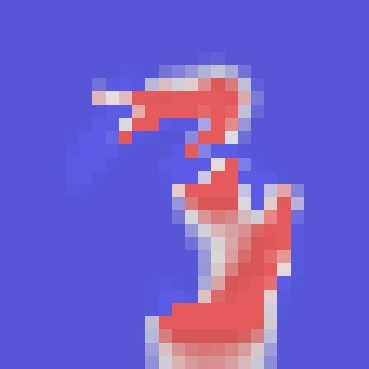}
    \includegraphics[width=1.0cm]{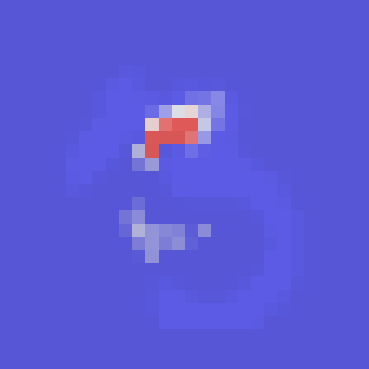}
    \includegraphics[width=1.0cm]{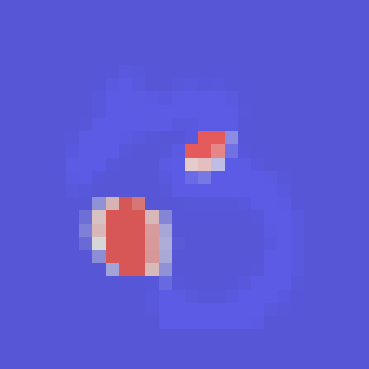}
    \includegraphics[width=1.0cm]{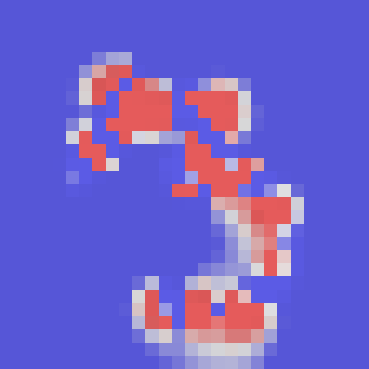}
    \includegraphics[width=1.0cm]{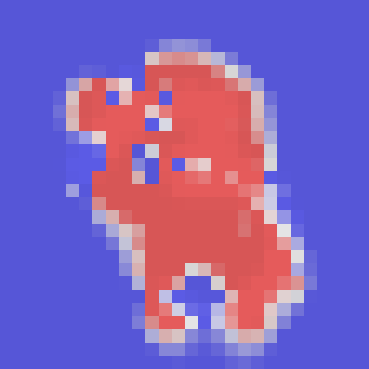}
    \includegraphics[width=1.0cm]{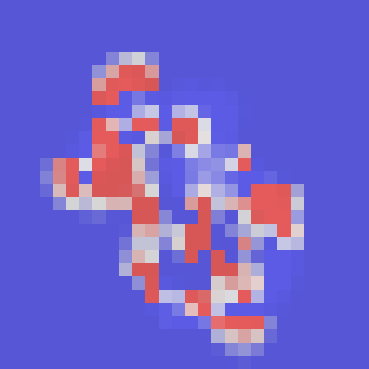}

    \includegraphics[width=1.0cm]{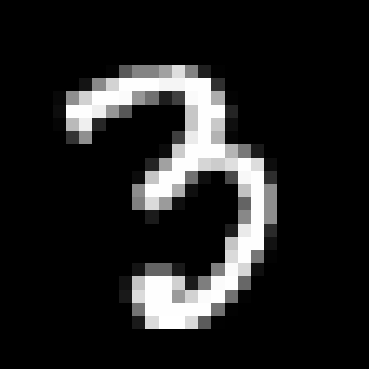}
    \includegraphics[width=1.0cm]{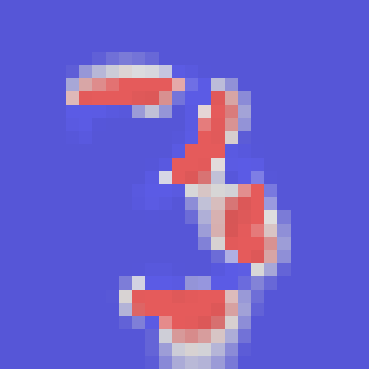}
    \includegraphics[width=1.0cm]{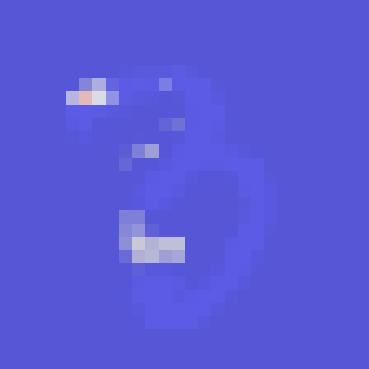}
    \includegraphics[width=1.0cm]{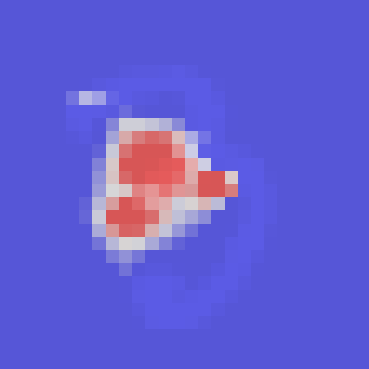}
    \includegraphics[width=1.0cm]{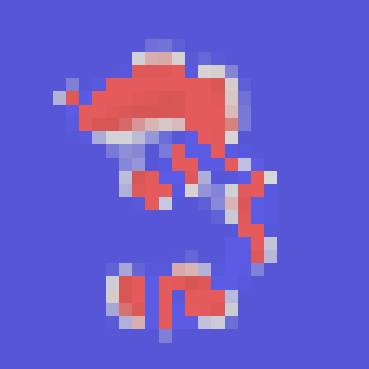}
    \includegraphics[width=1.0cm]{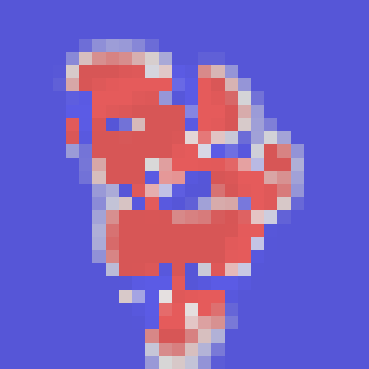}
    \includegraphics[width=1.0cm]{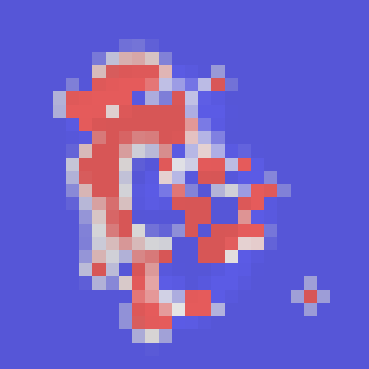}
    
    \includegraphics[width=1.0cm]{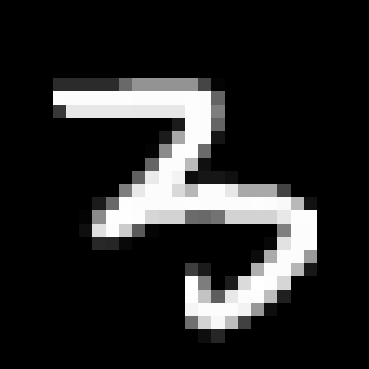}
    \includegraphics[width=1.0cm]{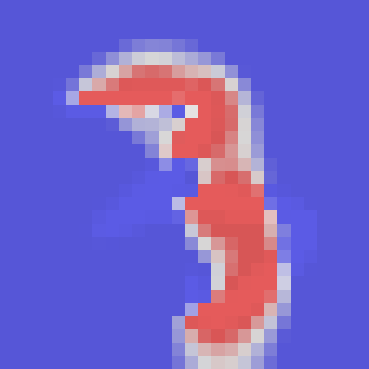}
    \includegraphics[width=1.0cm]{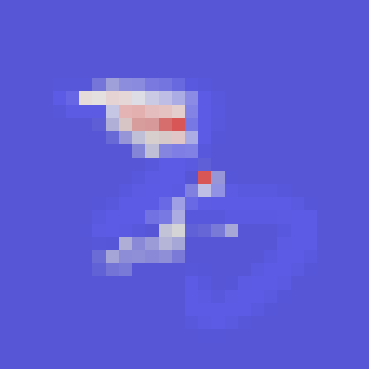}
    \includegraphics[width=1.0cm]{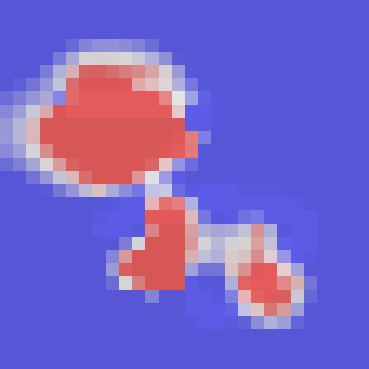}
    \includegraphics[width=1.0cm]{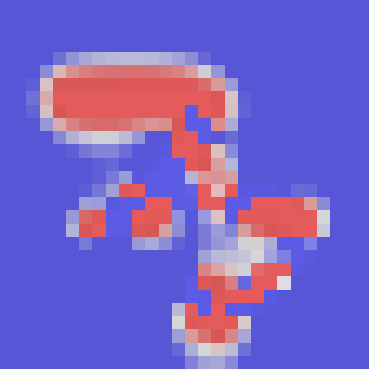}
    \includegraphics[width=1.0cm]{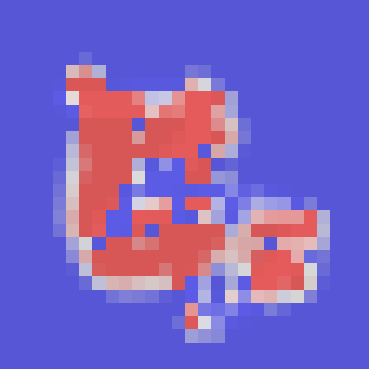}
    \includegraphics[width=1.0cm]{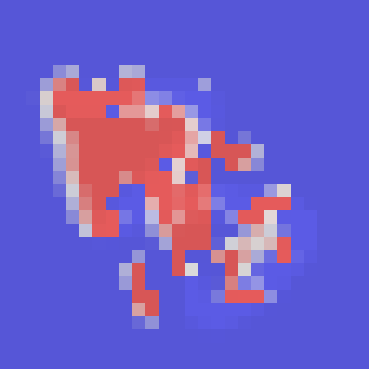}

    \includegraphics[width=1.0cm]{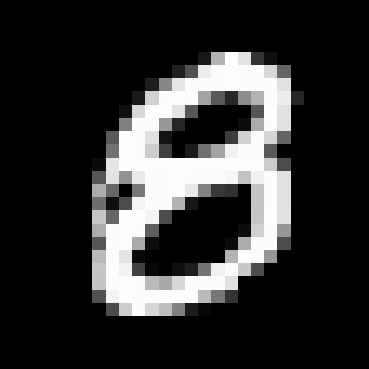}
    \includegraphics[width=1.0cm]{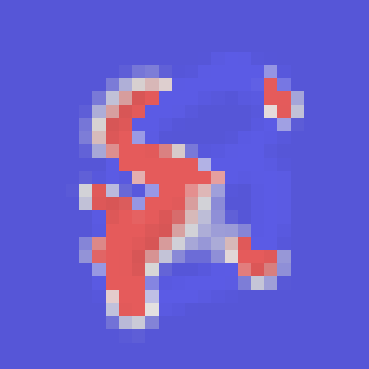}
    \includegraphics[width=1.0cm]{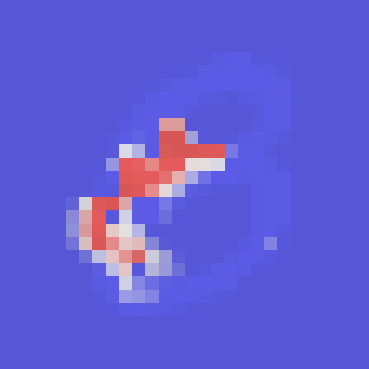}
    \includegraphics[width=1.0cm]{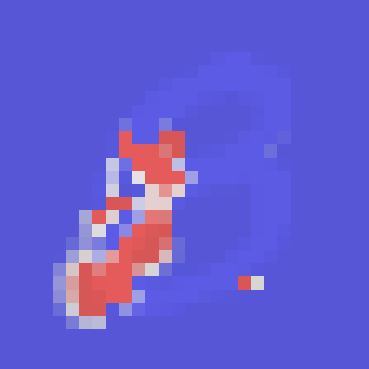}
    \includegraphics[width=1.0cm]{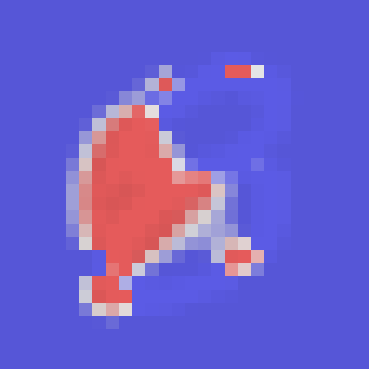}
    \includegraphics[width=1.0cm]{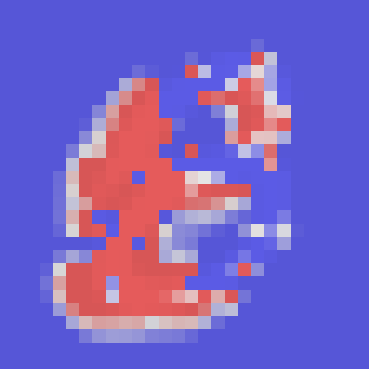}
    \includegraphics[width=1.0cm]{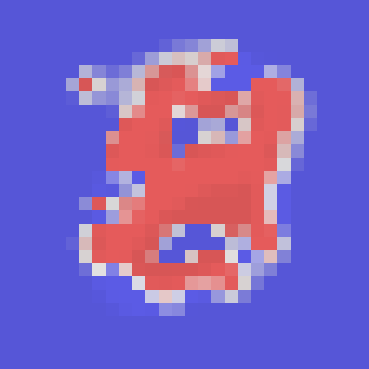}
    
    \includegraphics[width=1.0cm]{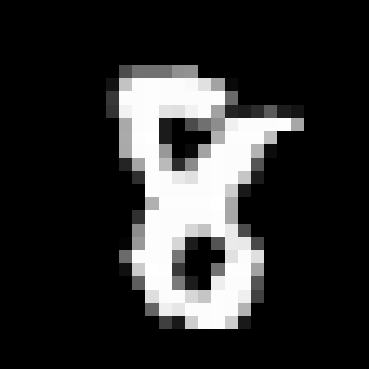}
    \includegraphics[width=1.0cm]{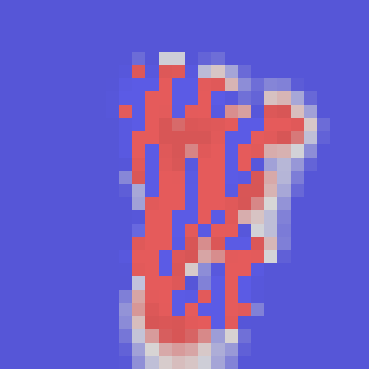}
    \includegraphics[width=1.0cm]{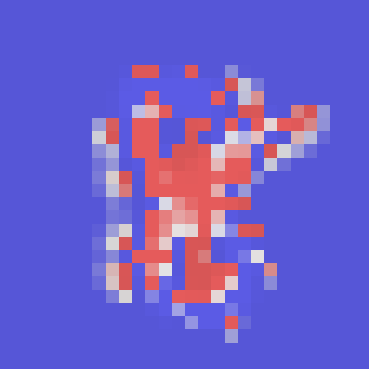}
    \includegraphics[width=1.0cm]{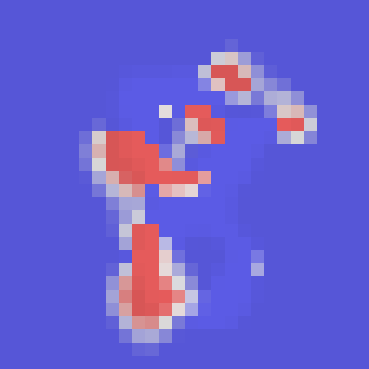}
    \includegraphics[width=1.0cm]{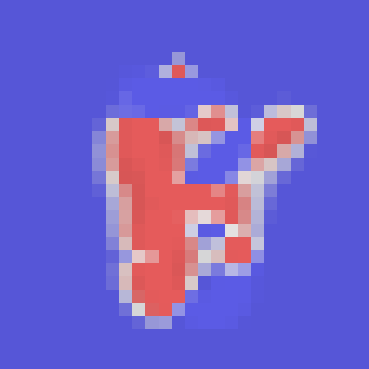}
    \includegraphics[width=1.0cm]{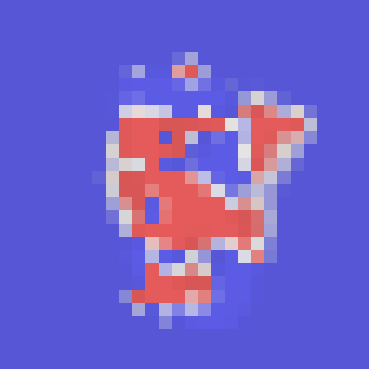}
    \includegraphics[width=1.0cm]{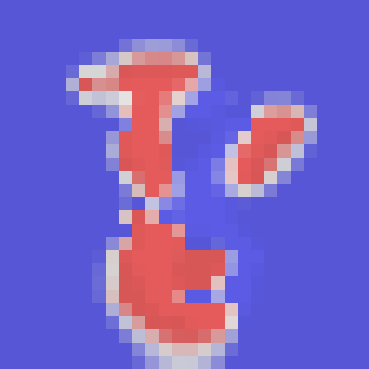}
    
    \includegraphics[width=1.0cm]{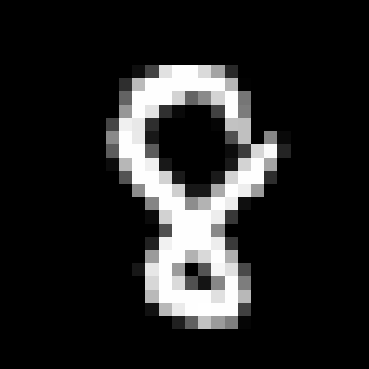}
    \includegraphics[width=1.0cm]{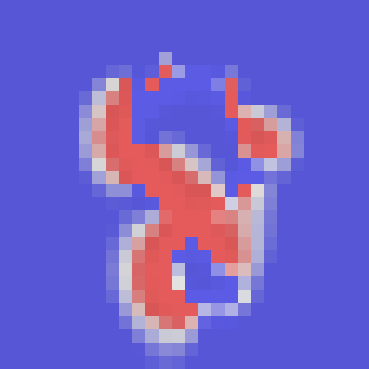}
    \includegraphics[width=1.0cm]{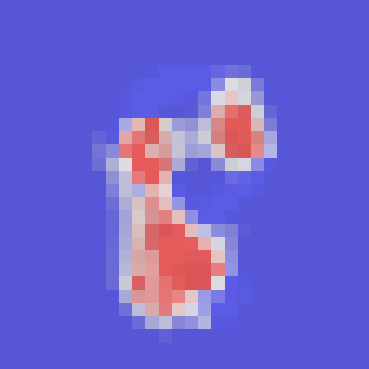}
    \includegraphics[width=1.0cm]{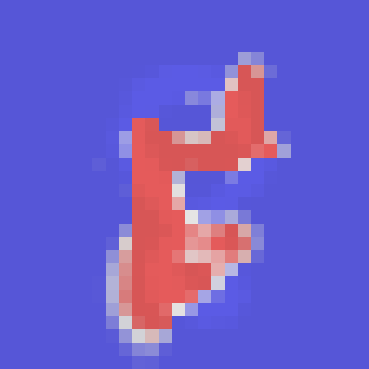}
    \includegraphics[width=1.0cm]{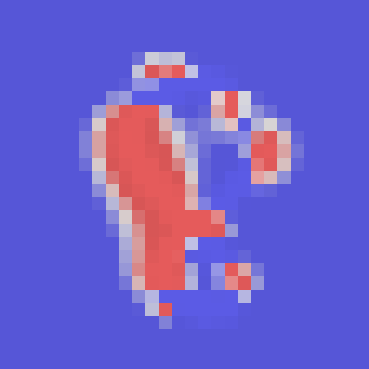}
    \includegraphics[width=1.0cm]{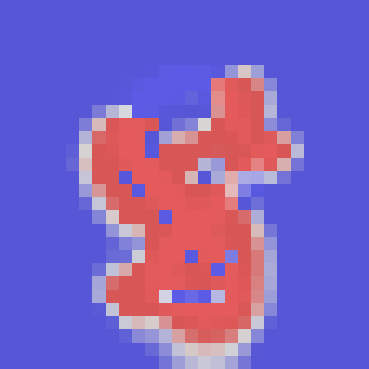}
    \includegraphics[width=1.0cm]{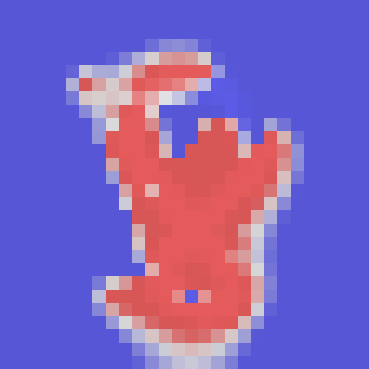}
    
    \includegraphics[width=1.0cm]{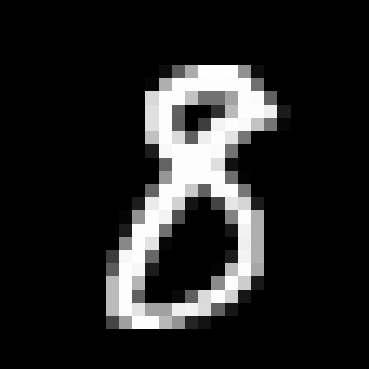}
    \includegraphics[width=1.0cm]{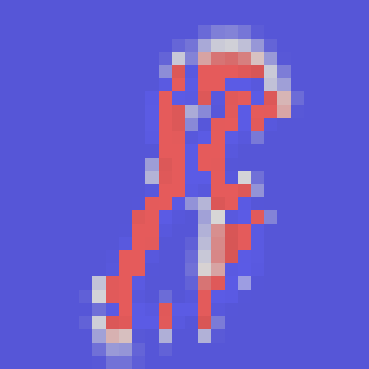}
    \includegraphics[width=1.0cm]{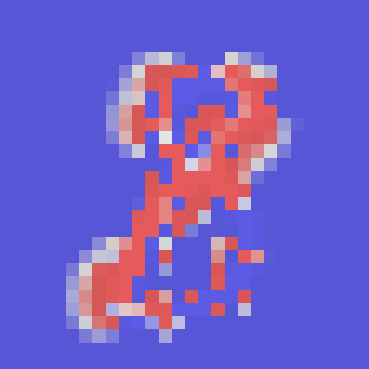}
    \includegraphics[width=1.0cm]{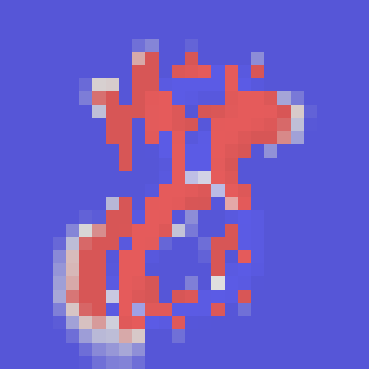}
    \includegraphics[width=1.0cm]{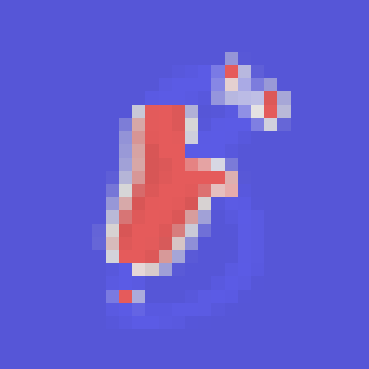}
    \includegraphics[width=1.0cm]{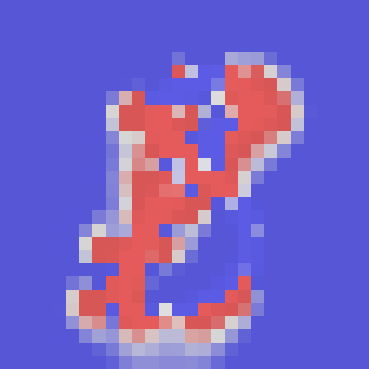}
    \includegraphics[width=1.0cm]{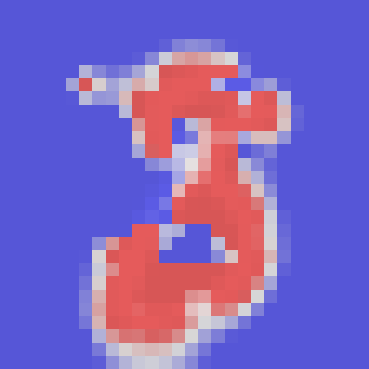}
    \caption{Examples of counterfactuals for numbers 3 and 8 from the complementary MNIST 3/8 subset. Left to right: original image, SSR flip, SSR VAE, SSR knockoff, SDR flip, SDR VAE, SDR knockoff.}
\end{minipage} \hfill %
\begin{minipage}[t]{.48\textwidth}
\centering
\includegraphics[width=1.0cm]{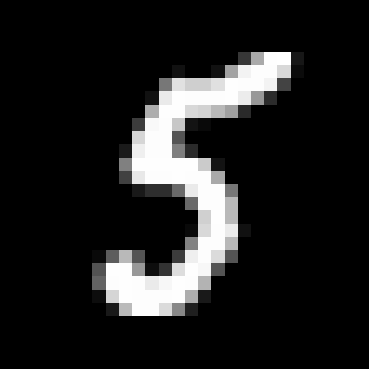}
    \includegraphics[width=1.0cm]{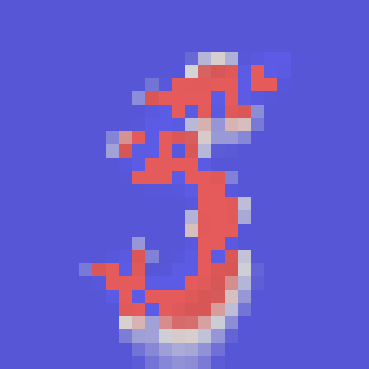}
    \includegraphics[width=1.0cm]{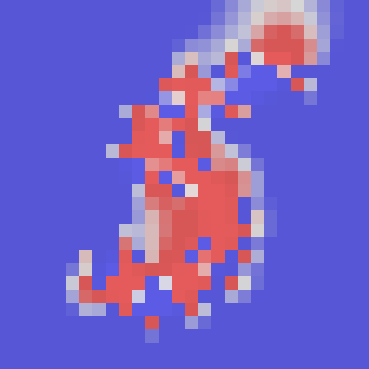}
    \includegraphics[width=1.0cm]{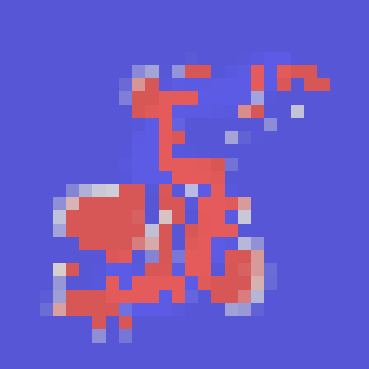}
    \includegraphics[width=1.0cm]{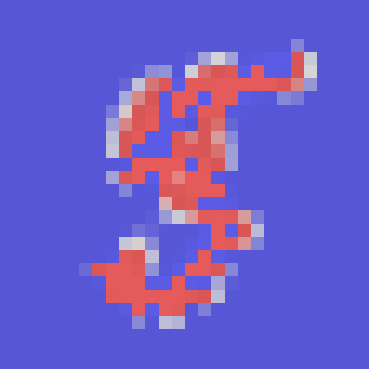}
    \includegraphics[width=1.0cm]{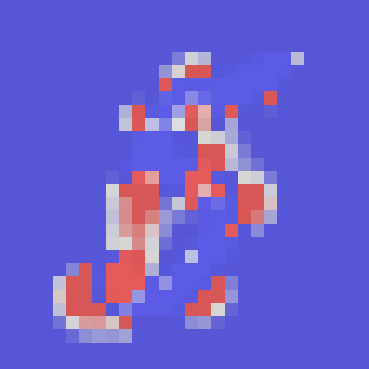}
    \includegraphics[width=1.0cm]{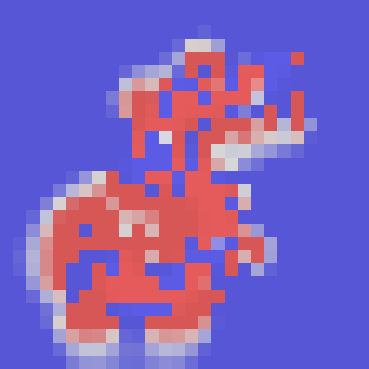}

    \includegraphics[width=1.0cm]{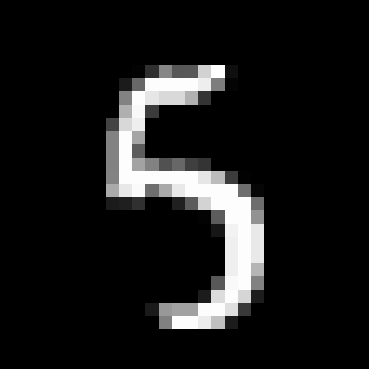}
    \includegraphics[width=1.0cm]{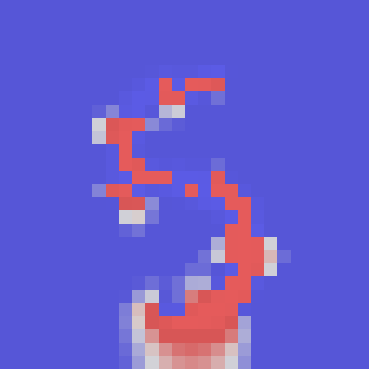}
    \includegraphics[width=1.0cm]{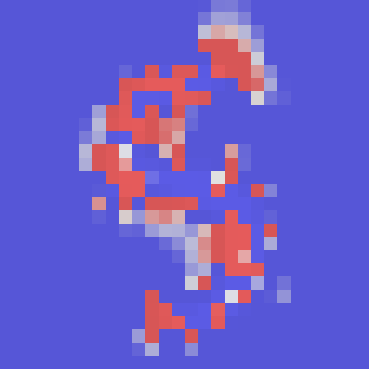}
    \includegraphics[width=1.0cm]{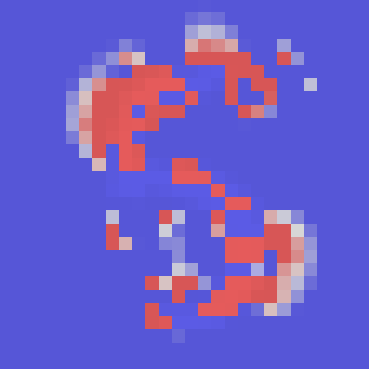}
    \includegraphics[width=1.0cm]{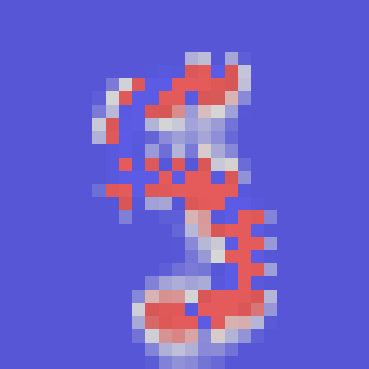}
    \includegraphics[width=1.0cm]{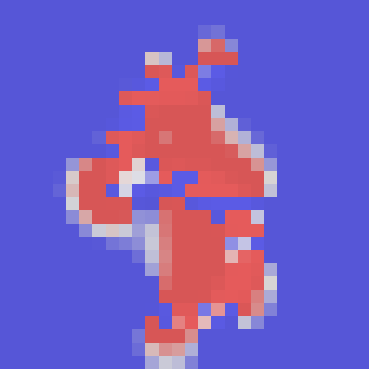}
    \includegraphics[width=1.0cm]{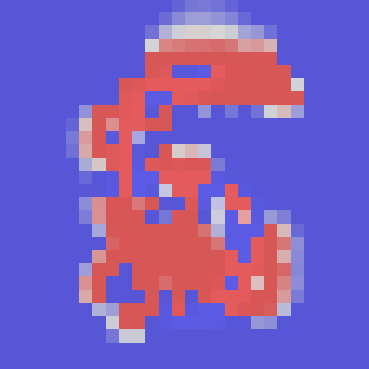}
    
    \includegraphics[width=1.0cm]{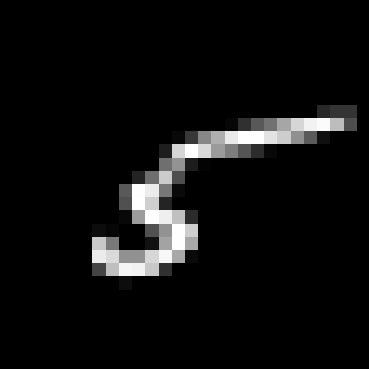}
    \includegraphics[width=1.0cm]{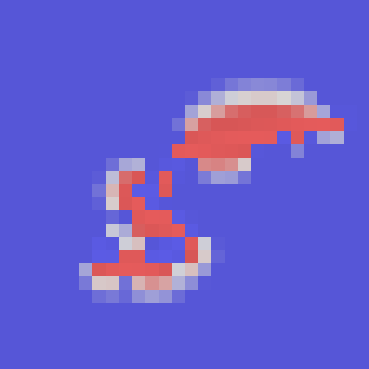}
    \includegraphics[width=1.0cm]{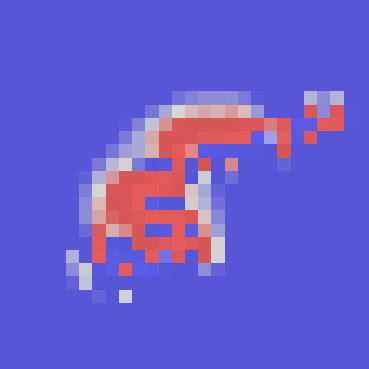}
    \includegraphics[width=1.0cm]{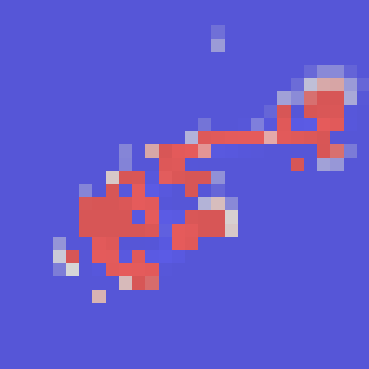}
    \includegraphics[width=1.0cm]{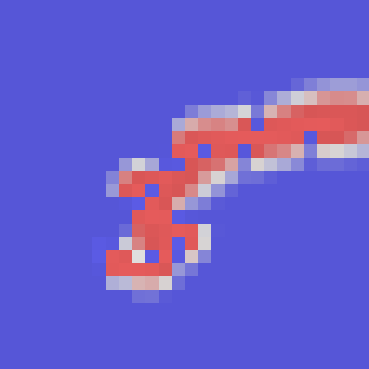}
    \includegraphics[width=1.0cm]{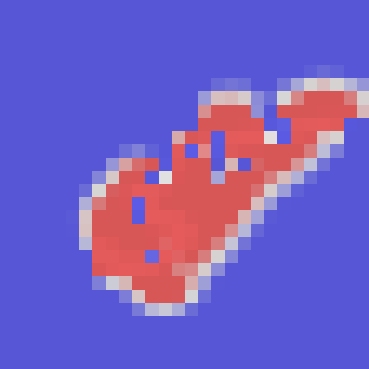}
    \includegraphics[width=1.0cm]{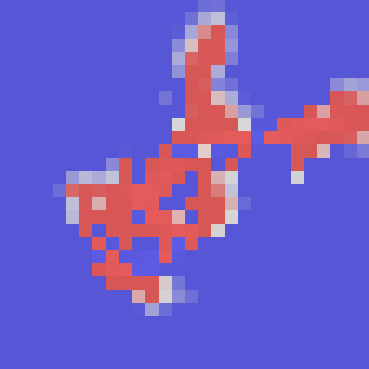}
    
    \includegraphics[width=1.0cm]{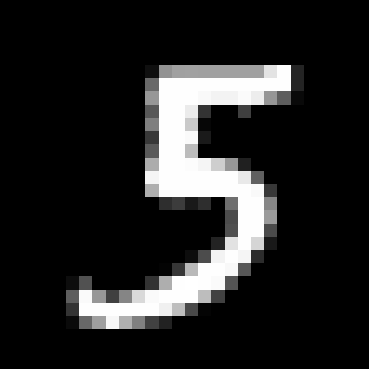}
    \includegraphics[width=1.0cm]{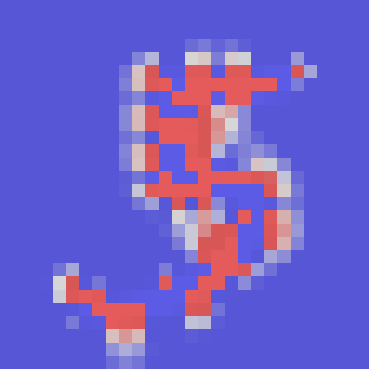}
    \includegraphics[width=1.0cm]{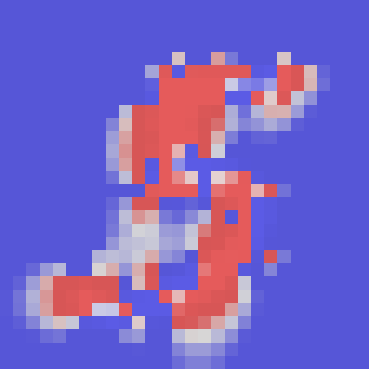}
    \includegraphics[width=1.0cm]{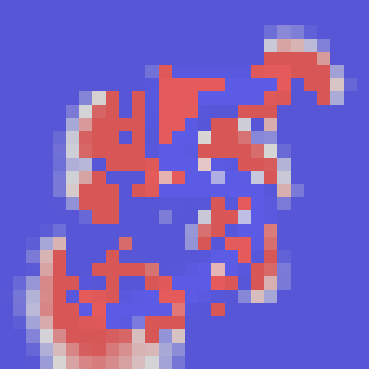}
    \includegraphics[width=1.0cm]{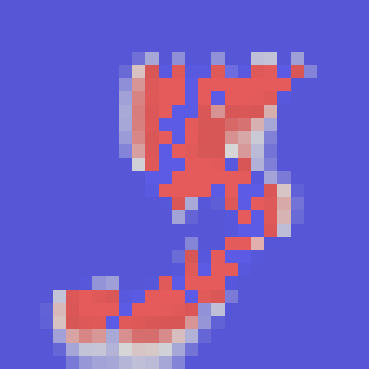}
    \includegraphics[width=1.0cm]{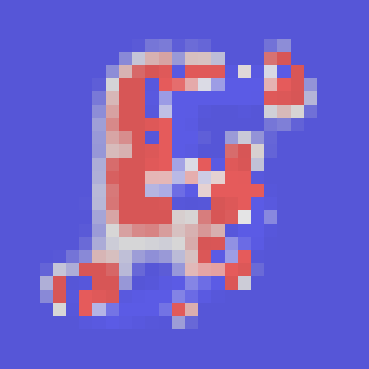}
    \includegraphics[width=1.0cm]{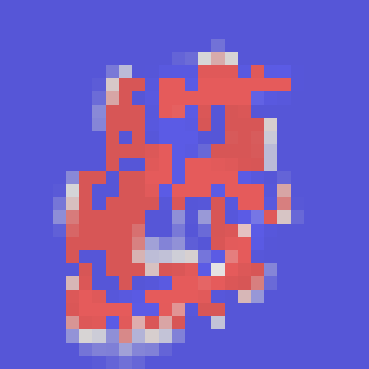}

    \includegraphics[width=1.0cm]{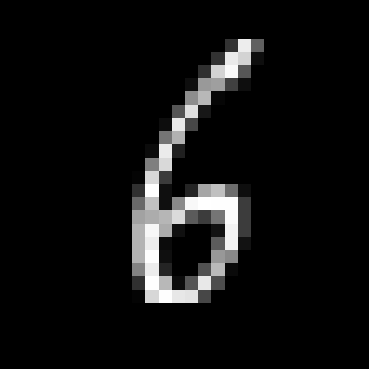}
    \includegraphics[width=1.0cm]{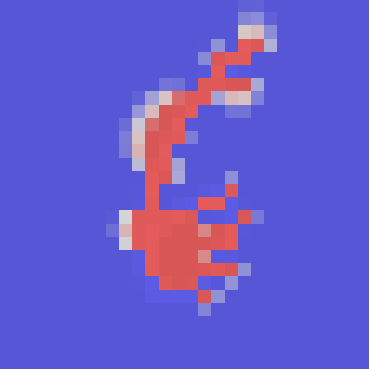}
    \includegraphics[width=1.0cm]{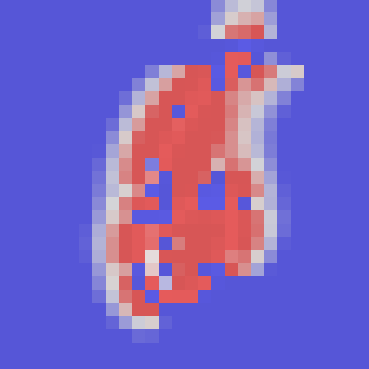}
    \includegraphics[width=1.0cm]{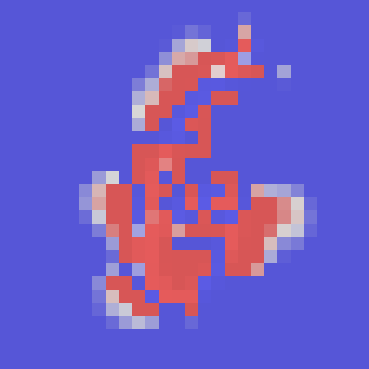}
    \includegraphics[width=1.0cm]{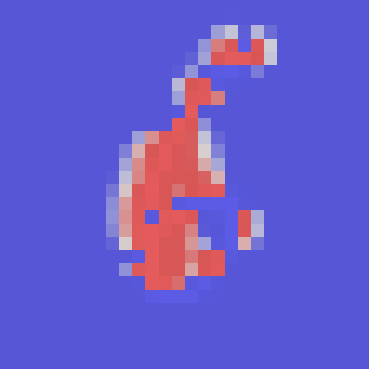}
    \includegraphics[width=1.0cm]{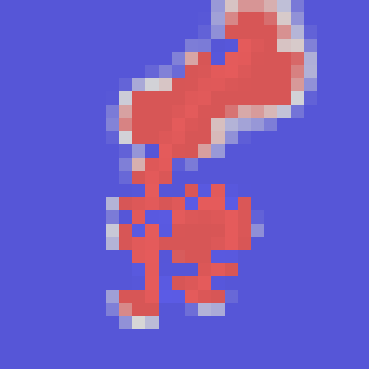}
    \includegraphics[width=1.0cm]{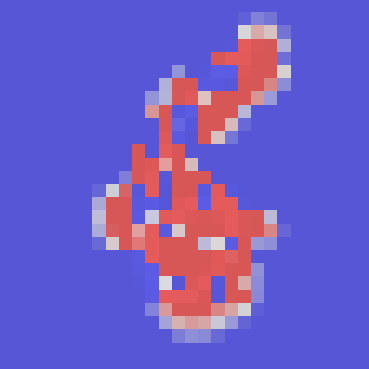}
    
    \includegraphics[width=1.0cm]{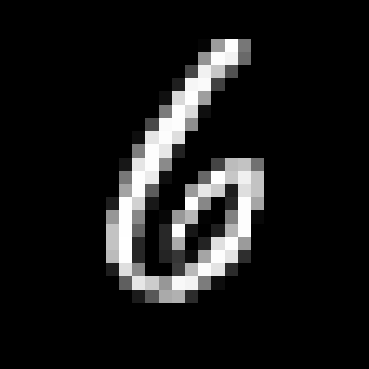}
    \includegraphics[width=1.0cm]{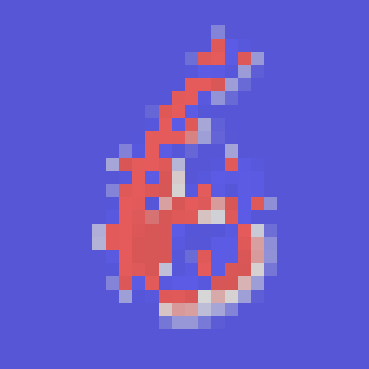}
    \includegraphics[width=1.0cm]{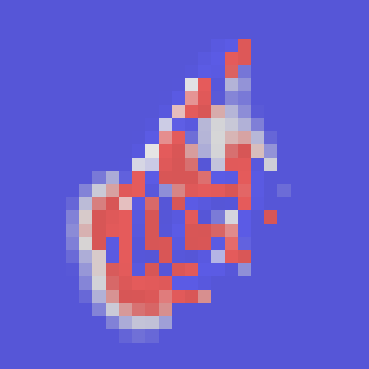}
    \includegraphics[width=1.0cm]{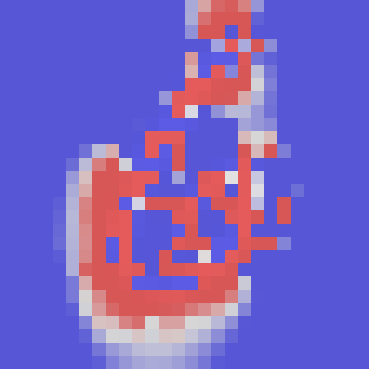}
    \includegraphics[width=1.0cm]{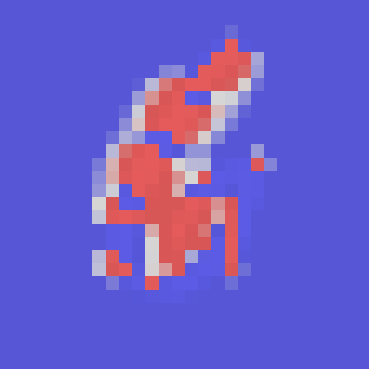}
    \includegraphics[width=1.0cm]{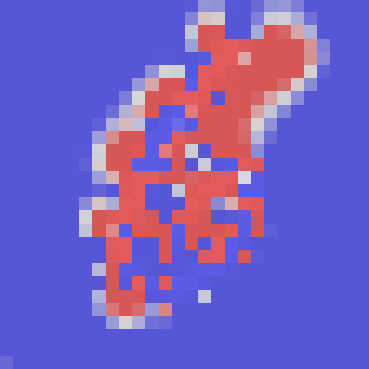}
    \includegraphics[width=1.0cm]{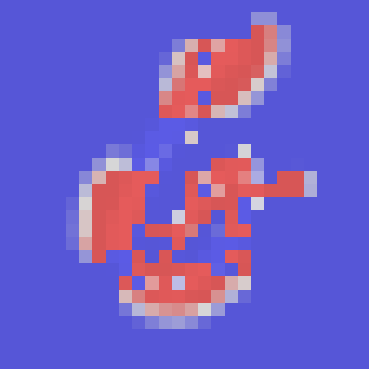}
    
    \includegraphics[width=1.0cm]{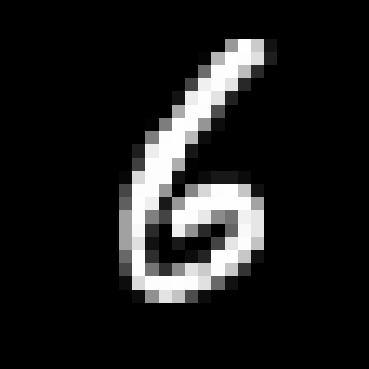}
    \includegraphics[width=1.0cm]{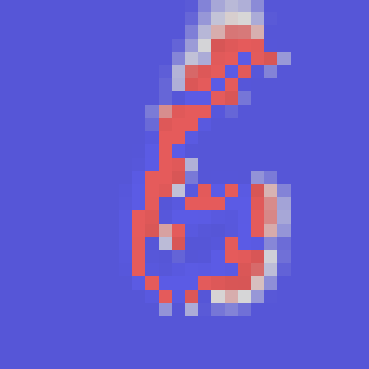}
    \includegraphics[width=1.0cm]{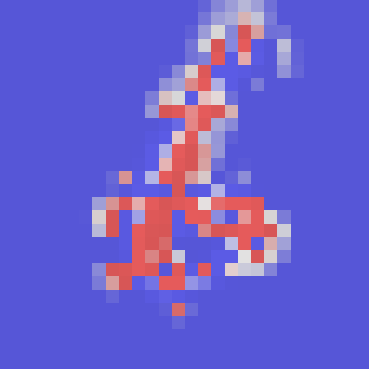}
    \includegraphics[width=1.0cm]{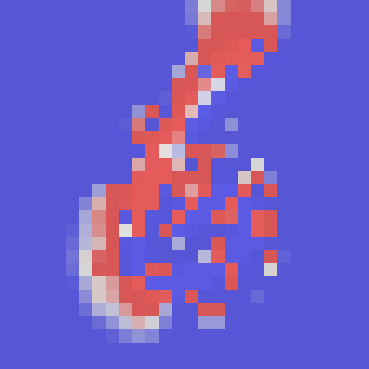}
    \includegraphics[width=1.0cm]{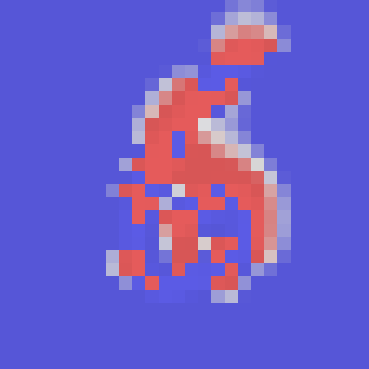}
    \includegraphics[width=1.0cm]{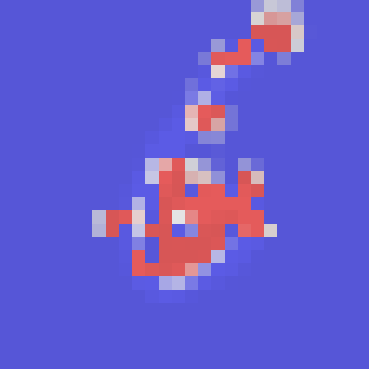}
    \includegraphics[width=1.0cm]{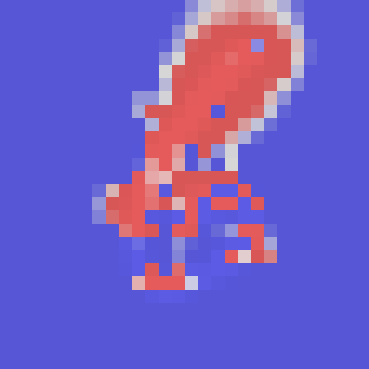}
    
    \includegraphics[width=1.0cm]{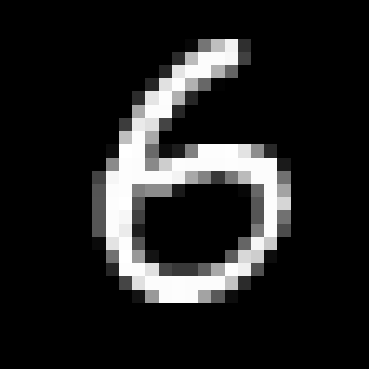}
    \includegraphics[width=1.0cm]{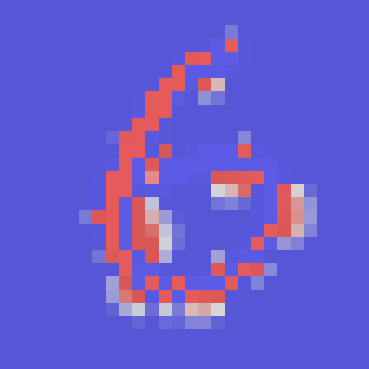}
    \includegraphics[width=1.0cm]{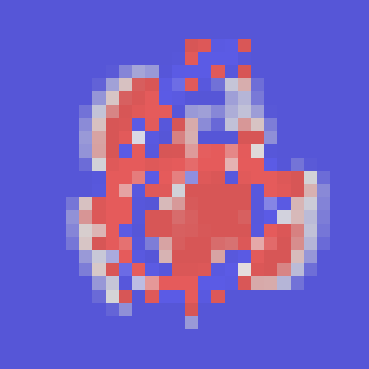}
    \includegraphics[width=1.0cm]{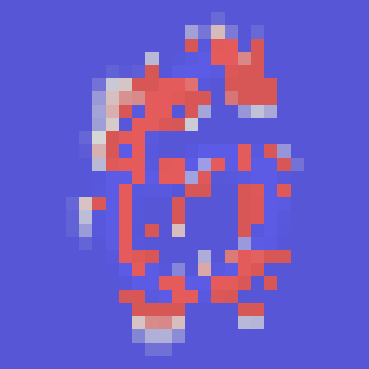}
    \includegraphics[width=1.0cm]{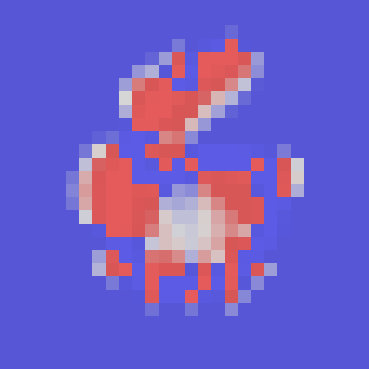}
    \includegraphics[width=1.0cm]{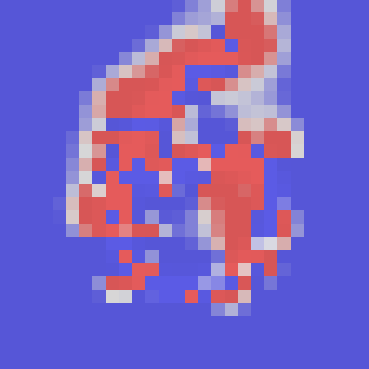}
    \includegraphics[width=1.0cm]{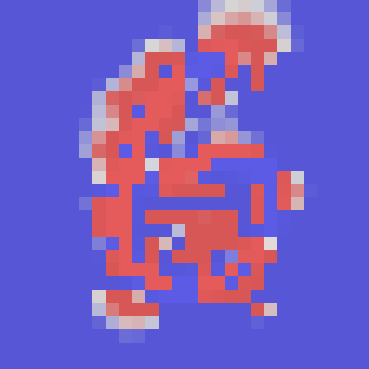}

    \caption{Examples of counterfactuals for numbers 5 and 6 from the complementary MNIST 5/6 subset. Left to right: original image, SSR flip, SSR VAE, SSR knockoff, SDR flip, SDR VAE, SDR knockoff.}
\end{minipage}
\end{figure}

\end{document}